\documentclass{article} 
\usepackage{iclr2022_conference,times}

\usepackage{longtable}
\usepackage{float}
\usepackage{placeins}
\usepackage{wrapfig,lipsum,booktabs}
\usepackage{anyfontsize}

\usepackage{amsmath,amsfonts,bm}

\def\eqref#1{equation~\ref{#1}}

\def\1{\bm{1}}

\DeclareMathAlphabet{\mathsfit}{\encodingdefault}{\sfdefault}{m}{sl}
\SetMathAlphabet{\mathsfit}{bold}{\encodingdefault}{\sfdefault}{bx}{n}

\newcommand{\inserttfiifkappa}{
\begin{table*}
\centering
\begin{tabular}{ll|lllll}
\hline
\hline
 $\kappa$ &  $\tau$ & mean recall & median recall & mean $\delta_l$ & median $\delta_l$ & $\delta_l > 10$ \\
 \hline
 5000 & 1000 & 94.3 & 98.9 & 8.6 & 1.2 & 20 \\
 10000 & 1000 &  94.2 & 98.9 & 7.3 & 0.7 & 16 \\
 20000 & 1000 & 94.1 & 98.7 & 5.7 & 0.4 & 14 \\
 80000 & 1000 & 93.4 & 98.5 & 2.9 & 0.1 & 8 \\
 80000 & 2000 & 94.9 & 99.3 & 6.8 &  0.4 &  21  \\
\hline
\end{tabular}

\caption{Investigating the effect of varying the IIF-thresholding parameter $\kappa$ on the recall and distractibility ($\delta_l$) of \tfiif{} filtering, over 1598 languages. As we increase $\kappa$, the distractibility steadily trends down, with only small losses in recall. Increasing $\tau$ recoups lost recall but increases distractibility.}
\label{tab:tfiif_kappa}
\end{table*}
}

\newcommand{\insertmtcapacity}{\begin{table}[t]
\centering
\begin{tabular}{lcccccccccc}
\toprule
                   & \multicolumn{2}{c}{en-mfa}    & \multicolumn{2}{c}{en-ff}     & \multicolumn{2}{c}{en-mni}    & \multicolumn{2}{c}{en-kri}    & \multicolumn{2}{c}{en-mad} \\
Mono Size          & \multicolumn{2}{c}{7k}      & \multicolumn{2}{c}{86k}     & \multicolumn{2}{c}{106k}    & \multicolumn{2}{c}{129k}   & \multicolumn{2}{c}{138k} \\

            Direction       & $\leftarrow$ & $\rightarrow$ & $\leftarrow$  & $\rightarrow$ & $\leftarrow$  & $\rightarrow$ & $\leftarrow$ & $\rightarrow$ & $\leftarrow$ & $\rightarrow$ \\
                   \midrule
1.5B  & 57.8 &	\textbf{26.0} & 33.7 &	\textbf{31.1} & 54.8 &	\textbf{35.7} & 48.7 &	\textbf{32.7} & 44.9 &	\textbf{29.8}\\
6B & \textbf{62.1} & 	17.9 & \textbf{34.3} & 	25.0  & 55.1 & 	31.9  & \textbf{54.8} & 	28.2 &  \textbf{52.0} & 	25.9 \\
\midrule
\midrule
                       & \multicolumn{2}{c}{en-doi}     & \multicolumn{2}{c}{en-bm}    & \multicolumn{2}{c}{en-ban}     & \multicolumn{2}{c}{en-quc}    & \multicolumn{2}{c}{en-ady}\\
Mono Size          & \multicolumn{2}{c}{179k}     & \multicolumn{2}{c}{187k}      & \multicolumn{2}{c}{188k}     & \multicolumn{2}{c}{250k}    & \multicolumn{2}{c}{296k} \\

            Direction       & $\leftarrow$ & $\rightarrow$ & $\leftarrow$  & $\rightarrow$ & $\leftarrow$  & $\rightarrow$ & $\leftarrow$ & $\rightarrow$ & $\leftarrow$ & $\rightarrow$ \\
      \midrule
1.5B  & 57.4 &	\textbf{29.9} & 27.6 &	28.7 & 32.2 &	\textbf{31.8} & 24.4 &	\textbf{22.8} & 53.7 &	\textbf{21.8} \\
6B &  \textbf{60.4} & 	24.4  & \textbf{33.4} & 	28.5 &  \textbf{35.5} & 	29.4 &  \textbf{27.5} & 	22.2 &  \textbf{56.0} & 	21.4 \\
\midrule
\midrule     & \multicolumn{2}{c}{en-av} 
                      & \multicolumn{2}{c}{en-gom*}     & \multicolumn{2}{c}{en-iso*}    & \multicolumn{2}{c}{en-yua}     & \multicolumn{2}{c}{en-min}      \\
Mono Size           & \multicolumn{2}{c}{301k}  & \multicolumn{2}{c}{311k}     & \multicolumn{2}{c}{409k}      & \multicolumn{2}{c}{419k}     & \multicolumn{2}{c}{533k} \\

            Direction       & $\leftarrow$ & $\rightarrow$ & $\leftarrow$  & $\rightarrow$ & $\leftarrow$  & $\rightarrow$ & $\leftarrow$ & $\rightarrow$ & $\leftarrow$ & $\rightarrow$ \\
         \midrule

1.5B   & 48.1 &	26.1 & 50.7 &	1.1 &
17.9 &	16.5  &34.6 &	30.6 &
58.8 &	51.8 \\
6B &  \textbf{50.5} & 	\textbf{27.3} &  \textbf{53.3} & 	0.9 &  \textbf{19.3} & 	16.8 &  \textbf{39.3} & 	\textbf{31.3} &  \textbf{62.6} & 	\textbf{54.1} \\
\midrule
\midrule     & \multicolumn{2}{c}{en-bho}    & \multicolumn{2}{c}{en-kl*} & \multicolumn{2}{c}{en-qu}    & \multicolumn{2}{c}{en-gn}
                      & \multicolumn{2}{c}{en-bbc}      \\
Mono Size               & \multicolumn{2}{c}{734k}    & \multicolumn{2}{c}{741k}  & \multicolumn{2}{c}{842k}   & \multicolumn{2}{c}{861k} & \multicolumn{2}{c}{923k}    \\

            Direction       & $\leftarrow$ & $\rightarrow$ & $\leftarrow$  & $\rightarrow$ & $\leftarrow$  & $\rightarrow$ & $\leftarrow$ & $\rightarrow$ & $\leftarrow$ & $\rightarrow$ \\
         \midrule

1.5B &
54.8&	38.9&
18.4&	12.5  &
29.9 &	29.8 &
26.7 &	24.3 &
40.9 &	34.9\\
6B&  
\textbf{56.9}&	38.8&
\textbf{20.0}&	\textbf{13.0} & \textbf{32.7} & 	30.2 &  \textbf{35.6} & 	\textbf{28.8} &  \textbf{43.9} & 	\textbf{35.6}  \\
\midrule
\midrule      & \multicolumn{2}{c}{en-skr}    & \multicolumn{2}{c}{en-ak}    & \multicolumn{2}{c}{en-ts*}    & \multicolumn{2}{c}{en-mai}    & \multicolumn{2}{c}{en-ce}
                          \\
Mono Size           & \multicolumn{2}{c}{974k}      & \multicolumn{2}{c}{1.29m}     & \multicolumn{2}{c}{1.3m}    & \multicolumn{2}{c}{1.35m}   & \multicolumn{2}{c}{1.36m}   \\

            Direction       & $\leftarrow$ & $\rightarrow$ & $\leftarrow$  & $\rightarrow$ & $\leftarrow$  & $\rightarrow$ & $\leftarrow$ & $\rightarrow$ & $\leftarrow$ & $\rightarrow$ \\
         \midrule

1.5B   &
44.4 &	\textbf{32.8} &
31.7 &	32.1 &
25.5 &	17.4 &
57.8 &	\textbf{34.1} &
44.6 &	23.5 \\
6B&  \textbf{46.1} & 	28.5 &  \textbf{34.6} & 	\textbf{32.8} &  \textbf{28.3} & 	17.5 &  \textbf{59.9} & 	32.5 &  \textbf{48.1} & 	\textbf{24.4} \\
\midrule
\midrule     & \multicolumn{2}{c}{en-pcm}     & \multicolumn{2}{c}{en-nso}   & \multicolumn{2}{c}{en-ilo}     & \multicolumn{2}{c}{en-cv}    & \multicolumn{2}{c}{en-ti} \\
Mono Size          & \multicolumn{2}{c}{1.59m}     & \multicolumn{2}{c}{1.87m}    & \multicolumn{2}{c}{2.6m}     & \multicolumn{2}{c}{2.9m}    & \multicolumn{2}{c}{3.9m} \\

            Direction       & $\leftarrow$ & $\rightarrow$ & $\leftarrow$  & $\rightarrow$ & $\leftarrow$  & $\rightarrow$ & $\leftarrow$ & $\rightarrow$ & $\leftarrow$ & $\rightarrow$ \\
         \midrule

1.5B   &
\textbf{54.9} &	\textbf{54.7} &
45.8 &	45  &
\textbf{50.8} &	\textbf{49.7} &
39.9 &	\textbf{28.1} &
37.9	 &19.9\\
6B  &  52.5 & 	52.4 &  \textbf{49.1} & 	\textbf{46.9} &  
39.9 & 	47.8 &
\textbf{44.4} & 	27.5 &  
\textbf{40.9} & 	20.0\\
\midrule
\midrule       & \multicolumn{2}{c}{en-om}
                      & \multicolumn{2}{c}{en-sa}  & \multicolumn{2}{c}{en-dv}    & \multicolumn{2}{c}{en-lus*}     & \multicolumn{2}{c}{en-as}    \\
Mono Size              & \multicolumn{2}{c}{5.6m}  & \multicolumn{2}{c}{6.2m}  & \multicolumn{2}{c}{7.9m}     & \multicolumn{2}{c}{8.3m}    & \multicolumn{2}{c}{9.3m}  \\

            Direction       & $\leftarrow$ & $\rightarrow$ & $\leftarrow$  & $\rightarrow$ & $\leftarrow$  & $\rightarrow$ & $\leftarrow$ & $\rightarrow$ & $\leftarrow$ & $\rightarrow$ \\
         \midrule

1.5B &
32.1	 &36 &
43.5 &	\textbf{26.9}   &
\textbf{37} &	\textbf{43.2} &
22.6 &	16.9 &
\textbf{52.9} &	36.3\\
6B &  
\textbf{37.0} & 	\textbf{36.7} &  
\textbf{45.8} & 	24.7&  35.8 & 	39.2 &  \textbf{33.0} & 	\textbf{19.1} &  49.6 & 	36.2   \\
\midrule         & \multicolumn{2}{c}{en-mzn}    & \multicolumn{2}{c}{en-bew}  & \multicolumn{2}{c}{en-ckb*}    & & & \multicolumn{2}{c}{\textbf{Average}} \\

Mono Size       & \multicolumn{2}{c}{11.6m}   & \multicolumn{2}{c}{33.3m}  & \multicolumn{2}{c}{76.9m}    &  & \\

            Direction       & $\leftarrow$ & $\rightarrow$ & $\leftarrow$  & $\rightarrow$ & $\leftarrow$  & $\rightarrow$ && &$\leftarrow$  & $\rightarrow$  \\
         \midrule

1.5B &
50 &	\textbf{41.3} &
49.8 &	\textbf{45.7}  &
46.6 & 40.8\\
6B  &  \textbf{53.3} & 	40.2 &  \textbf{51.6} & 	44.0 &  
\textbf{51.7} & \textbf{41.6}&
&&
\textbf{+2.7} & \textbf{-1.0}\\
\bottomrule
\end{tabular}
\caption{Comparison of our 206-language 1.5B and 6B models trained on supervised data from 112 languages to and from English and monolingual data from 206 languages. Languages marked with * weren't included in the set of 206 languages, so monolingual data was not used for these languages in this experiment.}
\label{tab:mtcapacity}
\end{table}
}

\newcommand{\insertmtmultilinguality}{\begin{table}[t]
\centering
\begin{tabular}{lcccccccccc}
\toprule
                   & \multicolumn{2}{c}{en-mfa}    & \multicolumn{2}{c}{en-ff}     & \multicolumn{2}{c}{en-mni}    & \multicolumn{2}{c}{en-kri}    & \multicolumn{2}{c}{en-mad} \\
Mono Size          & \multicolumn{2}{c}{7k}      & \multicolumn{2}{c}{86k}     & \multicolumn{2}{c}{106k}    & \multicolumn{2}{c}{129k}   & \multicolumn{2}{c}{138k} \\

            Direction       & $\leftarrow$ & $\rightarrow$ & $\leftarrow$  & $\rightarrow$ & $\leftarrow$  & $\rightarrow$ & $\leftarrow$ & $\rightarrow$ & $\leftarrow$ & $\rightarrow$ \\
                   \midrule
200L & 62.1 & 	17.9 & 34.3 & 	25.0  & 55.1 & 	31.9  & 54.8 & 	28.2 &  \textbf{52.0} & 	25.9 \\
1000L&
\textbf{65.4}&	\textbf{27.5}&
\textbf{41.2}&	\textbf{32.3}&
\textbf{56.4}&	\textbf{38.2}&
\textbf{56.5}&	\textbf{34.9}&
50.5&	\textbf{30.2}\\
\midrule
\midrule
                       & \multicolumn{2}{c}{en-doi}     & \multicolumn{2}{c}{en-bm}    & \multicolumn{2}{c}{en-ban}     & \multicolumn{2}{c}{en-quc}    & \multicolumn{2}{c}{en-ady}\\
Mono Size          & \multicolumn{2}{c}{179k}     & \multicolumn{2}{c}{187k}      & \multicolumn{2}{c}{188k}     & \multicolumn{2}{c}{250k}    & \multicolumn{2}{c}{296k} \\

            Direction       & $\leftarrow$ & $\rightarrow$ & $\leftarrow$  & $\rightarrow$ & $\leftarrow$  & $\rightarrow$ & $\leftarrow$ & $\rightarrow$ & $\leftarrow$ & $\rightarrow$ \\
      \midrule
200L &  60.4 & 	24.4  & 33.4 & 	28.5 &  35.5 & 	29.4 &  \textbf{27.5} & 	22.2 & \textbf{56.0} & 	21.4 \\
1000L&
\textbf{63.1}&	\textbf{25.5}&
\textbf{36.3}&	\textbf{34.3}&
35.4&	\textbf{33.1}&
26.7&	\textbf{22.9}&
54.8&	\textbf{28.2}\\
\midrule
\midrule     & \multicolumn{2}{c}{en-av} 
                      & \multicolumn{2}{c}{en-gom*}     & \multicolumn{2}{c}{en-iso*}    & \multicolumn{2}{c}{en-yua}     & \multicolumn{2}{c}{en-min}      \\
Mono Size           & \multicolumn{2}{c}{301k}  & \multicolumn{2}{c}{311k}     & \multicolumn{2}{c}{409k}      & \multicolumn{2}{c}{419k}     & \multicolumn{2}{c}{533k} \\

            Direction       & $\leftarrow$ & $\rightarrow$ & $\leftarrow$  & $\rightarrow$ & $\leftarrow$  & $\rightarrow$ & $\leftarrow$ & $\rightarrow$ & $\leftarrow$ & $\rightarrow$ \\
         \midrule

200L &  \textbf{50.5} & 	27.3 &  53.3 & 	0.9 &  19.3 & 	16.8 &  39.3 & 	31.3 &  62.6 & 	54.1 \\
1000L &
48.1&	\textbf{28.1}&
\textbf{55.5}&	\textbf{39.1}&
\textbf{29.4}&	\textbf{30.5}&
\textbf{40.7}&	31.6&
62.4&	\textbf{56.1}\\
\midrule
\midrule     & \multicolumn{2}{c}{en-bho}    & \multicolumn{2}{c}{en-kl*} & \multicolumn{2}{c}{en-qu}    & \multicolumn{2}{c}{en-gn}
                      & \multicolumn{2}{c}{en-bbc}      \\
Mono Size               & \multicolumn{2}{c}{734k}    & \multicolumn{2}{c}{741k}  & \multicolumn{2}{c}{842k}   & \multicolumn{2}{c}{861k} & \multicolumn{2}{c}{923k}    \\

            Direction       & $\leftarrow$ & $\rightarrow$ & $\leftarrow$  & $\rightarrow$ & $\leftarrow$  & $\rightarrow$ & $\leftarrow$ & $\rightarrow$ & $\leftarrow$ & $\rightarrow$ \\
         \midrule

200L &  
56.9&	38.8&
20.0&	13.0& 32.7 & 	30.2 &  35.6 & 	28.8 &  43.9 & 	35.6 \\
1000L &  
\textbf{58.3}&	\textbf{40.6}&
\textbf{29.7}&	\textbf{23.1}&
32.5&	\textbf{33.1}&
\textbf{38.9}&	\textbf{32.2}&
44.0	&35.4\\
\midrule
\midrule      & \multicolumn{2}{c}{en-skr}    & \multicolumn{2}{c}{en-ak}    & \multicolumn{2}{c}{en-ts*}    & \multicolumn{2}{c}{en-mai}    & \multicolumn{2}{c}{en-ce}
                          \\
Mono Size           & \multicolumn{2}{c}{974k}      & \multicolumn{2}{c}{1.29m}     & \multicolumn{2}{c}{1.3m}    & \multicolumn{2}{c}{1.35m}   & \multicolumn{2}{c}{1.36m}   \\

            Direction       & $\leftarrow$ & $\rightarrow$ & $\leftarrow$  & $\rightarrow$ & $\leftarrow$  & $\rightarrow$ & $\leftarrow$ & $\rightarrow$ & $\leftarrow$ & $\rightarrow$ \\
         \midrule

200L  &  46.1 & 	28.5 &  34.6 & 	32.8&  28.3 & 	17.5 &  59.9 & 	32.5 &  \textbf{48.1} & 	\textbf{24.4} \\
1000L &
\textbf{48.4}&	\textbf{31.3}&
\textbf{36.3}&	\textbf{34.3}&
\textbf{43.0}	&\textbf{45.5}&
\textbf{61.6}&	\textbf{37.6}&
44.9&	23.7\\
\midrule
\midrule     & \multicolumn{2}{c}{en-pcm}     & \multicolumn{2}{c}{en-nso}   & \multicolumn{2}{c}{en-ilo}     & \multicolumn{2}{c}{en-cv}    & \multicolumn{2}{c}{en-ti} \\
Mono Size          & \multicolumn{2}{c}{1.59m}     & \multicolumn{2}{c}{1.87m}    & \multicolumn{2}{c}{2.6m}     & \multicolumn{2}{c}{2.9m}    & \multicolumn{2}{c}{3.9m} \\

            Direction       & $\leftarrow$ & $\rightarrow$ & $\leftarrow$  & $\rightarrow$ & $\leftarrow$  & $\rightarrow$ & $\leftarrow$ & $\rightarrow$ & $\leftarrow$ & $\rightarrow$ \\
         \midrule

200L &  \textbf{52.5} & 	52.4 &  49.1 & 	\textbf{46.9} &
39.9 & 	47.8 &
44.4 & 	27.5 &
40.9 & 	20.0\\
1000L &
51.2&	\textbf{53.5}&
\textbf{51.3}&	41.6&
\textbf{43.4}&	\textbf{52.4}&
\textbf{46.3}&	\textbf{32.1}&
\textbf{44.2}&	\textbf{21.1}\\
\midrule
\midrule       & \multicolumn{2}{c}{en-om}
                      & \multicolumn{2}{c}{en-sa}  & \multicolumn{2}{c}{en-dv}    & \multicolumn{2}{c}{en-lus}     & \multicolumn{2}{c}{en-as}    \\
Mono Size              & \multicolumn{2}{c}{5.6m}  & \multicolumn{2}{c}{6.2m}  & \multicolumn{2}{c}{7.9m}     & \multicolumn{2}{c}{8.3m}    & \multicolumn{2}{c}{9.3m}  \\

            Direction       & $\leftarrow$ & $\rightarrow$ & $\leftarrow$  & $\rightarrow$ & $\leftarrow$  & $\rightarrow$ & $\leftarrow$ & $\rightarrow$ & $\leftarrow$ & $\rightarrow$ \\
         \midrule

200L  &  
37.0 & 	36.7 &
45.8 & 	24.7 &  
35.8 & 	39.2 &  
33.0 & 	19.1 &  
49.6 & 	36.2\\
1000L&
\textbf{38.1}&	\textbf{39.1}&
\textbf{46.3}&	\textbf{28.4}&
\textbf{45.2}&	\textbf{43.7}&
\textbf{34.5}&	\textbf{39.3}&
\textbf{58.6}&	\textbf{36.7}\\
\midrule         & \multicolumn{2}{c}{en-mzn}    & \multicolumn{2}{c}{en-bew}  & \multicolumn{2}{c}{en-ckb*}    & & & \multicolumn{2}{c}{\textbf{Average}} \\

Mono Size       & \multicolumn{2}{c}{11.6m}   & \multicolumn{2}{c}{33.3m}  & \multicolumn{2}{c}{25.1m}    &  & \\

            Direction       & $\leftarrow$ & $\rightarrow$ & $\leftarrow$  & $\rightarrow$ & $\leftarrow$  & $\rightarrow$ && &$\leftarrow$  & $\rightarrow$  \\
         \midrule

200L  & 
53.3 & 	40.2 & 
51.6 & 	44.0 &  
51.7 & 41.6\\
1000L &
\textbf{55.2}&	\textbf{41.4}&
51.8&	\textbf{46.0} &
\textbf{54.4}&	41.6
&&&
\textbf{+2.5} & \textbf{+5.3}\\
\bottomrule
\end{tabular}
\caption{Comparison of our 206-language 6B and 1000-language 6B models trained on monolingual data from 206 languages and 1000 languages respectively, while sharing supervised data from 112 languages. Languages marked with * weren't included in the set of 206 languages, so monolingual data was not used for these languages in the baseline model, and the 1000-language model consequently sees larger performance improvements.}
\label{tab:mtmultilinguality}
\end{table}
}

\newcommand{\insertmtfinetuning}{\begin{table}[t]
\centering
\begin{tabular}{lcccccccccc}
\toprule
                       & \multicolumn{2}{c}{en-ff}     & \multicolumn{2}{c}{en-mni}    & \multicolumn{2}{c}{en-kri}& \multicolumn{2}{c}{en-doi}     & \multicolumn{2}{c}{en-bm}     \\
Mono Size          & \multicolumn{2}{c}{86k}     & \multicolumn{2}{c}{106k}    & \multicolumn{2}{c}{129k} & \multicolumn{2}{c}{179k}     & \multicolumn{2}{c}{187k}   \\

            Direction       & $\leftarrow$ & $\rightarrow$ & $\leftarrow$  & $\rightarrow$ & $\leftarrow$  & $\rightarrow$ & $\leftarrow$ & $\rightarrow$ & $\leftarrow$ & $\rightarrow$ \\
                   \midrule
1000L&
41.2&	32.3&
56.4&	38.2&
56.5&	34.9&
63.1&	25.5&
36.3&	34.3\\
Finetuned&
\textbf{44.5}&	\textbf{35.7}&
\textbf{60.5}&	\textbf{40.8}&
\textbf{62.2}&	\textbf{36.8}&
\textbf{64.6}&	\textbf{37.2}&
\textbf{37.9}&	34.7\\
\midrule
\midrule
                            & \multicolumn{2}{c}{en-quc}
                      & \multicolumn{2}{c}{en-gom}     & \multicolumn{2}{c}{en-yua}& \multicolumn{2}{c}{en-bho}    & \multicolumn{2}{c}{en-kl}\\
Mono Size             & \multicolumn{2}{c}{250k}     &  \multicolumn{2}{c}{311k}     &  \multicolumn{2}{c}{419k}  & \multicolumn{2}{c}{734k}    & \multicolumn{2}{c}{741k} \\

            Direction       & $\leftarrow$ & $\rightarrow$ & $\leftarrow$  & $\rightarrow$ & $\leftarrow$  & $\rightarrow$ & $\leftarrow$ & $\rightarrow$ & $\leftarrow$ & $\rightarrow$ \\
      \midrule
1000L&
26.7&	22.9 &
55.5&	39.1&
40.7&	31.6 &  
58.3&	40.6&
29.7&	23.1\\
Finetuned&
\textbf{29.2}&	\textbf{23.5}&
\textbf{57.5}&	\textbf{40.0}&
\textbf{43.0}&   31.8 &  
\textbf{60.5}&	\textbf{41.4}&
\textbf{39.1}&	\textbf{27.8}\\
\midrule
\midrule
                          & \multicolumn{2}{c}{en-qu}    & \multicolumn{2}{c}{en-gn} & \multicolumn{2}{c}{en-ak} & \multicolumn{2}{c}{en-ts}    & \multicolumn{2}{c}{en-mai}      \\
Mono Size            & \multicolumn{2}{c}{842k}   & \multicolumn{2}{c}{861k}    & \multicolumn{2}{c}{1.29m}& \multicolumn{2}{c}{1.3m}    & \multicolumn{2}{c}{1.35m}   \\

            Direction       & $\leftarrow$ & $\rightarrow$ & $\leftarrow$  & $\rightarrow$ & $\leftarrow$  & $\rightarrow$ & $\leftarrow$ & $\rightarrow$ & $\leftarrow$ & $\rightarrow$ \\
         \midrule

1000L&
32.5&	33.1&
38.9&	32.2&
36.3&	34.3&
43.0	&45.5&
61.6&	37.6\\
Finetuned&
\textbf{35.3}&	\textbf{36.1}&
\textbf{42.7}&	32.3&
\textbf{38.6}&	34.1&
\textbf{46.2}&	\textbf{46.5}&
\textbf{64.3}&	\textbf{39.2}\\
\midrule
\midrule       & \multicolumn{2}{c}{en-pcm}     & \multicolumn{2}{c}{en-nso}    & \multicolumn{2}{c}{en-ilo}     & \multicolumn{2}{c}{en-ti}    & \multicolumn{2}{c}{en-om} \\
Mono Size           & \multicolumn{2}{c}{1.59m}     & \multicolumn{2}{c}{1.87m}      & \multicolumn{2}{c}{2.6m}     & \multicolumn{2}{c}{3.9m}   & \multicolumn{2}{c}{5.6m} \\
            Direction       & $\leftarrow$ & $\rightarrow$ & $\leftarrow$  & $\rightarrow$ & $\leftarrow$  & $\rightarrow$ & $\leftarrow$ & $\rightarrow$ & $\leftarrow$ & $\rightarrow$ \\
         \midrule

1000L &
51.2&	53.5&
51.3&	41.6&
43.4&	52.4&
44.2&	21.1&
38.1&	39.1\\
Finetuned&
\textbf{59.3}&	\textbf{54.5}&
\textbf{52.2}&	\textbf{45.4}&
\textbf{61.7}&	\textbf{53.4}&
\textbf{46.0}&	21.0&
\textbf{41.2}&	\textbf{39.9}\\
\midrule
\midrule   
                      & \multicolumn{2}{c}{en-sa}       & \multicolumn{2}{c}{en-dv}   & \multicolumn{2}{c}{en-lus}     & \multicolumn{2}{c}{en-as}  & \multicolumn{2}{c}{en-ckb} \\
Mono Size            & \multicolumn{2}{c}{6.2m}    & \multicolumn{2}{c}{7.9m}  & \multicolumn{2}{c}{8.3m}    & \multicolumn{2}{c}{9.3m}  & \multicolumn{2}{c}{25.1m}\\

            Direction       & $\leftarrow$ & $\rightarrow$ & $\leftarrow$  & $\rightarrow$ & $\leftarrow$  & $\rightarrow$ & $\leftarrow$ & $\rightarrow$ & $\leftarrow$ & $\rightarrow$ \\
         \midrule

1000L&
46.3&	28.4 &
45.2&	43.7&
34.5&	39.3&
58.6&	36.7&
54.4&	41.6\\
Finetuned&
\textbf{49.0}	&\textbf{30.9}&
\textbf{47.3}&	43.8&
\textbf{40.0}	&\textbf{40.3}&
\textbf{59.8}&	\textbf{39.0} &  
54.7&	\textbf{42.7}\\
\bottomrule
\end{tabular}
\caption{Comparison of our 1000-language 6B models against the 30-language version fine-tuned from this model. Scores are shown for the languages that were fine-tuned on.}
\label{tab:mtfinetuning}
\end{table}
}

\newcommand{\insertmtfiltering}{\begin{table}[t]
\centering
\begin{tabular}{lcccccccccc}
\toprule
                       & \multicolumn{2}{c}{en-ff}     & \multicolumn{2}{c}{en-mni}    & \multicolumn{2}{c}{en-kri}& \multicolumn{2}{c}{en-doi}     & \multicolumn{2}{c}{en-bm}     \\
Mono Size          & \multicolumn{2}{c}{86k}     & \multicolumn{2}{c}{106k}    & \multicolumn{2}{c}{129k} & \multicolumn{2}{c}{179k}     & \multicolumn{2}{c}{187k}   \\

            Direction       & $\leftarrow$ & $\rightarrow$ & $\leftarrow$  & $\rightarrow$ & $\leftarrow$  & $\rightarrow$ & $\leftarrow$ & $\rightarrow$ & $\leftarrow$ & $\rightarrow$ \\
                   \midrule
Finetuned&
44.5&	35.7&
60.5&	40.8&
62.2&	\textbf{36.8}&
64.6&	37.2&
37.9&	34.7\\
Filtered&
\textbf{45.3}&	35.3&
\textbf{62.1}&	40.8&
\textbf{64.2}&	35.4&
\textbf{65.5}&	36.9&
\textbf{38.4}&	34.7\\
\midrule
\midrule
                            & \multicolumn{2}{c}{en-quc}
                      & \multicolumn{2}{c}{en-gom}     & \multicolumn{2}{c}{en-yua} & \multicolumn{2}{c}{en-bho}    & \multicolumn{2}{c}{en-kl}\\
Mono Size             & \multicolumn{2}{c}{250k}     &  \multicolumn{2}{c}{311k}     &  \multicolumn{2}{c}{419k}  & \multicolumn{2}{c}{734k}    & \multicolumn{2}{c}{741k}  \\

            Direction       & $\leftarrow$ & $\rightarrow$ & $\leftarrow$  & $\rightarrow$ & $\leftarrow$  & $\rightarrow$ & $\leftarrow$ & $\rightarrow$ & $\leftarrow$ & $\rightarrow$ \\
      \midrule
Finetuned&
29.2&	23.5&
57.5&	40&
43	&31.8 &  
60.5&	41.4&
39.1&	27.8\\
Filtered&
\textbf{29.8}&	23.5&
57.9&	\textbf{40.8}&
\textbf{43.7}&	32.0 &  
60.9&	41.2&
39.0&	\textbf{35.8}\\

\midrule
\midrule
                        &   \multicolumn{2}{c}{en-qu}    & \multicolumn{2}{c}{en-gn}& \multicolumn{2}{c}{en-ak} & \multicolumn{2}{c}{en-ts}    & \multicolumn{2}{c}{en-mai}        \\
Mono Size            & \multicolumn{2}{c}{842k}   & \multicolumn{2}{c}{861k}    & \multicolumn{2}{c}{1.29m}& \multicolumn{2}{c}{1.3m}    & \multicolumn{2}{c}{1.35m}   \\

            Direction       & $\leftarrow$ & $\rightarrow$ & $\leftarrow$  & $\rightarrow$ & $\leftarrow$  & $\rightarrow$ & $\leftarrow$ & $\rightarrow$ & $\leftarrow$ & $\rightarrow$ \\
         \midrule

Finetuned&
35.3&	36.1&
42.7&	32.3&
38.6&	34.1&
46.2&	46.5&
64.3&	39.2\\
Filtered&
35.3&	35.7&
43.0&	32.1&
38.6&	34.3&
\textbf{47.5}&	46.5&
\textbf{65.5}&	39.2\\
\midrule
\midrule      & \multicolumn{2}{c}{en-pcm}     & \multicolumn{2}{c}{en-nso}  & \multicolumn{2}{c}{en-ilo}     & \multicolumn{2}{c}{en-ti}    & \multicolumn{2}{c}{en-om} \\
Mono Size              & \multicolumn{2}{c}{1.59m}     & \multicolumn{2}{c}{1.87m}  & \multicolumn{2}{c}{2.6m}     & \multicolumn{2}{c}{3.9m}   & \multicolumn{2}{c}{5.6m} \\
            Direction       & $\leftarrow$ & $\rightarrow$ & $\leftarrow$  & $\rightarrow$ & $\leftarrow$  & $\rightarrow$ & $\leftarrow$ & $\rightarrow$ & $\leftarrow$ & $\rightarrow$ \\
         \midrule

Finetuned&
59.3&	54.5&
52.2&	45.4&
61.7&	53.4&
46.0&	21.0&
\textbf{41.2}&	39.9\\
Filtered &
\textbf{64.6}&	\textbf{57.2}&
52.5&	\textbf{47.1}&
\textbf{62.7}&	\textbf{54.1}&
46.1&	21.5&
40.7&	40.0\\
\midrule
\midrule   
                      & \multicolumn{2}{c}{en-sa}       & \multicolumn{2}{c}{en-dv}   & \multicolumn{2}{c}{en-lus}     & \multicolumn{2}{c}{en-as}  & \multicolumn{2}{c}{en-ckb} \\
Mono Size            & \multicolumn{2}{c}{6.2m}    & \multicolumn{2}{c}{7.9m}  & \multicolumn{2}{c}{8.3m}    & \multicolumn{2}{c}{9.3m}  & \multicolumn{2}{c}{25.1m}\\
         \midrule

Finetuned&
49.0	&30.9&
47.3&	43.8&
40.0	&\textbf{40.3}&
59.8&	39.0 &  
54.7&	42.7\\
Filtered&
49.2&	31.0&
\textbf{48.4}&	\textbf{45.0}&
\textbf{42.1}&	39.7&
\textbf{60.6}&	\textbf{39.6} &  
\textbf{55.3}&	42.6\\
\bottomrule
\end{tabular}
\caption{Comparison of our 30-language fine-tuned model against one fine-tuned with RTT and LangID filtered data.}
\label{tab:mtfiltering}
\end{table}
}

\newcommand{\insertdistilledperf}{\begin{table}[t]

\centering
\small
\begin{tabular}{l|r:rrrrrrr:r:r}
LP & avg & 0 & 1 & 2 & 3 & 4 & 5 & 6 & entropy & \chrf{} \\
\hline

en$\rightarrow$ti &     5.4 &   0 &     0 &     1 &     4 &     11 &    18 &    66 &    0.45 &  22.0 \\ 
en$\rightarrow$ay &     5.1 &   0 &     0 &     2 &     5 &     17 &    36 &    40 &    0.56 &  33.3 \\ 
en$\rightarrow$bm &     5.0 &   0 &     1 &     1 &     3 &     22 &    40 &    34 &    0.55 &  36.6 \\ 
en$\rightarrow$ts &     4.9 &   1 &     0 &     2 &     6 &     24 &    32 &    37 &    0.59 &  46.5 \\ 
en$\rightarrow$lus &    4.9 &   3 &     0 &     5 &     0 &     33 &    13 &    47 &    0.55 &  41.1 \\ 
en$\rightarrow$ilo &    4.7 &   0 &     0 &     3 &     2 &     25 &    64 &    7 &     0.44 &  54.8 \\ 
en$\rightarrow$ff &     4.6 &   1 &     1 &     4 &     7 &     24 &    45 &    18 &    0.61 &  37.2 \\ 
en$\rightarrow$ln &     4.6 &   0 &     0 &     3 &     10 &    28 &    41 &    18 &    0.61 &  34.9 \\ 
en$\rightarrow$doi &    4.5 &   0 &     0 &     2 &     3 &     38 &    50 &    6 &     0.49 &  40.9 \\ 
en$\rightarrow$kri &    4.4 &   0 &     2 &     9 &     14 &    26 &    23 &    27 &    0.71 &  36.6 \\ 
en$\rightarrow$om &     4.4 &   0 &     0 &     3 &     10 &    42 &    37 &    8 &     0.56 &  40.3 \\ 
en$\rightarrow$dv &     4.1 &   2 &     2 &     8 &     12 &    33 &    29 &    14 &    0.71 &  45.6 \\ 
en$\rightarrow$sa &     4.1 &   4 &     1 &     3 &     7 &     52 &    30 &    4 &     0.57 &  33.3 \\ 
en$\rightarrow$as &     4.0 &   0 &     0 &     4 &     12 &    60 &    23 &    0 &     0.47 &  40.8 \\ 
en$\rightarrow$gom &    4.0 &   4 &     1 &     14 &    11 &    38 &    12 &    21 &    0.71 &  41.5 \\ 
en$\rightarrow$lg &     3.9 &   0 &     8 &     11 &    18 &    22 &    25 &    16 &    0.76 &  39.7 \\ 
en$\rightarrow$ak &     3.8 &   0 &     0 &     19 &    0 &     69 &    0 &     12 &    0.39 &  34.6 \\ 
en$\rightarrow$ckb &    3.6 &   1 &     2 &     14 &    32 &    30 &    13 &    9 &     0.70 &  44.3 \\ 
en$\rightarrow$nso &    3.4 &   7 &     13 &    13 &    16 &    19 &    19 &    14 &    0.83 &  47.6 \\ 
en$\rightarrow$kl &     3.2 &   13 &    3 &     19 &    13 &    29 &    10 &    13 &    0.79 &  40.7 \\ 
en$\rightarrow$mni-Mtei &       3.2 &   0 &     0 &     24 &    32 &    43 &    1 &     0 &     0.51 &  47.7 \\ 
en$\rightarrow$gn &     3.1 &   0 &     8 &     30 &    18 &    35 &    8 &     2 &     0.66 &  32.2 \\ 
en$\rightarrow$qu &     3.1 &   4 &     4 &     20 &    30 &    36 &    6 &     0 &     0.65 &  37.2 \\ 
en$\rightarrow$ee &     3.0 &   7 &     14 &    18 &    21 &    19 &    12 &    9 &     0.82 &  39.9 \\ 
en$\rightarrow$mai &    3.0 &   8 &     1 &     30 &    20 &    30 &    8 &     4 &     0.71 &  40.2 \\ 
en$\rightarrow$pcm &    2.7 &   20 &    9 &     22 &    4 &     27 &    10 &    9 &     0.78 &  57.5 \\ 
en$\rightarrow$bho &    2.3 &   0 &     6 &     70 &    14 &    8 &     1 &     0 &     0.44 &  42.4 \\ 
en$\rightarrow$yua &    2.1 &   0 &     30 &    42 &    17 &    8 &     2 &     1 &     0.59 &  32.8 \\ 
\hdashline                                                                                      
ti$\rightarrow$en &     4.6 &   0 &     0 &     3 &     9 &     31 &    37 &    21 &    0.62 &  45.8 \\ 
ay$\rightarrow$en &     4.8 &   2 &     1 &     5 &     6 &     12 &    48 &    27 &    0.61 &  38.8 \\ 
ln$\rightarrow$en &     4.8 &   0 &     1 &     1 &     6 &     25 &    46 &    22 &    0.58 &  32.3 \\ 
lg$\rightarrow$en &     5.6 &   0 &     1 &     2 &     3 &     4 &     8 &     82 &    0.34 &  41.1 \\ 
mni-Mtei$\rightarrow$en &       3.5 &   0 &     1 &     20 &    14 &    65 &    1 &     0 &     0.44 &  62.9 \\ 
bm$\rightarrow$en &     5.1 &   0 &     0 &     0 &     4 &     12 &    52 &    32 &    0.50 &  38.7 \\ 
ts$\rightarrow$en &     4.3 &   1 &     1 &     8 &     14 &    27 &    33 &    17 &    0.68 &  47.5 \\ 
lus$\rightarrow$en &    4.6 &   4 &     4 &     5 &     4 &     13 &    40 &    31 &    0.66 &  41.7 \\ 
ilo$\rightarrow$en &    4.6 &   0 &     0 &     3 &     3 &     30 &    59 &    6 &     0.47 &  62.4 \\ 
ff$\rightarrow$en &     4.6 &   7 &     1 &     10 &    4 &     13 &    11 &    54 &    0.63 &  45.8 \\ 
doi$\rightarrow$en &    4.6 &   1 &     0 &     1 &     4 &     41 &    33 &    20 &    0.56 &  65.8 \\ 
kri$\rightarrow$en &    5.0 &   0 &     1 &     6 &     8 &     13 &    18 &    53 &    0.59 &  64.8 \\ 
om$\rightarrow$en &     4.6 &   0 &     0 &     1 &     7 &     37 &    50 &    7 &     0.50 &  41.9 \\ 
dv$\rightarrow$en &     3.4 &   0 &     2 &     19 &    33 &    33 &    14 &    0 &     0.61 &  48.7 \\ 
sa$\rightarrow$en &     4.4 &   0 &     1 &     1 &     4 &     55 &    33 &    6 &     0.48 &  48.9 \\ 
as$\rightarrow$en &     5.2 &   0 &     2 &     8 &     2 &     10 &    15 &    64 &    0.51 &  60.4 \\ 
gom$\rightarrow$en &    5.5 &   0 &     0 &     2 &     1 &     5 &     30 &    62 &    0.43 &  57.2 \\ 
ak$\rightarrow$en &     4.8 &   0 &     1 &     4 &     9 &     19 &    31 &    37 &    0.64 &  39.4 \\ 
ckb$\rightarrow$en &    2.9 &   4 &     12 &    27 &    14 &    36 &    7 &     0 &     0.69 &  56.4 \\ 
nso$\rightarrow$en &    3.4 &   2 &     12 &    14 &    18 &    34 &    9 &     12 &    0.76 &  52.8 \\ 
kl$\rightarrow$en &     4.6 &   2 &     3 &     11 &    5 &     25 &    9 &     46 &    0.65 &  39.6 \\ 
gn$\rightarrow$en &     3.5 &   0 &     0 &     12 &    31 &    49 &    7 &     1 &     0.54 &  43.6 \\ 
qu$\rightarrow$en &     2.9 &   0 &     4 &     28 &    47 &    18 &    5 &     0 &     0.57 &  35.6 \\ 
ee$\rightarrow$en &     4.5 &   3 &     3 &     8 &     12 &    13 &    22 &    39 &    0.71 &  37.3 \\ 
mai$\rightarrow$en &    3.2 &   0 &     2 &     38 &    15 &    25 &    19 &    0 &     0.62 &  65.4 \\ 
pcm$\rightarrow$en &    4.6 &   1 &     0 &     3 &     7 &     31 &    39 &    20 &    0.61 &  65.4 \\ 
bho$\rightarrow$en &    3.4 &   0 &     0 &     15 &    44 &    31 &    10 &    0 &     0.56 &  61.7 \\ 
yua$\rightarrow$en &    3.4 &   0 &     1 &     20 &    36 &    29 &    13 &    2 &     0.63 &  42.1 \\ 
\bottomrule
\end{tabular}
\caption{Performance of the distilled student model in \chrf{} and human evaluation, on a scale from 0 (nonsense/wrong language) to 6 (perfect). The value in the ``avg'' column reports the weighted average score across all sentences. The value under each of the numbers from 0 to 6 is the percent of sentences that were given that rating. The entropy is included to flag suspicious rating patterns: a low entropy may mean that most raters are assigning the same score to all sentences.}
\label{tab:distilledperf}
\end{table}
}

\newcommand{\insertsinglewordserrorcats}{
\begin{table}[H]
\centering
\small
\begin{tabular}{ll| lllll}
\hline
\hline
direction & 	model & 	copies & 	multi-word & 	duplicate & 	repeats & 	other \\

\hline
\enxx & 	teacher & 	1.28 & 	66.23 & 	26.48 & 	5.55 & 	17.43 \\ 
\enxx & 	teacher. & 	0.85 & 	71.71 & 	31.66 & 	2.79 & 	13.87 \\ 
\hdashline
\enxx & 	student & 	1.15 & 	48.21 & 	37.16 & 	1.43 & 	27.38 \\ 
\enxx & 	student. & 	8.28 & 	45.53 & 	36.18 & 	2.05 & 	22.73 \\
\hline
\xxen & 	teacher & 	61.87 & 	6.19 & 	21.11 & 	0.74 & 	14.74 \\
\xxen & 	teacher. & 	47.3 & 	17.02 & 	24.8 & 	0.47 & 	18.21 \\
\hdashline
\xxen & 	student & 	18.33 & 	28.54 & 	43.23 & 	2.65 & 	22.65 \\
\xxen & 	student. & 	20.1 & 	35.14 & 	40.41 & 	0.27 & 	21.70 \\

\hline
\end{tabular}
\caption{Statistics for single-token translation on the top ten thousand most common tokens from the monolingual datasets, compared for 48 language pairs, for the fine-tuned teacher model. Metrics shown are 1) percent of outputs that copy the source; 2) percent of input that is multi-word; 3) percent of outputs that are identical to other outputs, suggesting incorrect translation; 4) percent of outputs with low character diversity, suggesting repeats; and 5) percent of outputs with none of those features. The entries with full-stops (.) after them use the ``period trick'' (Section \ref{sec:period}). Single word translation is a particularly tricky task as the model needs to solve translation and LangID simultaneously, which can be undefined for short queries. 
}
\label{tab:singlewordserrorcats}
\end{table}
}

\newcommand{\inserterrorssinglewordsshort}{
\begin{table}[H]
\centering
\small
\begin{tabular}{ll|lll}
\hline
\hline
sl    & tl & source     & translation \\
\hline
en & ff & devices & ka\AS{\m{b}}ir\AS{\m{d}}e (dispositifs) \\
en & lus &  fragile & hring hring (fragile) \\
en & pcm & removes & \textbf{dey} remove  \\
en & pcm & blame  & blame wetin?  \\
en & lus & four   & pali \textbf{a awm a}  \\
en & lus & freedom& zalenna \textbf{a awm}  \\
en & quc & solved & xsol rij \textbf{ri}  \\
en & quc & dam  & \textbf{ri} k'o pa \textbf{ri} cho  \\
en & kri & juvenile & \AS{pikin we n\m{o} rich 18 ia yet} \\
en & kri & notorious & \AS{pipul d\m{e}n we g\m{e}t badnem} \\
\hline
\end{tabular}

\caption{Our models frequently displayed exceeding verbosity with single-word inputs. In some cases they would add extra definitions for words in parentheses or commas afterwards (top half of table). In other instances they added function words from that language after the translation (\textbf{boldface}), or gave whole definitions, as with Krio.}
\label{tab:errors_singlewords_short}
\end{table}
}

\newcommand{\insertinflecting}{
\begin{table*}
\centering
\fontsize{8.5pt}{9pt}\selectfont
\begin{tabular}{p{0.42\linewidth} | p{0.52\linewidth}}
\hline
source     & translation \\
\hline
I can't sing so well so I don't want to be a singer. & Erinangippallaannginnama erinarsortartunngorusunngilanga.\\
Without my car, I wouldn't be able to get to work.	& Biileqanngikkuma suliartorsinnaanavianngilanga.\\
I would like to raise a dog instead of a cat	 & Qimmiuteqarusunnerussangaluarpunga qitsuuteqarnissannit.\\
I have a cat who is lazy. &	Qitsuuteqarpunga eqiasuttuuvoq. \\
\hline
\end{tabular}

\caption{Example translations from English to Kalaallisut from our evaluation set. A token-based metric like \bleu{} is unsuitable for such a highly inflecting language, as exact match is very unlikely. Character-based metrics, like \chrf{}, are more appropriate. }
\label{tab:inflecting}
\end{table*}
}

\newcommand{\insertbleuvschrf}{
\begin{table*}
\centering
\small
\begin{tabular}{lrrr||lrrr}
\hline
\hline
lp &	\bleu{} & 	\chrf{} & 	ratio & lp &	\bleu{} & 	\chrf{} & 	ratio \\ 
\hline

en$\rightarrow$om &     3.8 &   40.1 &  4.0  & en$\rightarrow$ts &      12.8 &  46.7 &  1.6 \\ 
en$\rightarrow$lg &     3.7 &   39.8 &  4.0  & en$\rightarrow$gom &     10.8 &  42.3 &  1.6 \\ 
en$\rightarrow$dv &     5.4 &   45.5 &  3.5  & en$\rightarrow$ee &      11.9 &  40.5 &  1.3 \\ 
en$\rightarrow$sa &     2.8 &   33.1 &  3.5  & en$\rightarrow$ff &      9.9 &   37.2 &  1.3 \\ 
en$\rightarrow$qu &     3.6 &   36.2 &  3.4  & en$\rightarrow$bm &      9.5 &   36.4 &  1.3 \\ 
en$\rightarrow$ln &     3.5 &   34.9 &  3.2  & en$\rightarrow$lus &     13.2 &  41.5 &  1.2 \\ 
en$\rightarrow$kl &     5.3 &   40.6 &  2.9  & en$\rightarrow$kri &     10.6 &  36.2 &  1.1 \\ 
en$\rightarrow$ay &     4.6 &   34.3 &  2.4  & en$\rightarrow$ilo &     24.2 &  54.9 &  1.1 \\ 
en$\rightarrow$ckb &    9.4 &   44.2 &  1.9  & en$\rightarrow$ak &      10.4 &  34.6 &  1.0 \\ 
en$\rightarrow$gn &     5.2 &   32.1 &  1.7  & en$\rightarrow$nso &     20.3 &  47.6 &  1.0 \\ 
en$\rightarrow$mai &    8.6 &   39.8 &  1.7  & en$\rightarrow$bho &     16.7 &  42.7 &  1.0 \\ 
en$\rightarrow$as &     9.6 &   40.6 &  1.6  & en$\rightarrow$doi &     15.3 &  40.7 &  1.0 \\ 
en$\rightarrow$mni-Mtei &       12.9 &  47.1 &  1.6  & en$\rightarrow$ti &      3.7 &   22.1 &  0.4 \\ 

\hline
\end{tabular}

\caption{\chrf{} versus \bleu{} scores on \enxx language pairs in the distilled model. Although it is almost impossible to compare these scores directly, as explained for a variety of reasons in the text, it is still clear that they give a very different picture of the performance on the languages. In order to make an approximate comparison between the two, we have included the ratio of the \scaledchrf{} score to the \bleu{} score. Language pairs in this table are sorted from the highest ratio (where \bleu{} underestimates performance) on the top to the lowest ratio on the bottom. As expected, polysynthetic and agglutinative languages are misjudged my \bleu{}.}
\label{tab:chrfvsbleu}
\end{table*}
}

\newcommand{\inserttabledatasizesummary}{
\begin{table*}
\centering
\begin{tabular}{l|rrrrr}
 
 dataset & N & total & median &  $>$ 1M &  $>$ 100K \\
\hline
LRL-full & 1503 & 1.7B & 25k & 122 & 322 \\
LRL-train  & 1057 & 1.7B & 38k & 122 & 321 \\
all-train  & 1140 & 28B & 43k & 205 & 404 \\
\end{tabular}
\caption{Summary statistics about monolingual data for 1) the full dataset of low-resource languages; 2) the portion thereof used to train our model; and 3) the full training set including high resource languages. Columns are the number of languages covered, the total number of sentences across all languages, the median number of sentences per language, the number of languages with more than 1 million sentences, and the number of languages with more than 100,000 sentences.
 \label{tab:datasize_summary}}
\end{table*}}
\newcommand{\inserttablegroundingtokens}{\begin{table*}
\centering
\scriptsize
\begin{tabular}{lllllllllllllll}
\hline
\hline
sents&	pts&	lb&	ub&	metric &	bew &	mzn &	as &	lus &	dv &	sa &	om &	ti &	cv &	ilo \\
\hline										
&       &       &       &       N mono &   33M &   12M &   9M &    8M &    8M &    6M &    6M &    4M &    3M &    3M \\
&       &       &       &       \chrf &   51.5 &  54.6 &  59.1 &  34.5 &  45.1 &  46.6 &  38 &    43.8 &  46.6 &  43.5 \\
\hline
1191&   9682&   0&      125&    $\Xi$ &       \cellcolor{scale8} 64 & \cellcolor{scale8} 65 & \cellcolor{scale9} 73 & \cellcolor{scale5} 43 & \cellcolor{scale7} 58 & \cellcolor{scale8} 64 & \cellcolor{scale7} 55 & \cellcolor{scale8} 60 & \cellcolor{scale8} 62 & \cellcolor{scale5} 41 \\
994&    2537&   125&    500&    $\Xi$ &       \cellcolor{scale7} 56 & \cellcolor{scale7} 60 & \cellcolor{scale8} 65 & \cellcolor{scale4} 32 & \cellcolor{scale6} 49 & \cellcolor{scale6} 51 & \cellcolor{scale5} 42 & \cellcolor{scale6} 52 & \cellcolor{scale6} 51 & \cellcolor{scale5} 39 \\
1034&   3102&   500&    2k&     $\Xi$ &       \cellcolor{scale7} 55 & \cellcolor{scale8} 61 & \cellcolor{scale8} 66 & \cellcolor{scale4} 31 & \cellcolor{scale6} 48 & \cellcolor{scale6} 50 & \cellcolor{scale5} 39 & \cellcolor{scale6} 45 & \cellcolor{scale6} 49 & \cellcolor{scale5} 42 \\
958&    2420&   2k&     8k&     $\Xi$ &       \cellcolor{scale6} 51 & \cellcolor{scale7} 56 & \cellcolor{scale7} 56 & \cellcolor{scale3} 27 & \cellcolor{scale6} 44 & \cellcolor{scale5} 42 & \cellcolor{scale4} 28 & \cellcolor{scale5} 38 & \cellcolor{scale5} 43 & \cellcolor{scale5} 41 \\

\multicolumn{15}{c}{}\\
\hline
\hline
metric &        nso &   pcm &   ce &    mai &   ts &    ak &    skr &   bbc &   gn &    qu &    min &   yua &   iso &   gom \\
\hline

N mono &   2M &    2M &    1M &    1M &    1M &    1M &    974k &  932k &  861k &  842k &  533k &  419k &  409k &  311k \\
\chrf &   51.1 &  52.3 &  45.2 &  61.4 &  43.6 &  36.1 &  47.8 &  44.2 &  38.2 &  32.2 &  62.2 &  40.8 &  29.1 &  54.9 \\

\hline
$\Xi$ &       \cellcolor{scale9} 73 & \cellcolor{scale8} 61 & \cellcolor{scale7} 55 & \cellcolor{scale9} 68 & \cellcolor{scale8} 61 & \cellcolor{scale7} 57 & \cellcolor{scale8} 64 & \cellcolor{scale7} 56 & \cellcolor{scale7} 57 & \cellcolor{scale5} 40 & \cellcolor{scale9} 72 & \cellcolor{scale7} 58 & \cellcolor{scale6} 45 & \cellcolor{scale9} 69 \\
$\Xi$ &       \cellcolor{scale7} 57 & \cellcolor{scale7} 53 & \cellcolor{scale6} 47 & \cellcolor{scale8} 66 & \cellcolor{scale6} 45 & \cellcolor{scale4} 34 & \cellcolor{scale7} 53 & \cellcolor{scale6} 47 & \cellcolor{scale4} 35 & \cellcolor{scale4} 28 & \cellcolor{scale8} 66 & \cellcolor{scale5} 40 & \cellcolor{scale3} 23 & \cellcolor{scale7} 59 \\
$\Xi$ &       \cellcolor{scale6} 51 & \cellcolor{scale7} 54 & \cellcolor{scale6} 45 & \cellcolor{scale8} 67 & \cellcolor{scale5} 42 & \cellcolor{scale4} 30 & \cellcolor{scale7} 53 & \cellcolor{scale6} 46 & \cellcolor{scale4} 32 & \cellcolor{scale4} 30 & \cellcolor{scale8} 67 & \cellcolor{scale5} 39 & \cellcolor{scale2} 20 & \cellcolor{scale7} 59 \\
$\Xi$ &       \cellcolor{scale5} 41 & \cellcolor{scale7} 53 & \cellcolor{scale5} 43 & \cellcolor{scale8} 62 & \cellcolor{scale5} 38 & \cellcolor{scale3} 24 & \cellcolor{scale6} 47 & \cellcolor{scale5} 39 & \cellcolor{scale4} 28 & \cellcolor{scale4} 29 & \cellcolor{scale8} 63 & \cellcolor{scale4} 31 & \cellcolor{scale2} 17 & \cellcolor{scale6} 52 \\

\multicolumn{15}{c}{}\\
\hline
\hline
metric &        av &    ady &   quc &   ban &   bm &    doi &   mad &   kri &   mni &   ff &    mfa \\
\hline

N mono &   301k &  296k &  250k &  188k &  187k &  179k &  138k &  129k &  106k &  86k &   7k \\
\chrf &   48.4 &  55.2 &  27 &    35.4 &  35.8 &  62.8 &  50.1 &  56.1 &  56.2 &  40.7 &  65.1 \\
\hline
$\Xi$ &       \cellcolor{scale7} 55 & \cellcolor{scale7} 56 & \cellcolor{scale5} 42 & \cellcolor{scale6} 48 & \cellcolor{scale7} 57 & \cellcolor{scale10} 77 &        \cellcolor{scale8} 61 & \cellcolor{scale8} 65 & \cellcolor{scale9} 71 & \cellcolor{scale8} 66 & \cellcolor{scale10} 77 \\
$\Xi$ &       \cellcolor{scale7} 53 & \cellcolor{scale8} 63 & \cellcolor{scale3} 20 & \cellcolor{scale4} 33 & \cellcolor{scale4} 34 & \cellcolor{scale9} 68 & \cellcolor{scale6} 50 & \cellcolor{scale7} 58 & \cellcolor{scale8} 61 & \cellcolor{scale5} 41 & \cellcolor{scale9} 71 \\
$\Xi$ &       \cellcolor{scale7} 57 & \cellcolor{scale8} 67 & \cellcolor{scale2} 19 & \cellcolor{scale4} 33 & \cellcolor{scale4} 30 & \cellcolor{scale9} 69 & \cellcolor{scale6} 49 & \cellcolor{scale7} 59 & \cellcolor{scale7} 59 & \cellcolor{scale4} 33 & \cellcolor{scale9} 69 \\
$\Xi$ &       \cellcolor{scale7} 55 & \cellcolor{scale8} 65 & \cellcolor{scale2} 16 & \cellcolor{scale4} 29 & \cellcolor{scale3} 22 & \cellcolor{scale8} 62 & \cellcolor{scale6} 45 & \cellcolor{scale6} 50 & \cellcolor{scale6} 51 & \cellcolor{scale3} 26 & \cellcolor{scale8} 62 \\

\end{tabular}
\caption{Token hit-rate ($\Xi$) for different sets of tokens, binned by frequency. The lower and upper bound of the token frequency rank per bin are given in the \texttt{lb} and \texttt{ub} columns; thus, the top row is the hit-rate on the 125 most frequent tokens, and the bottom row is on the 6,000 least frequent tokens. Columns \texttt{sents} and \texttt{pts} give the number of sentences in each bin-specific eval set and the number of tokens in that bin occurring in those references, respectively. The \chrf{} score and number of monolingual training sentences is also given. 
 The interesting results are on the languages, like Fulfulde (\texttt{ff}), that have a high hit-rate on more frequent tokens and a lower hit-rate on rarer tokens. All results are only in the \xxen direction.}
\label{table:grounding_tokens}
\end{table*}
}

\newcommand{\inserteachervsstudent}{\begin{table}[t]	
\centering	
\begin{tabular}{lcccccccccc}	
\toprule	
 & \multicolumn{2}{c}{en-ff}  & \multicolumn{2}{c}{en-kri}  & \multicolumn{2}{c}{en-doi}  & \multicolumn{2}{c}{en-bm}  & \multicolumn{2}{c}{en-ay}  \\	
Mono Size          &\multicolumn{2}{c}{86k}   & \multicolumn{2}{c}{129k}   & \multicolumn{2}{c}{179k}   & \multicolumn{2}{c}{187k}   & \multicolumn{2}{c}{267k}   \\	
	
Direction       & $\leftarrow$ & $\rightarrow$ & $\leftarrow$  & $\rightarrow$ & $\leftarrow$  & $\rightarrow$ & $\leftarrow$ & $\rightarrow$ & $\leftarrow$ & $\rightarrow$ \\	
\midrule	
Teacher &	
45.3 &	35.3 &
64.2 &	35.4 &
65.5 &	36.9 &
38.4 &	34.7 &
39.8 &	28.6 \\
Student &	
45.5 &	37.2 &
64.6 &	36.1&
65.7 &	40.5&
38.6 &	36.4&
40.1 &	34.2 \\
$\Delta$ &	
+0.2 &	+1.9 &
+0.4 &	+0.7 &
+0.2 &	+3.6 &
+0.2 &	+1.7 &
+0.4 &	+5.6 \\
\midrule	
\midrule	
	
 & \multicolumn{2}{c}{en-gom}  & \multicolumn{2}{c}{en-bho}  & \multicolumn{2}{c}{en-kl}  & \multicolumn{2}{c}{en-ee}  & \multicolumn{2}{c}{en-qu}  \\	
Mono Size          &\multicolumn{2}{c}{311k}   & \multicolumn{2}{c}{734k}   & \multicolumn{2}{c}{741k}   & \multicolumn{2}{c}{796k}   & \multicolumn{2}{c}{842k}   \\	
	
Direction       & $\leftarrow$ & $\rightarrow$ & $\leftarrow$  & $\rightarrow$ & $\leftarrow$  & $\rightarrow$ & $\leftarrow$ & $\rightarrow$ & $\leftarrow$ & $\rightarrow$ \\	
\midrule	
Teacher &	
57.9 &	40.8 &  
60.9&	41.2&
39.0&	35.8 &
37.0 &	40.2 &
35.3 &	35.7 \\
Student &	
57.4 &	42 &
61.3 &	42.7&
39.5 &	41.0 &
37.5 &	40.7 &
35.1 &	36.1 \\
$\Delta$ &	
-0.4 &	+1.2 &
+0.4 &	+1.5 &
+0.5 &	+5.2 &
+0.5 &	+0.4 &
-0.1 &	+0.4 \\
	
\midrule	
\midrule	
 & \multicolumn{2}{c}{en-gn}  & \multicolumn{2}{c}{en-ak}  & \multicolumn{2}{c}{en-ts}  & \multicolumn{2}{c}{en-mai}  & \multicolumn{2}{c}{en-ln}  \\	
Mono Size          &\multicolumn{2}{c}{861k}   & \multicolumn{2}{c}{1.3m}   & \multicolumn{2}{c}{1.3m}   & \multicolumn{2}{c}{1.3m}   & \multicolumn{2}{c}{1.4m}   \\	
	
Direction       & $\leftarrow$ & $\rightarrow$ & $\leftarrow$  & $\rightarrow$ & $\leftarrow$  & $\rightarrow$ & $\leftarrow$ & $\rightarrow$ & $\leftarrow$ & $\rightarrow$ \\	
\midrule	
	
Teacher &	
43.0 &	32.1 &
38.6 &	34.3 &
47.5 &	46.5 &
65.5 &	39.2 &
31.7 &	34.6 \\
Student &	
43.4 &	31.9 &
38.9 &	34.4 &
47.7 &	46.8 &
65.5 &	40.0 &
32.1 &	34.6 \\
$\Delta$ &	
+0.4 &	-0.2 &
+0.4 &	+0.1 &
+0.2 &	+0.3 &
0.0 &	+0.8 &
+0.4 &	0.0 \\
\midrule	
\midrule	
	
 & \multicolumn{2}{c}{en-nso}  & \multicolumn{2}{c}{en-lg}  & \multicolumn{2}{c}{en-ilo}  & \multicolumn{2}{c}{en-ti}  & \multicolumn{2}{c}{en-om}  \\	
Mono Size          &\multicolumn{2}{c}{1.9m}   & \multicolumn{2}{c}{2m}   & \multicolumn{2}{c}{2.6m}   & \multicolumn{2}{c}{3.9m}   & \multicolumn{2}{c}{5.6m}   \\	
	
Direction       & $\leftarrow$ & $\rightarrow$ & $\leftarrow$  & $\rightarrow$ & $\leftarrow$  & $\rightarrow$ & $\leftarrow$ & $\rightarrow$ & $\leftarrow$ & $\rightarrow$ \\	
\midrule	
	
Teacher &	
52.5 &	47.1 &
40.5 &	39.3 &
62.7 &	54.1 &
46.1 &	21.5 &
40.7 &	40.0 \\
Student &	
52.8 &	48.2 &
41.0 &	39.8&
62.9 &	54.8&
46.0 &	21.9&
41.5 &	40.1 \\
$\Delta$ &	
+0.3 &	+1.1 &
+0.5 &	+0.5 &
+0.2 &	+0.8 &
0.0 &	+0.5 &
+0.8 &	+0.2 \\
\midrule	
\midrule	
 & \multicolumn{2}{c}{en-sa}  & \multicolumn{2}{c}{en-dv}  & \multicolumn{2}{c}{en-lus}  & \multicolumn{2}{c}{en-as}  & \multicolumn{2}{c}{en-ckb}  \\	
Mono Size          &\multicolumn{2}{c}{6.2m}   & \multicolumn{2}{c}{7.9m}   & \multicolumn{2}{c}{8.3m}   & \multicolumn{2}{c}{9.3m}   & \multicolumn{2}{c}{25.1m}   \\	
	
Direction       & $\leftarrow$ & $\rightarrow$ & $\leftarrow$  & $\rightarrow$ & $\leftarrow$  & $\rightarrow$ & $\leftarrow$ & $\rightarrow$ & $\leftarrow$ & $\rightarrow$ \\	
\midrule	
	
Teacher &	
49.2 &	31.0 &
48.4 &	45.0 &
42.1 &	39.7 &
60.6 &	39.6 &
55.3&	42.6 \\
Student &	
49.2 &	33.3 &
48.4 &	45.3&
41.9 &	41.3&
61.1 &	40.3&
56.4 &	44.3 \\
$\Delta$ &	
0.0 &	+2.3 &
0.0 &	+0.3 &
-0.1 &	+1.6 &
+0.6 &	+0.7 &
+1.1 &	+1.7 \\
	
\bottomrule	
\end{tabular}	
\caption{Performance of teacher model versus student model in \chrf{}.}	
 \label{tab:teacher_vs_student}	
\end{table}	
}	
\newcommand{\insertunmterrorsmain}{
\begin{table*}
\centering
\fontsize{7.0pt}{9pt}\selectfont
\begin{tabular}{l| p{0.45\linewidth}  p{0.45\linewidth}}
\hline
\hline
sl    & reference     & translation \\
\hline
ak & I believe a \textbf{lion} is \textbf{stronger} than a \textbf{tiger}. & 
I think the \mistake{hyena}'s \mistake{hotter} than the \mistake{elephant}. \\
dv & I believe a \textbf{lion} is \textbf{stronger} than a \textbf{tiger}.& I believe a \correct{lion} would be \correct{stronger} than a \mistake{miniature crocodile}. \\
mni & I believe a \textbf{lion} is \textbf{stronger} than a \textbf{tiger}.& I believe a \mistake{snake} is \correct{stronger} than a \mistake{crocodile}. \\
doi & I believe a \textbf{lion} is \textbf{stronger} than a \textbf{tiger}. & I believe \mistake{seizures} are \correct{more severe} than \mistake{epilepsy}. \\
ff & I believe a \textbf{lion} is \textbf{stronger} than a \textbf{tiger}. & I think a \mistake{rabbit} is \correct{stronger} than a \mistake{squirrel}. \\
\hline
mni & The first three colors are \textbf{red}, \textbf{orange},and \textbf{yellow}. & The first three colors are \correct{red}, \mistake{yellow}, and \mistake{blue}. \\

qu &  The first three colors are \textbf{red}, \textbf{orange},and \textbf{yellow}. &The first three colors are \correct{red}, \mistake{fire red}, and \correct{yellow}. \\
sa &  The first three colors are \textbf{red}, \textbf{orange}, and \textbf{yellow}. & The first three colors are \correct{red}, \mistake{yellow} and \mistake{saffron}. \\

ts &  The first three colors are \textbf{red}, \textbf{orange}, and \textbf{yellow}. & the first three colors are \correct{red}, \mistake{purple} and \mistake{pink} \\
yua &  The first three colors are \textbf{red}, \textbf{orange}, and \textbf{yellow}. & the first colors are \correct{red}, \mistake{red} and \correct{yellow}. \\
\hline
sa & In this I use \textbf{capsicum, tomatoes, onions, garlic, green chilies, olives}, etc. and do not use \textbf{cheese} and \textbf{butter} in it.  & Here I deal with \mistake{ greater marjoram, blood fruit, plantain, lassi, green marjoram, jujube}, etc., but I do not deal with \mistake{curd} and \mistake{yogurt}. \\

\hline

ak & I would want to be a \textbf{dog} for a day.  & I want to be a \mistake{crocodile} just one day.   \\
\hline
ak & I would ask my \textbf{cat} some questions about what he's always trying to tell \textbf{him} by \textbf{meowing}. & I will ask my \mistake{friend} some questions about why \mistake{she} is \mistake{crying}.\\
\hline

ak & my \textbf{dog} keeps me moving and enjoying life.  he is man's best friend. & my pet \mistake{cat} teaches me how to live a healthy lifestyle and enjoy being with people. \\

\hline
ak & I went to a \textbf{carnival} yesterday that was located in the middle of nowhere under a huge \textbf{red and white} \textbf{circular striped tent}  & I went to a \mistake{place of pleasure} in a desolate place, unknown to me, where there was a \mistake{parrot} that had a \mistake{golden-yellow} \mistake{coat} \\

\hline
ak & John was working in the \textbf{lighthouse} and went for a walk on the beach one \textbf{night}.    & John was working in the \mistake{synagogue} and he was sound asleep one \mistake{afternoon}. \\        

\hline
ak & this is why \textbf{hair} turns \textbf{grey} with age.  & this is the reason why \mistake{ticks} change into \mistake{worms} after a certain period of time. \\

\hline
sa &  my bad habit is that I \textbf{eat too much}.  &  My bad behaviour is that I \mistake{eat poison}. \\

\hline
sa & \textbf{Dogs} are very intelligent animals, they understand very much about humans.   &  \mistake{Cockroaches} are extremely sharp minded animals, they know about humans correctly \\

\hline
sa & Susy loved her \textbf{space book} and would ask her parents to read it to her every \textbf{night}. & Susie loved her \mistake{newspaper}, and she asked her parents to read it to her in the \mistake{morning}.  \\

\hline
\end{tabular}

\caption{Examples of correct translations (\correct{blue}) and mistranslations (\mistake{orange}), illustrating the model's tendency to make mistakes on distributionally similar nouns.
}
\label{tab:unmt_mistakes_full}
\end{table*}
}

\newcommand{\insertmagnification}{
\begin{table*}
{
\centering
\begin{tabular}{c|ccc}
\hline
\hline
language  & \% mono & \% synth (clean)  & \% synth (noisy)  \\
\hline
pcm & 8\% & 20\% & - \\
kl & 14\% & 18\% & 30\%\\
sa & 13\% & 16\% &  37\% \\
ja & 0.0004\% & - &  1\% \\
\hline
\end{tabular}
\caption{Percent of each data set removed after applying specialized filters. mono refers to the monolingual data from the corpus mined in Section~\ref{sec:data:data}. synth(clean) refers to synthetic data generated by forward translating a clean English monolingual corpus and synth (noisy) refers to synthetic data generated from a noisy web-scraped English corpus. We see that the synthetic text generated by the teacher model exhibited these problems to a greater degree than the monolingual web-sourced data, and that these problems intensified on noisier data.
 \label{tab:error_magnification}}
}
\end{table*}
}

\newcommand{\insertperiod}{
\begin{table*}{
\centering
\begin{tabular}{l| llll}
\hline
\hline
direction &	no TP. & TP & 	$\Delta$ & 	W/L  \\
\hline
\enxx & 	39.32 &	\textbf{39.56} &	+0.24 &	0.77 \\
\xxen & 	48.54 &	\textbf{49.23} &	+0.69 &	1.0 \\

\hline
\end{tabular}

\caption{Comparing \chrf{} on versions of the evaluation sets with and without terminal punctuation (TP) on 26-language distilled models. The Win/Loss ratio is also reported, meaning the fraction of language pairs that saw an increase in \chrf{} by applying the terminal punctuation.  \label{tab:period}
}
}
\end{table*}
}
\newcommand{\inserttablenonuniappendix}{\begin{table*}
\centering
\footnotesize 
\begin{tabular}{ll|ll|ll}
\hline
\multicolumn{2}{c}{Tamazight (ber-Latn)} &  \multicolumn{2}{|c|}{Ewe (ee)} &  \multicolumn{2}{c}{Mooré (mos)} \\
\hline
ASCII & codepoint(s) & ASCII & codepoint(s) & ASCII & codepoint(s) \\
\hline
â & 	U+025B & 	0 & 	U+025B U+0303 & 	à & 	U+0269 \\ 
ç & 	U+010D & 	1 & 	U+025B & 	â & 	U+00E3 \\ 
é & 	U+1E93 & 	2 & 	U+0256 & 	è & 	U+025B \\ 
ê & 	U+1E25 & 	3 & 	U+028B & 	ê & 	U+1EBD \\ 
î & 	U+1E6D & 	4 & 	U+0254 & 	î & 	U+0129 \\ 
o & 	U+01E7 & 	5 & 	U+0192 & 	Î & 	U+0128 \\ 
ô & 	U+1E5B & 	6 & 	U+0263 & 	ô & 	U+00F5 \\ 
û & 	U+1E63 & 	7 & 	U+00E3 & 	ù & 	U+028B \\ 
v & 	U+1E0D & 	8 & 	U+1EBD & 	û & 	U+0169 \\ 
Ä & 	U+0190 & 	- & 	U+0254 U+0303 & 	À & 	U+0196 \\ 
Ç & 	U+010C & 	[ & 	U+0169 & 	Â & 	U+00C3 \\ 
É & 	U+1E92 & 	] & 	U+0292 & 	È & 	U+0190 \\ 
Ë & 	U+1E24 & 	@ & 	U+0189 & 	Ê & 	U+1EBC \\ 
Ï & 	U+1E6C & 	\& & 	U+00C3 & 	Ô & 	U+00D5 \\ 
O & 	U+01E6 & 	\% & 	U+0191 & 	Ù & 	U+01B2 \\ 
Ö & 	U+1E5A & 	` & 	U+014B & 	Û & 	U+0168 \\ 
Ü & 	U+1E62 & 	\^{} & 	U+0194 & 	 & 	 \\ 
V & 	U+1E0C & 	= & 	U+0129 & 	 & 	 \\ 
\$ & 	U+0263 & 	\~{} & 	U+014A & 	 & 	 \\ 
£ & 	U+0194 & 	\$ & 	U+0186 & 	 & 	 \\ 
\hline
\end{tabular}

\caption{Three (possibly incomplete) fonts we reconstructed for three African languages that use an extended Latin character set. Table shows the mapping from the ASCII character to the ``correct'' Unicode codepoint.}
\label{tab:nonuni}
\end{table*}

}

\newcommand{\insertauditinstructions}{\begin{table*}[ht]
\centering
\begin{tabular}{l|rp{0.8\linewidth}}
Code & Weight & 	Description \\ 
\hline
CC &  100	& Natural in-language sentence. It's ok if it has a few small issues, like spelling errors or a few words from another language, or if it's a sentence fragment of reasonable length (about 5 words or more) \\ 
\hdashline
CB & 	50 & In-language, but low-quality. This could be ungrammatical text, boilerplate, or very short fragments. \\ 
\hdashline
CA &  30 &	Correct but ambiguous whether it's in the correct language. This code is only applicable for dialects that are closely related to a major language. For instance, many short sentences in Gulf Arabic may also be valid in MSA, and many written Cantonese sentences might also be valid in Mandarin. \\ 
\hdashline
WD & 	20 & This sentence is in a related but different dialect to the language it's supposed to be in.  This code is only applicable for dialects that are closely related to a similar dialect. For instance, it's supposed to be in Sa'idi Arabic but it's in Egyptian Arabic. \\ 
\hdashline
WL &  0 &	Wrong Language, but still linguistic content \\ 
\hdashline
NL &  0 &	Not a language -- any sort of non-linguistic content. Proper nouns like ``Ibuprofin'', ``Calvin Klein'', or ``Washington DC'' also count as NL. \\ 
\hdashline

\end{tabular}
\caption{Instructions descriptions of the error codes we used to rate samples of our datasets, along with the weight each one is given in the combined quality score.
 \label{tab:audit_instructions}}
\end{table*}
}

\newcommand{\insertntlaudit}{
\begin{longtable}{lrrrrrrr}

Language Name (BCP-47) &         score &         cc &    cb &    ca &    wl &    nl &    wd \\
\hline
Northeastern Dinka (dip) &      100 &   100 &   0 &     0 &     0 &     0 &     0 \\
Zarma (dje) &   100 &   100 &   0 &     0 &     0 &     0 &     0 \\
Dombe (dov) &   100 &   100 &   0 &     0 &     0 &     0 &     0 \\
Dyula (dyu) &   100 &   100 &   0 &     0 &     0 &     0 &     0 \\
Wayuu (guc) &   100 &   100 &   0 &     0 &     0 &     0 &     0 \\
Kalenjin (kln) &        100 &   100 &   0 &     0 &     0 &     0 &     0 \\
Wolaytta (wal) &        100 &   100 &   0 &     0 &     0 &     0 &     0 \\
Assyrian Neo-Aramaic (aii) &        100 &   100 &   0 &     0 &     0 &     0 &     0 \\
Igbo (ig) &     99 &    98 &    2 &     0 &     0 &     0 &     0 \\ 
Balinese (ban) &        98 &    96 &    4 &     0 &     0 &     0 &     0 \\ 
Latinized Hindi (hi-Latn) &       98 &    98 &    0 &     0 &     2 &     0 &     0 \\ 
Twi (ak) &     97 &    94 &    6 &     0 &     0 &     0 &     0 \\ 
Luyia (luy) &   97 &    97 &    0 &     0 &     0 &     0 &     3 \\ 
Ewe (ee) &      96 &    91 &    9 &     0 &     0 &     0 &     0 \\ 
Seselwa Creole French (crs) &   94 &    92 &    4 &     1 &     3 &     0 &     0 \\ 
Bhojpuri (bho) &        94 &    94 &    0 &     0 &     6 &     0 &     0 \\ 
Ilocano (ilo) &   94 &    88 &    12 &    0 &     0 &     0 &     0 \\ 
Caribbean Javanese (jvn) &      94 &    87 &    13 &    0 &     0 &     0 &     0 \\ 
Meiteilon (Manipuri) (mni) &        93 &    87 &    12 &    0 &     0 &     1 &     0 \\ 
Luba-Katanga (lu) &     92 &    83 &    17 &    0 &     0 &     0 &     0 \\ 
Krio (kri) &    92 &    89 &    5 &     0 &     6 &     0 &     0 \\ 
Latinized Tamil (ta-Latn) &       91 &    81 &    19 &    0 &     0 &     0 &     0 \\ 
Lingala (ln) &  90 &    80 &    19 &    0 &     1 &     0 &     0 \\ 
Maharasthra Konkani (knn) &     89 &    85 &    8 &     1 &     6 &     0 &     0 \\ 
Northern Sami (se) &    89 &    86 &    5 &     0 &     9 &     0 &     0 \\ 
Cherokee (chr) &        88 &    78 &    20 &    0 &     1 &     1 &     0 \\ 
Latinized Malayalam (ml-Latn) &   87 &    76 &    21 &    1 &     2 &     0 &     0 \\ 
Latinized Bengali (bn-Latn) &     86 &    72 &    28 &    0 &     0 &     0 &     0 \\ 
Maithili (mai) &        86 &    83 &    5 &     1 &     11 &    0 &     0 \\ 
Hiligaynon (hil) &      84 &    80 &    8 &     0 &     9 &     3 &     0 \\ 
Tok Pisin (tpi) &       84 &    83 &    2 &     0 &     15 &    0 &     0 \\ 
Latinized Chinese	(zh-Latn)     & 84 &80 &	7 &	0 &	13 &	0 &	0 \\
Gulf Arabic (afb) &     83 &    74 &    17 &    1 &     8 &     0 &     0 \\ 
Minangkabau (min) &     81 &    64 &    34 &    0 &     2 &     0 &     0 \\ 
Chuvash (cv) &  80 &    78 &    4 &     0 &     2 &     2 &     14 \\ 
Tamazight (ber-Latn) &  79 &    69 &    20 &    0 &     10 &    1 &     0 \\ 
Latinized Telugu (te-Latn) &      79 &    74 &    10 &    0 &     0 &     16 &    0 \\ 
Libyan Arabic (ayl) &   78 &    68 &    14 &    10 &    7 &     1 &     0 \\ 
Newari (new) &  78 &    76 &    3 &     0 &     21 &    0 &     0 \\ 
Pangasinan (pag) &      75 &    67 &    17 &    0 &     0 &     17 &    0 \\ 
Waray (war) &     75 &    67 &    17 &    0 &     17 &    0 &     0 \\ 
Goan Konkani (gom) &    73 &    67 &    5 &     13 &    13 &    2 &     0 \\ 
Sanskrit (sa) &         73 &    66 &    14 &    0 &     15 &    5 &     0 \\ 
North Levantine Arabic (apc) &  73 &    62 &    21 &    0 &     16 &    1 &     0 \\ 
Sudanese Arabic (apd-SD) &      72 &    60 &    4 &     34 &    2 &     0 &     0 \\ 
Ibibio (ibb) &  72 &    63 &    17 &    0 &     20 &    0 &     0 \\ 
Shona (sn) &    72 &    43 &    57 &    0 &     0 &     0 &     0 \\ 
Sena (seh) &    71 &    71 &    0 &     0 &     29 &    0 &     0 \\ 
Latinized Marathi (mr-Latn) &     67 &    36 &    62 &    0 &     1 &     1 &     0 \\ 
Ancient Greek (grc) &         67 &    67 &    0 &     0 &     29 &    4 &     0 \\ 
Makhuwa-Meetto (mgh) &  67 &    67 &    0 &     0 &     33 &    0 &     0 \\ 
Algerian Arabic (arq) &         64 &    47 &    6 &     47 &    0 &     0 &     0 \\ 
Latinized Goan Konkani (gom-Latn) &       64 &    63 &    0 &     2 &     28 &    7 &     0 \\ 
Ga (gaa) &      58 &    58 &    0 &     0 &     0 &     0 &     41 \\ 
Wolof (wo) &    57 &    16 &    81 &    0 &     1 &     2 &     0 \\ 
Nigerian Pidgin (pcm) &         51 &    43 &    14 &    3 &     40 &    0 &     0 \\ 
Saint Lucian Creole French (acf) &      50 &    50 &    0 &     0 &     50 &    0 &     0 \\ 
Kashmiri (ks-Deva) &    48 &    47 &    1 &     2 &     17 &    33 &    0 \\ 
Latinized Arabic (ar-Latn) &      47 &    33 &    15 &    21 &    6 &     24 &    0 \\ 
Kuanyama (kj) &         43 &    43 &    0 &     0 &     57 &    0 &     0 \\ 
Adangme (ada) &         42 &    42 &    0 &     0 &     0 &     0 &     58 \\ 
Anaang (anw) &  36 &    26 &    19 &    0 &     54 &    1 &     0 \\ 
Mesopotamian Arabic (acm) &     34 &    6 &     0 &     93 &    1 &     0 &     0 \\ 
North Ndebele (nd) &    25 &    25 &    0 &     0 &     50 &    25 &    0 \\ 
Moroccan Arabic (ar-MA) &       16 &    1 &     24 &    9 &     59 &    7 &     0 \\ 
Baluchi (bal) &         10 &    10 &    0 &     0 &     90 &    0 &     0 \\ 
Saidi Arabic (aec) &    0 &     0 &     0 &     0 &     100 &   0 &     0 \\ 
Eastern Baluchi (bgp) &         0 &     0 &     0 &     0 &     100 &   0 &     0 \\ 
Eastern Baluchi (bgp-Arab) &    0 &     0 &     0 &     0 &     100 &   0 &     0 \\ 
\hdashline
mean &     74 &    68 &    10 &    3 &     15 &    2 &     2 \\ 
median &    80 &    74 &    5 &     0 &     3 &     0 &     0 \\ 
\caption{Results of an audit of the datasets we collected, conducted on sampled of 100 sentences each by a mix of native speakers and non-native speakers. The values are the percent of the audited sample that head each label. The ``score'' metric combines these numbers for an approximate notion of the percent of the data that is usable, and is described in Section \ref{sec:audit}.
 \label{tab:audit}}
\end{longtable}
}

\usepackage{bbm}

\usepackage{hyperref}
\usepackage{url}
\usepackage{xcolor}
\usepackage{graphicx}

\usepackage{booktabs}       
\usepackage{amsfonts}       
\usepackage{nicefrac}       
\usepackage{microtype}      
\usepackage{hyperref}
\usepackage{multirow, makecell}
\usepackage[font=small,labelfont=bf]{caption} 

\usepackage{colortbl}

\definecolor{lgreen}{RGB}{73,174,137}
\definecolor{lred}{RGB}{182,49,54}
\definecolor{lorange}{RGB}{255, 128, 0}
\definecolor{lblue}{RGB}{0, 0, 255}

\colorlet{scale10}{lgreen!100}
\colorlet{scale9}{lgreen!60}
\colorlet{scale8}{lgreen!40}
\colorlet{scale7}{lgreen!20}
\colorlet{scale6}{lgreen!10}
\colorlet{scale5}{lred!5}
\colorlet{scale4}{lred!20}
\colorlet{scale3}{lred!40}
\colorlet{scale2}{lred!60}
\colorlet{scale1}{lred!75}

\colorlet{orangescale0}{lorange!0}
\colorlet{orangescale1}{lorange!10}
\colorlet{orangescale2}{lorange!20}
\colorlet{orangescale3}{lorange!30}
\colorlet{orangescale4}{lorange!40}
\colorlet{orangescale5}{lorange!50}
\colorlet{orangescale6}{lorange!60}
\colorlet{orangescale7}{lorange!70}
\colorlet{orangescale8}{lorange!80}
\colorlet{orangescale9}{lorange!90}
\colorlet{orangescale10}{lorange!100}

\colorlet{bluescale0}{lblue!0}
\colorlet{bluescale1}{lblue!10}
\colorlet{bluescale2}{lblue!20}
\colorlet{bluescale3}{lblue!30}
\colorlet{bluescale4}{lblue!40}
\colorlet{bluescale5}{lblue!50}
\colorlet{bluescale6}{lblue!60}
\colorlet{bluescale7}{lblue!70}
\colorlet{bluescale8}{lblue!80}
\colorlet{bluescale9}{lblue!90}
\colorlet{bluescale10}{lblue!100}

\newcommand{\mistake}[1]{{\hphantom{}\color{lorange!70} \bf #1}}
\newcommand{\correct}[1]{{\hphantom{}\bf \color{blue}  #1}}
\newcommand{\enxx}{en$\rightarrow$xx }
\newcommand{\xxen}{xx$\rightarrow$en }
\newcommand{\bleu}{\textsc{Bleu}}
\newcommand{\chrf}{\textsc{ChrF}}
\newcommand{\scaledchrf}{\textsc{ScaledChrF}}
\newcommand{\rttlangidchrf}{\textsc{RttLangIDChrF}}
\newcommand{\tfiif}{\textsc{Tf-iif}}

\newcommand{\howmanylaguagesaudited}{72}
\newcommand{\medianauditscore}{80}

\usepackage{adjustbox}
\usepackage{tikz}
\usetikzlibrary{automata,arrows,arrows.meta,calc,positioning,shadows,decorations.pathreplacing}

\usepackage{enumitem}

\usepackage{arydshln}
\usepackage{stackrel}
\usepackage[T4,T1]{fontenc}
\usepackage{subcaption}

\newcommand{\AS}[1]{{\fontencoding{T4}\selectfont#1}}

\title{\centering \LARGE Building Machine Translation Systems \\for the Next Thousand Languages}

\author{ \centering
Ankur Bapna\thanks{Equal contributions. Correspondence to \{ankurbpn,icaswell\}@google.com. All authors affiliated with Google Research.}\And
Isaac Caswell\footnotemark[\value{footnote}] \And
Julia Kreutzer\And
Orhan Firat\And
Daan van Esch\And
Aditya Siddhant\And
Mengmeng Niu\And
Pallavi Baljekar\And
Xavier Garcia\And
Wolfgang Macherey\And
Theresa Breiner\And
Vera Axelrod\And
Jason Riesa\And
Yuan Cao\And
Mia Xu Chen\And
Klaus Macherey\And
Maxim Krikun\And
Pidong Wang\And
Alexander Gutkin\And
Apurva Shah\And
Yanping Huang\And
Zhifeng Chen\And
Yonghui Wu\And
Macduff Hughes}

\makeatletter
\def\nomarkfootnote{\xdef\@thefnmark{}\@footnotetext}
\makeatother

\iclrfinalcopy 
\begin{document}

\maketitle

\begin{abstract}
In this paper we share findings from our effort to build practical machine translation (MT) systems capable of translating across over one thousand languages. We describe results in three research domains:
(i) Building clean, web-mined datasets for 1500+ languages by leveraging semi-supervised pre-training for language identification and developing data-driven filtering techniques; (ii) Developing practical MT models for under-served languages by leveraging massively multilingual models trained with supervised parallel data for over $100$ high-resource languages and monolingual datasets for an additional $1000+$ languages; and (iii) Studying the limitations of evaluation metrics for these languages and conducting qualitative analysis of the outputs from our MT models, highlighting several frequent error modes of these types of models. Using this approach, we add 24 new languages to Google Translate, the product's largest increase in language coverage to-date. We hope that our work provides useful insights to practitioners working towards building MT systems for currently understudied languages, and highlights research directions that can complement the weaknesses of massively multilingual models in data-sparse settings.
\end{abstract}

\clearpage
\tableofcontents
\clearpage

\section{An Overview}
The past decade has seen tremendous improvements in the quality of academic and commercial machine translation (MT) systems. These improvements have largely been driven by advances in machine learning and the availability of large-scale web-mined datasets~\citep{resnik-smith-2003-web,uszkoreit-etal-2010-large,espla-gomis-2009-bitextor,espla-etal-2019-paracrawl,banon-etal-2020-paracrawl,schwenk-etal-2021-ccmatrix}. The advent of deep learning and sequence-to-sequence models~\citep{sutskever2014sequence,bahdanau2015neural,luong-etal-2015-effective,vaswani2017attention}, large parallel (and monolingual) datasets mined from the web, data augmentation approaches like back-translation~\citep{sennrich-etal-2016-improving, edunov-etal-2018-understanding} and self-training~\citep{he2019revisiting} and massively multilingual modeling~\citep{firat-etal-2016-multi,johnson-etal-2017-googles,aharoni-etal-2019-massively,arivazhagan2019massively,tang-etal-2021-multilingual,fan2021beyond} have enabled high quality machine translation systems that can support over 100 languages.

However, despite tremendous progress in low-resource MT, the number of languages for which widely-available, general-domain MT systems have been built has been limited to around $100$, which is a small fraction of the over $7000+$ languages that are spoken in the world today. Apart from the limited number of languages, the distribution of languages supported by current MT systems is highly skewed in favour of European languages. Despite high speaker populations, languages spoken in Africa, South and South-East Asia and indigenous languages of the Americas are relatively under-served. For example, Google Translate supports Frisian, Maltese, Icelandic, and Corsican, each with fewer than 1M L1 speakers, but not (up until this work) Bhojpuri (\textasciitilde 51M speakers), Oromo (\textasciitilde 24M speakers), Quechua (\textasciitilde 9M speakers), or Tigrinya (\textasciitilde 9M speakers)~\citep{51306}.
We will refer to these languages as \emph{long-tail languages}, since the data scarcity requires the application of machine learning techniques that can generalize beyond the languages for which ample training data is available.

\textbf{Web-crawled datasets for 1000 languages:} The progress towards building machine translation systems in these languages has largely been limited by the lack of digitized and accessible datasets and NLP tools like language identification (LangID) models; such resources are ubiquitous for higher resource languages.
The first stage of this paper describes our approach to building monolingual web text corpora in over 1500 languages, with a particular focus on dealing with common noise, data quality and scale challenges encountered when building a dataset from the web (Section \ref{sec:data}).

Towards this goal, we first scale LangID models to 1500+ languages (\ref{sec:langid}), both using traditional n-gram models and semi-supervised approaches. We next describe several practical approaches and filtering techniques that enable using these LangID models to identify long-tail language data on the web. To minimize recall loss during mining, we cluster languages by error rate (\ref{sec:data:cluster}). To reduce noise from LangID mis-predictions, we leverage document-level LangID consistency to filter our data (\ref{sec:data:firstpass}), followed by percent-threshold wordlist filtering (\ref{sec:data:wordlistfilter}), \tfiif{} filtering (\ref{sec:data:tfiif}), and hand-designed filters to address specific noise issues with certain languages (\ref{sec:data:anomalousness}). We describe the resulting dataset in Section~\ref{sec:data:data} and perform an audit of \howmanylaguagesaudited{} language corpora (\ref{sec:audit}), finding that the data is between 70\% and 100\% in-language, with a median score of \medianauditscore\%.


\textbf{Machine Translation for long-tail languages:} 
With monolingual data mined from the web, the next challenge is to build high quality, general-domain MT models from limited amounts of monolingual training data. We follow a practical approach, utilizing all the parallel data that is available for higher resource languages to boost the quality of long-tail languages where only monolingual data is available\footnote{Preliminary experiments indicated no quality improvements on our evaluation sets when incorporating widely available limited-domain parallel corpora in our models.}. We will refer to this setting as \emph{zero-resource} since no direct supervision is available for our long-tail languages. We want to emphasize that this term is used only in the technical sense, meaning that the model itself is not able to see parallel text; in reality, there is a richness of resources for these languages, including tens of millions of native speakers, centuries (or in some cases millenia) of scholarship, and even large segments of text inaccessible to digital methods. See \citet{bird-2020-decolonising} for some more reflections on this term.

We leverage several techniques that have been developed for MT over the last few years in order to boost the quality of zero-resource translation for long-tail languages. These include self-supervised learning from monolingual data, massively multilingual supervised training, large-scale back-translation and self-training, and high capacity models. We utilize these tools to build MT models capable of translating across over 1000 languages, utilizing our existing parallel corpus spanning around 100 languages and the 1000-language monolingual dataset built from the web (Section~\ref{sec:mt}).  

We first highlight the importance of model capacity in highly multilingual models by comparing the performance of 1.5B and 6B parameter Transformers on zero-resource translation (\ref{sec:mt:capacity}). We then scale up the number of self-supervised languages to 1000, demonstrating that the performance of most long-tail languages improves as more monolingual data from similar languages becomes available (\ref{sec:mt:multilinguality}). While our 1000-language models demonstrate reasonable performance, in order to understand the strengths and limitations of the approach we incorporate large-scale data augmentation. For practical purposes we fine-tune the resulting model on a subset of 30 languages with large amounts of synthetic data via self-training and back-translation (\ref{sec:mt:augmentation}). We further describe practical approaches to filter synthetic data to increase the robustness of these fine-tuned models to hallucinations and wrong-language translations (\ref{sec:mt:filtering}). We also describe our efforts to distill these models into smaller, more inference-friendly architectures using sequence level distillation, and highlight the performance gaps between the teacher and student models (\ref{sec:distill}).

\textbf{Evaluating MT for 1000 languages:} Existing MT systems heavily rely on n-gram overlap based lexical metrics (\bleu)~\citep{papineni-etal-2002-bleu} or the new and emerging model-based metrics like \textsc{YiSi}~\citep{lo-2019-yisi}, \textsc{BLEURT}~\citep{sellam-etal-2020-bleurt} or \textsc{COMET}~\citep{rei-etal-2020-comet} to evaluate translation quality. These metrics are usually computed on static evaluation sets with fixed references obtained from professional translators or crowd-workers.

To evaluate our MT models, we first build an evaluation set for 38 selected languages from the long-tail~(\ref{sec:evalsets}), by translating English sentences into these languages. We highlight the limitations of \bleu{} in the long-tail setting and make the case for evaluating these languages with \chrf~(\ref{sec:bleuvschrf}). We also propose an approximate, round-trip translation based, reference-free metric to understand the quality of our models on languages where reference sets were unavailable, and report the quality of our models as measured on this metric (\ref{sec:rtteval}). We conduct and report the findings from human evaluations of our models (on a subset of 28 languages), confirming that it is possible to build functioning MT systems by following the recipe described in this paper (\ref{sec:human}).

To understand the weaknesses of our massively-multilingual zero-resource models, we perform qualitative error analysis on several languages. We find that our models often confuse distributionally similar words and concepts, e.g. ``tiger'' becomes a ``miniature crocodile''  (\ref{sec:miniaturecrocodile}) and their ability to translate tokens deteriorates on less frequent tokens for lower-resource settings (\ref{sec:tokfreq}). We also find that these models often fail to adequately translate short, or single word inputs (\ref{sec:errorshort}). A study of our distilled models also reveals that all models are more likely to magnify biases or noise present in the training data (\ref{sec:errormagnification}). 

\textbf{Other miscellaneous findings and experiments: } We perform a few additional experiments on these models, demonstrating that they often perform better when translating directly between similar languages --- rather than using English as a pivot (\ref{sec:direct}) and that they can be utilized for zero-shot transliteration between different scripts (\ref{sec:translit}). We describe a practical technique of appending terminal punctuation to any input, called the ``period trick'', that improves the quality of translation (\ref{sec:period}). Furthermore, we demonstrate that these models are robust to nonstandard Unicode glyph usage for some but not all languages (\ref{sec:glyphs}), and we explore several non-Unicode fonts (\ref{sec:nonuni}).

Throughout the course of this work, we rely extensively on native speakers of these languages to guide, evaluate and often contribute technically to the development of these systems. Section~\ref{sec:nativespeakers} highlights the important role they play --- a role that researchers not familiar with the language and community could not have contributed.

As a result of the work outlined in this paper, we add support for 24 new languages to Google Translate. Adding a language to a user-facing product requires considerable effort and individual attention to evaluate each language thoroughly and consult members of affected communities; as a result we limited this effort to 30 languages, of which 24 met our launch bar. This set of languages was chosen to cover languages with large speaker populations in regions that are under-represented in technology, like the Americas, Africa, and South Asia.

\clearpage

\section{Building a 1000-Language Web text Dataset}
\label{sec:data}

It is difficult to uncover large, clean, and highly multilingual corpora from the web. As explored in \citet{caswell2020language}, the task of using LangID models for web-mining is exceedingly challenging for low-resource languages, with noise arising from domain mismatch between LangID training data and web text, idiosyncrasies of n-gram based LangID models, and the massive class imbalance between high- and low-resource languages. This is further borne out by \citet{kreutzer2022quality}, who demonstrate that a variety of public multilingual corpora, created using techniques that work for high-resource languages, are unusably noisy for many low-resource languages.

This problem is compounded by the size of the internet. For lower-resource languages, it is important to get as much data as possible to have any usable signal for a model to learn from. However, the sheer quantity of data online makes it infeasible to apply computationally intensive methods, like Transformer-based models; and even storing unfiltered data on disk can be challenging.

The following subsections explain in detail the approach we took to crawl a monolingual text dataset for 1500+ languages. Our approach focused on recovering high-precision data (high percentage clean, in-language text), so a large portion of the steps are various filtration approaches. The work in this section is an extension of and improvement over the methods presented in \citet{caswell2020language}.

In summary, our approach is as follows:

\begin{description}
    \item[\ref{sec:data:pare}] Omit languages from the LangID model with poor quality training data and poor LangID performance; train both a 1,629-language CLD3 (n-gram) LangID model and Semi-supervised LangID (SSLID) model
    \item[\ref{sec:data:cluster}] Cluster languages by their error rate in the CLD3 model
    \item[\ref{sec:data:firstpass}] Perform a first-pass webcrawl with the CLD3 model
    \item[\ref{sec:data:firstpass}] Filter sentences with document-consistency filtering
    \item[\ref{sec:data:wordlistfilter}] Filter all corpora with percent-threshold wordlist filtering
    \item[\ref{sec:data:sslid}] Filter all corpora with Semi-Supervised LangID (SSLID) filtering
    \item[\ref{sec:data:tfiif}] Detect outliers languages with Relative Recall Rate and filter them with Term-Frequency-Inverse-Internet-Frequency (\textbf{\tfiif{}}) filtering
    \item[\ref{sec:data:anomalousness}] Detect outliers with Token-Frequency Anomalousness score and hand-design filters for them
     \item[\ref{sec:data:dedupe}]  De-duplicate all corpora at a sentence level
\end{description}

\subsection{Steps necessary to create the dataset}
\label{sec:data:dummy}

\subsubsection{LangID modeling}
\label{sec:langid}
As described in \citet{caswell2020language}, the task of reliably detecting what language a given portion of text is in, known as Language Identification or \textit{LangID}, is surprisingly challenging for low-resource languages on the web.

There are two types of LangID models used in this work: CLD3 models \citep{cld3} and Semi-Supervised LangID (SSLID) models \citep{caswell2020language}. The CLD3 models are a multi-layer perceptron on top of a bag of character ngrams, and are very efficient, but suffer from various error pathologies. The semi-supervised models use the MASS objective~\citep{DBLP:journals/corr/abs-1905-02450} on noisy web text in addition to the LangID prediction task, and use the Transformer Big~\citep{vaswani2017attention} architecture. They are generally more accurate and also more robust to web noise, but are much slower to run inference with.

Overall, three LangID models were used for this work: 1) a 1,745-language CLD3 model to measure the quality of the training data (Section \ref{sec:data:pare}); 2) a 1,629-language CLD3 model to cluster languages and label sentences when performing a deep crawl of the web (Sections \ref{sec:data:cluster} and \ref{sec:data:firstpass}); and 3) a 1,629-language semi-supervised LangID model for filtering (Section \ref{sec:data:sslid}).

The training and evaluation data for our LangID models is identical to that described in \citet{caswell2020language}; namely, we trained on an aggregation of proprietary and publicly available text corpora, with an average of 800K tokens per language. Some of the data came from sources with language tags like Wikipedia or Corpus Crawler \citep{corpuscrawler}, while another subset was created using a text elicitation task where we prompted native speakers to write sentences in their language \citep{48745}.

\subsubsection{Paring down the set of languages}
\label{sec:data:pare}
An important first step is determining what languages to crawl the web for. In an ideal world we would crawl all possible languages, but some of them may have poor-quality training data, or be very close to each other and be hard to distinguish.

We trained a CLD3 LangID model on 1,745 language varieties, and investigated languages that seemed to be having large quality losses. Our metrics for considering a language as possibly presenting difficulties were 1) LangID precision was under 33\%; 2) language had over 50\% confusion (False Negative Rate, FNR, or False Discovery Rate, FDR) with respect to another language; 3) a preliminary web-crawl with this model indicated severe overtriggering with a high-resource language; 4) LangID training data had under 2000 examples. 
Following this analysis we removed 116 languages from the model, primarily based on poor-quality training data. Among these most were regional dialects in Europe. 

We additionally did not try to crawl data for the highest-resource 83 languages, in part because of storage space limitations, and in part as we already had monolingual data for these languages from a previous web crawl.

\subsubsection{False Negative Rate clustering}
\label{sec:data:cluster}
For the first-pass, we wanted to avoid lost recall between related dialects, of which many existed in the 1,629 languages. For instance, our training data contains many different varieties of Hindustani, e.g. Haryanvi, Garhwali, Magahi, and so on.

To mitigate this, we clustered languages based on their False Negative Rate (FNR), as measured on our LangID evaluation sets. Of the many error rates one could cluster on, we chose to cluster on FNR to minimize the loss in recall for related languages. We clustered using the Sklearn~\citep{scikit-learn} implementation of Hierarchical Agglomerative Clustering, using \verb|distance_threshold=None| and \texttt{affinity="precomputed"}, and \texttt{linkage=average}. These parameters were chosen largely because they seemed to produce the nicest distribution in cluster sizes. Since some of the clusters were still excessively large, we re-split the largest ones by hand such that no cluster had more than 20 languages. We split these clusters by hand since re-splitting using clustering algorithms still resulted in very uneven sizes, probably because these clusters have hubs with poor or noisy data.

To observe the effect of clustering predictions, we compare the size of a dataset for a language if document-consistency filtering (Section \ref{sec:data:firstpass}) were applied on the language-code level (which would result in lower recall) or on the cluster level. The median ratio between these sizes was 1.006, meaning that for most languages, clustering didn't significantly improve recall. However, for some languages it improved recall significantly, with 57 languages showing a dataset size increase of greater than 20x. The languages with large wins here were by-and-large to be expected, including Hindustani and Arabic varieties, and a variety of cases like Oromo (\texttt{om}) and Eastern Oromo (\texttt{hae}).

The higher-resource languages omitted in this crawl (Section \ref{sec:data:pare}) were all put in their own clusters.

\subsubsection{Document Consistency filtering and First-pass LangID}
\label{sec:data:firstpass}

To start with, we performed LangID prediction on every sentence in every document with the 1,629-language CLD3 n-gram LangID model. Having obtained these predictions, we applied \textit{document consistency filtering} \citep{caswell2020language}. This is one of the simplest and by far most effective filtering steps. We simply discarded any sentence whose sentence-level LangID cluster prediction did not match the document-level LangID cluster prediction. We defined the document-level LangID cluster prediction as the most-often predicted cluster among all sentences --- e.g. if a document had 20 sentences in cluster A, 19 sentences in cluster B, and 18 in cluster C, we gave it a document-level ID of cluster A.

\begin{figure}[t!]
\begin{center}
\includegraphics[scale=0.25]{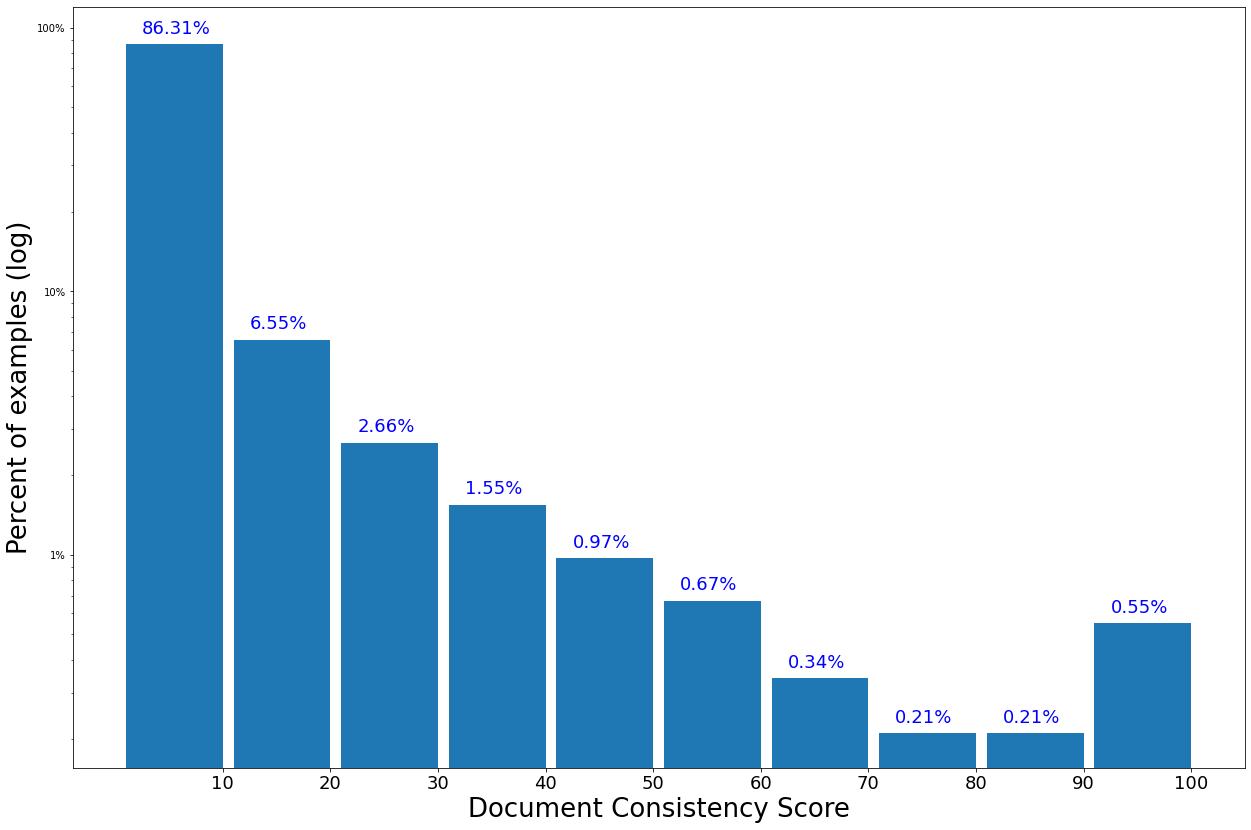}
\caption{Histogram of document consistency scores, using a 1,745-language CLD3 LangID model, on web text. The large majority of sentence-level LangID predictions had a score of under 10\%, indicating that they were likely noisy predictions on pages in other languages or non-linguistic content. Of the 0.55\% with document consistency score over 90\%, over half (0.3\%) had a score of 100\%. }
\label{fig:documentconsistency}
\end{center}
\end{figure}

To investigate the effectiveness of document consistency filtering, we looked at the \textit{document consistency score}. For a given sentence in a document, the document consistency score is simply the percent of sentences in that document sharing the same LangID prediction as that sentence. (Therefore, if the document consistency score is over 50\%, it will never be filtered out with document consistency filtering.) Figure \ref{fig:documentconsistency} shows the deciles for document consistency score across all languages, for a sample web crawl with the initial, 1,745-language CLD3 LangID model. The distribution is heavily weighted towards the lower values, with a whole 83\% of sentences having a score under 10\%.   The large mass of sentences with a low document consistency score indicates that there are many single, random sentences in larger documents, including non-linguistic content, whose language is mis-predicted. This suggests a more refined approach to document-consistency filtering. Using the method outlined above, a page with content in multiple languages could only yield data for one language; if instead any sentence were preserved with a document consistency score under a particular threshold (say, 0.3), documents with multilingual content would be handled much better, while preserving most of the benefits of filtering. This is left for future work.

\subsubsection{High-recall Wordlist-filtering}
\label{sec:data:wordlistfilter}
Following  \cite{caswell2020language}, we apply percent-threshold wordlist filtering to all languages. A sentence was discarded if it had $<20\%$ in-language words for any of the languages in the cluster, where the wordlists were the most frequent 800 words from the LangID training data.

\subsubsection{Filtering and Declustering with Semi-supervised LangID}
\label{sec:data:sslid}
Once the first-pass, clustered filtering was completed, we classified each sentence with the more computationally expensive Semi-Supervised LangID model (SSLID), as described in \cite{caswell2020language}, resulting in per-language corpora. If the SSLID model predicted any language outside of the cluster, we filtered that sentence. The CLD3 predictions were ignored.

\subsubsection{\tfiif{} Filtering}
\label{sec:data:tfiif}

Even after these three rounds of filtering --- CLD3, percent-threshold, and SSLID -- many languages still had extremely noisy data. The worst case was Tok Pisin (\texttt{tpi}), whose dataset consisted of 1.3B sentences, of which over $99.7\%$ were in Standard English (mostly containing the word ``long'', which is also a common function word in Tok Pisin). Therefore, as in \citet{caswell2020language}, we applied percent-threshold filtering with \textbf{\tfiif{}} (Term-Frequency Inverse-Internet-Frequency) lists \footnote{Open-sourced at \url{https://github.com/google-research-datasets/TF-IDF-IIF-top100-wordlists/}.}, meaning that we retained any sentence which contained at least 20\% of its tokens in our \tfiif{} wordlists. However, we optimize the approach used in \citet{caswell2020language}, and furthermore develop a heuristic metric to determine whether to apply the filtering on a per-language basis.

In contrast to \citet{caswell2020language}, we omitted the \texttt{IDF} term entirely, since this term is influenced by the set of languages one considers, and becomes less helpful as the number of languages scales. We also adjusted the values of the parameters $\kappa$ and $\tau$ (described below). To understand these changes, we can revisit the formulation for \tfiif{}. For a token $t$ in a language $l$, with a frequency function $f(term, corpus)$ and language-specific corpora $D_l$:

\begin{equation}
\textrm{TFIIF}_{t,l} = \textrm{TF}_{t,l} *  \textrm{IIF}_{t} = \frac{f ( t, D_l )}{\textrm{max}(f ( t, internet ), \alpha)}
\label{eq:tfiif}
\end{equation}

Note the clipping parameter $\alpha$ in the IIF term, which is introduced to account for OOV tokens (which are common) and noise near the tail of the distribution. For this work we set $\alpha = f(w_\kappa)$ --- in other words, we only consider the top $\kappa$ terms in the empirical frequency distribution, and give all less common terms the same weight as the $\kappa^{th}$ most common token. A higher value of $\kappa$ means that more words are covered by the frequency distribution -- which seems good -- but it also means that OOV words will have a higher weight, since the IIF term is an inverse. In the worst case this could mean that the resulting \tfiif{} wordlist would contain too many rare words. An especially low value of $\alpha$, e.g. $\alpha < 1.0$, would essentially guarantee this, pushing up only OOV words to the top of the list.

There is another parameter hidden in this formulation, namely $\tau$. This parameter is simply how many words to include in the wordlist. For instance, $\tau = 1000$ means that the wordlist used for filtering includes the top 1000 words by \tfiif{} score.

In order to investigate the effect of these parameters, we define a heuristic measure of \textit{distractibility} as a proxy for precision, which measures how much a given language's data is likely to be polluted by a common high-resource language. For a language $l$ and a set of distractor languages $\mathcal{D}$, we define the distractibility $\delta_l$:

\begin{equation}
\delta_l = \textrm{max}_{d \in \mathcal{D}} (\textrm{FDR}(d, l))
\label{eq:distractability}
\end{equation}

Where, for our purposes, we have chosen $\mathcal{D}$ \texttt{= \{en, de, es, hi, id, ar, ru\}}, and the False Discovery Rate (FDR) of a distractor language $d$ with regard to a language $l$ is defined in the standard way, as (on a balanced eval set)

\begin{equation*}
\textrm{FDR(}d, l\textrm{)} = \frac{\#  \textrm{examples in language } d \textrm{ mis-predicted as }l}{ \# \textrm{true examples in language } l}
\end{equation*}

Despite the concern that it would elevate rare and OOV words, we found that increasing the value of $\kappa$ steadily decreased the distractibility with little loss in recall on the target languages. Increasing the size of the wordlists, $\tau$, recouped the loss in recall, but increased distractibility to more dangerous levels. These results can be seen in Table \ref{tab:tfiif_kappa}.

Therefore, for our \tfiif{} lists, we use values of $\kappa=80000$ and $\tau=1000$. For comparison, \citet{caswell2020language} use $\kappa = 10000$, and the value of $\tau$ was set on a per-language basis from the recall on the dev set. Furthermore, \citet{caswell2020language} use an \texttt{IIF} list with only 980k unique tokens (7M webpages), whereas this work uses 41M unique tokens (250M web pages). The public GitHub repo has been updated to reflect these improvements.

\textbf{Deciding when to apply \tfiif{} filtering:} Many languages do not need extra filtering, and it is important not to overfilter and decrease recall. Therefore we needed an outlier detection metric to determine whether filtering needed to be applied. We wanted to filter those corpora where the loss in recall was small, but the reduction in dataset size was large (indicating that many out-of-language sentences were removed).

We looked at the percent of our LangID eval sets that remained after filtering ($r_{\textsc{tfiif}}(\textrm{gold})$), aka the filtering recall), and compared it to the percent of our web-crawled corpora remaining after filtering ($r_{\textsc{tfiif}}(\textrm{crawl})$. We used these to define the heuristic Relative Recall Rate (\textsc{RRR}) as follows:

\begin{equation}
    \textsc{RRR}_{\textsc{tfiif}} = \frac{r_{\textsc{tfiif}}(\textrm{gold})^\rho}{r_{\textsc{tfiif}}(\textrm{crawl})}
\end{equation}

This quantity measures, approximately, how many true positives are kept by this method for every false positive filtered out. The exponent $\rho$ is the trade-off between recall and percent filtered. A value of $\rho>1$ means that we weight a loss in recall (undesired outcome) more than an equivalently sized reduction in data size (desired outcome). One way of interpreting this is how much we trust the recall on our eval set. If $\rho=1$, we trust it perfectly. However, in practice, since we fear that the recall on our eval set overestimates the true recall on natural text on the web, we set $\rho>1$.

For our experiments we used a value of $\rho=2$. This flagged 895 out of 1503 languages. However since we were only concerned about datasets that were more severely polluted, we only filtered datasets where the filtering would remove 20\% or more of the data ($r_{\textsc{tfiif}}(\textrm{crawl}) \leq 80$). We also decided not to filter any language where the recall ($r_{\textsc{tfiif}}(\textrm{gold})$) was less than 80\%, in case we lost too much usable data. 
The result of this was that we applied \tfiif{} filtering to 210 out of 1503 language corpora.


\inserttfiifkappa

\subsubsection{Token distribution anomaly detection and Negative Token Filters}
\label{sec:data:anomalousness}

Even after the filtering steps described in the previous sections, a variety of languages still had data-quality issues. One issue that was not easily filtered out by approaches like \tfiif{} was templated content, which is technically in the right language, but is not useful for training data. To a lesser extent we also saw issues like the ``unlucky n-gram'' \citep{caswell2020language} effect. Examples of the type of content we found were:
\begin{enumerate}
    \item Scottish Gaelic (\texttt{gd}) found 570M in-language sentences even after \tfiif{} filtering. It turned out that this was mostly from one site, and the most common token was ``Luchdaich a-nois'' (``download'')
    \item Darija (\texttt{ar-MA}) came up with a dataset of over a billion sentences, but 94.9\% contained some reference to ``casinos'', ``gambling'', etc.
    \item Kurukh (\texttt{kru-Mlym}) was mainly A N T S P E A K. (And it later turned out that the training data was spurious.)
    \item The Arabic-script Indo-Aryan languages of Kalami (\texttt{gwc}), Hindko (\texttt{hnd}), and Torwali (\texttt{trw}) had picked up masses of templates from unit conversion websites (``X lbs is Y ounces'', ``convert Euro to US American Dollar'', etc.)
    \item Many Latn-script Indic languages (\texttt{hi-Latn}, \texttt{ml-Latn}, etc.) had large amounts of content that were just download links or titles for videos, songs, and so on.
    \item Cree (\texttt{cr-Latn}) was almost 100\% ``Lorem ipsum'' sentences
\end{enumerate}

While some of these are actually incorrect content, much of it is technically in-language, and therefore can't be filtered out by \tfiif{} filtering or straightforward application of LangID. Therefore, we decided to develop an approach to detect anomalous data and investigate datasets that looked like they had issues.

To detect these extreme domain shifts, we hypothesized that the token distribution would be severely skewed. Therefore, we compared the distribution of the tokens in the LangID train data (the \textit{reference distribution}) to the token distribution in the crawled data (the \textit{empirical distribution}). To compare these distributions we looked at several scores for the top N=40 tokens in the empirical distribution:

\begin{itemize}
    \item \textbf{2n-overlap}: This is simply the percentage of the top N tokens that appear in the top 2N tokens of the reference distribution; this metric is very simple and highly interpretable.
    \item \textbf{Euclidean}: This is the Euclidean distance between the frequencies of the top N tokens and their corresponding frequencies from the reference distribution.
\end{itemize}

Different scores, like Jenson-Shannon divergence and Pearson R, were initially considered but found to be ineffective.

We then combined these two scores together with the harmonic mean, yielding the \textit{Harmonic Token Anomalousness Score}. This is a very approximate measure, but still gives a useful score for outlier detection. Based on some qualitative analysis, we determined that a score of $<0.70$ is a sign of a questionable dataset quality. Interestingly enough, however, datasets with scores that were too high also had issues: a score of $> 0.97$ often indicated that the web crawl had merely recovered the training data, which in these cases was often religious material.

After computing this score for all datasets, we manually observed samples of the data for all languages with Harmonic Token Anomalousness Score $< 0.7$ and more than 20,000 sentences. There were 179 languages flagged as suspicious in this way. It was relatively straightforward to make filters for 62 of these, for instance excluding sentences containing ``casino'' in Arabic dialects. For some of the others, we made notes that they were the wrong language. For many others, there was no clear or obvious solution, so we left them as-is. These filters removed on median 21\% of the data for these 62 languages.

We acknowledge that this measure of quality is very approximate, so it should be used judiciously.

\subsubsection{Deduplication}
\label{sec:data:dedupe}

The last step was simply to remove duplicate sentences in all datasources. This reduced the median dataset size by a factor of 1.8x, and the average dataset by a factor of 1.4x.

\subsection{Description of Resultant Dataset}
\label{sec:data:data}

\subsubsection{Monolingual Data}
The result of the process described in section \ref{sec:data:dummy} was a dataset with corpora for 1503 low-resource languages, ranging in size from one sentence (Mape) to 83 million sentences (Sabah Malay). For our experiments we chose to experiment only on those 1057 languages where we recovered more than 25,000 monolingual sentences (before deduplication). We combined this with our existing monolingual datasources for the 83 high-resource languages. Table \ref{tab:datasize_summary} shows statistics for these three datasets --- the full low-resource dataset (``LRL-full''), the portion of the full low-resource dataset used for model training (``LRL-train''), and the full training dataset (``all-train''), which combines in the 83 higher-resource languages.

\subsubsection{Monolingual Data Quality Audit}
\label{sec:audit}
We conducted an audit of our data as in \citet{kreutzer2022quality} with a variety of volunteers, comprising native speakers and non-speakers willing to do detective work. From the sample of \howmanylaguagesaudited{} languages we audited, the median data score was \medianauditscore\%. The ``score'' is a simple heuristic to estimate the percent usable data given the different error codes, and is defined as $1.0*\textrm{cc} + 0.5*\textrm{cb} + 0.3*\textrm{ca} + 0.2*\textrm{wd}$, where $\textrm{cc}$ is the percent of the sample labeled as ``Correct'', $\textrm{cb}$ is the percent labeled as ``correct, Low quality'',  $\textrm{ca}$ is the percent labeled as ``Correct, ambiguous dialect'', and   $\textrm{wd}$ is the percent labeled as ``Correct, wrong dialect'' (See Appendix Section \ref{appendix:audit} for details).

The largest error category tended to be Wrong Language, with an average value of 16\%, followed by Low Quality/Boilerplate, with an average of 10\%. Languages with the poorest quality tended to be close dialects of major languages, especially varieties of Arabic (Sa'idi, Moroccan, Mesopotamian, Latinized, Algerian: \texttt{aec,ar-MA,acm,ar-Latn,arq}), but also varieties of French (Saint Lucian Creole: \texttt{acf}) and English (Nigerian Pidgin: \texttt{pcm}). There were also some close varieties that had higher quality, including varieties of French (Seychellois Creole: \texttt{crs}), Hindi (Bhojpuri: \texttt{bho}) and English (Krio: \texttt{kri}). Of these three, \texttt{kri} and \texttt{crs} use quite different orthographies than their high-resource relatives. In addition to close varieties, some African languages (Ndebele, Anaang, Kwanyama, Wolof: \texttt{nd,anw,kj,wo}) also had very poor quality. Less common languages with extremely \textit{high} quality, e.g. Northeastern Dinka (\texttt{dip}), Zarma (\texttt{dje}), and Dombe (\texttt{dov}), each with a score of 100, may also be viewed with suspicion, as this may mean that a very narrow domain has been recovered (usually religious text).  Details on the per-language performance and the set of error codes we used can be seen in Appendix Section \ref{appendix:audit}.

\inserttabledatasizesummary

\subsubsection{Parallel Data}
In addition to the monolingual data described above, we also utilize the web-crawled parallel corpora available to us. This corpus is a slightly extended version of the corpus described in~\citet{arivazhagan2019massively}, containing approximately $25$ billion sentence pairs spanning $112$ languages, to and from English. 

\clearpage

\section{Building Machine Translation Models for Long-Tail Languages}
\label{sec:mt}

Next, we utilize the datasets described in Section~\ref{sec:data:data} to build our MT models. As we see from the monolingual data statistics in Table~\ref{tab:datasize_summary}, while there are over 1M sentences per language present in our corpus for the highest-resource 205 languages, the median number of sentences in our full training corpus is only around 43K sentences per language. This is a relatively small amount of data, given that traditional MT systems often require a few hundred thousand to a few million \textbf{parallel} sentences to reach good quality, while we are limited to around 43K \textbf{monolingual} sentences per language.

Given the highly data sparse setting, it is clear that the approaches that work for high resource languages cannot apply directly in our setting. In order to build high quality models for long-tail languages, we build on the approach developed previously in~\citet{siddhant2022towards}
to enable zero-resource translation, by leveraging (i) Self-supervised training on in-language monolingual data, (ii) Massively multilingual supervised translation on out-of-language data, (iii) Large-scale data augmentation via back-translation and self-training and (iv) Model scaling. In this section we describe our recipe for training MT models for long-tail languages, starting with massive massively multilingual models and concluding with smaller, inference-friendly models trained on a selected subset of languages.

We start by highlighting the role of capacity in our highly multilingual setting by comparing the performance of $1.5$B and $6$B parameter models in Section~\ref{sec:mt:capacity}. These experiments were conducted on an earlier version of our dataset, spanning $112$ supervised and $94$ zero-resource languages. In Section~\ref{sec:mt:multilinguality} we scale up our models to cover over $1000$ languages and highlight the effect of increasing multilinguality in the long-tail setting. To further increase performance, we incorporate large-scale back-translation and self-training. In Section~\ref{sec:mt:augmentation}, We evaluate these approaches on a subset of $30$ languages for practical considerations. In Section~\ref{sec:mt:filtering}, we further highlight the role of filtering synthetic data and its effect on model quality for close dialects. Finally we describe our distillation approach and compare the resulting student models against the fine-tuned teachers, in Section~\ref{sec:distill}.

Of the languages we evaluate on in this work, only Sorani Kurdish (\texttt{ckb}) uses any in-language supervised data; all other languages are evaluated under a zero-resource setting. While small amounts of parallel resources weren't difficult to obtain for many of these languages, preliminary experiments did not result in any quality improvements with including limited amounts of (often religious-domain) parallel text. 

\subsection{Experimental Setup}
\label{sec:mt:setup}
We utilize the 1000-language monolingual corpus available to us, together with large amounts of massively multilingual parallel data in 112 languages (including English), to build zero-resource MT models for 1000 languages following the approach described in~\citet{siddhant2022towards}. To elaborate, we train Transformer-based MT models on the translation task on 112 languages, simultaneously with the MASS masked denoising task~\citep{DBLP:journals/corr/abs-1905-02450} to first build zero-resource models capable of translating from \xxen with reasonable quality. These models follow a two-stage training procedure, where the first stage consists of training on just the MASS and supervised translation tasks, and the second stage also incorporates data generated using online translation from the model's latest checkpoints (in the \xxen direction). This data is used as back-translated and self-training (aka forward-translated) data simultaneously.

All our models are based on the standard Transformer architecture, with 32 layer encoders and decoders. We use two variants of the Transformer; a smaller model, with approximately 1.5B parameters, and a larger one with 2x larger model dimension and 2x wider hidden dimension, with approximately 6B parameters. Our models utilize the vocabulary from a 64K-token SentencePiece model (SPM)~\citep{kudo-richardson-2018-sentencepiece} trained on monolingual data from the entire set of languages covered by the model. Data is upsampled with a temperature of $T=5$ \citep{arivazhagan2019massively} before creation of the SPM. An analysis of the vocabulary suggests that our vocabulary is close to character-level for all but the highest resource languages. We use GPipe pipeline parallelism~\citep{huang2019gpipe} and the Tensorflow-Lingvo~\citep{shen2019lingvo} framework for all our MT experiments.

Different from~\citet{siddhant2022towards}, in addition to the \texttt{<2xx>} token that was prepended to the source sequence to signify the target language for both translation and MASS tasks, we add a \texttt{<2task>} token (\texttt{<2translation>} for the translation task, and \texttt{<2mass>} for the MASS task) that specifies the task to be performed by the model. We find this to be critical for zero-resource performance, especially when model sizes are scaled up. In the absence of this task token, our models learnt to `infer' the task from the source languages instead of relying on the \texttt{<2xx>} token, resulting in copying sentences when provided zero-resource language sentences with the \texttt{<2en>} token.

All our models are evaluated with \chrf{} on standard evaluation sets built by translating 1200 English sentences into the target language, for 38 languages. Details of our evaluation metrics and datasets are provided in Section~\ref{sec:eval}.

\insertmtcapacity
\subsection{Experiments comparing effect of model capacity}
\label{sec:mt:capacity}
We first compare the effect of model capacity on zero-resource performance. In this experiment we train our 1.5B and 6B Transformer variants on supervised data from 112 languages, to and from English, and monolingual data from 206 languages (including the above 112 languages). These models are first trained with MASS and translation, and as a second stage, additionally trained on online-translated monolingual sentences as back-translation and self-training data. The results of these experiments are listed in Table~\ref{tab:mtcapacity}. We demonstrate results for 38 languages, of which we didn't include any monolingual data for 6 languages (which weren't present in our initial scrape). We find that increasing the capacity of the model from 1.5B to 6B has a significant effect on the quality of translation, improving by an average of 2.7 \chrf{} on the \xxen translation direction. On the other hand, the \enxx direction regresses in quality with an average loss of -1.0 \chrf. This degradation was associated with the same issue that required us to append extra \texttt{<2task>} tokens to the source input (models learning to infer the task from the input language, resulting in copying instead of translation). 

Looking at the overall results, we notice that the performance of these models reaches above 40 \chrf{} in the \xxen direction for a majority of the languages. Performance is especially high for languages which either have similar languages in our parallel corpus (including South-Asian languages and Pidgins like Bhojpuri (\texttt{bho}), Dogri (\texttt{doi}), and Nigerian Pidgin (\texttt{pcm})), while \chrf{} scores are relatively low for lower-resource and those with no related languages in the model (including Native American languages like Quechua (\texttt{qu}), K'iche' (\texttt{quc}), Aymara (\texttt{ay}), and Kalaallisut (\texttt{kl})). We observe that our model learns to translate Goan Konkani (\texttt{gom}) into English with surprisingly high quality despite the lack of \texttt{gom} monolingual data seen by the model, perhaps owing to its similarity with Marathi (\texttt{mr}), which is present in our supervised data. 

\insertmtmultilinguality
\subsection{Continual learning, extending models from 200 to 1000 Languages}
\label{sec:mt:multilinguality}
We next compare our 6B model trained on monolingual data from 200 languages against one trained on our entire monolingual set, spanning 1000 languages. To train the 1000-language translation model we utilize the continual learning technique described in~\citet{garcia-etal-2021-towards}. To elaborate, we replace the vocabulary used by our 200-language MT model with one trained on monolingual data from all 1000 languages. To be able to reuse the model learnt on 200 languages to initialize the newer, 1000 language model, we align the SPM tokens shared across the two vocabularies and assign them the same IDs. The tokens in the new vocabulary that do not map to any token in the original vocabulary are assigned random IDs not assigned to the aligned tokens. This allows to continue training our 1000-language model from the model trained on 200-language subset.


We compare the performance of our 200-language model against the 1000 language MT model in Table~\ref{tab:mtmultilinguality}. Unsurprisingly, languages that were not covered by our 200-language model but are now covered by our new monolingual data, including Goan Konkani (\texttt{gom}), Isoko (\texttt{iso}), Kalaallisut (\texttt{kl}) and Tsonga (\texttt{ts}), see large quality improvements. However, we also observe quality improvements almost for all languages in our evaluation set, including both, the \xxen and \enxx translation directions with average improvements of +2.5 and +5.3 \chrf{} points respectively. This is counter-intuitive since we are increasing the number of self-supervised languages, which should presumably increase the amount of interference and worsen capacity contention within this model, similar to what was observed in~\citet{siddhant2022towards}. What is different from earlier studies is the extent of multilinguality in the model; any new added language in our model is likely to be very similar to one of the existing languages, resulting in stronger transfer between those languages. This transfer effect is strengthened by the low quantity of monolingual data for long-tail languages; in the massively multilingual and resource constrained setting the cross-lingual transfer effect dominates the interference observed in massively multitask models. This is also supported by the fact that languages that benefit the least from increased multilinguality are Native American languages like Quechua (\texttt{qu}), K'iche' (\texttt{quc}), Aymara (\texttt{ay}),  Yucatec Maya (\texttt{yua}), and Kalaallisut (\texttt{kl}), which tend to have few or no similar languages in the training data.

While our evaluation sets are limited to 38 languages, we provide \rttlangidchrf{} as a highly approximate measure of the quality of the model on all 1000 languages in Appendix Table~\ref{tab:datasize_full}. More details of the metric and the performance of the model on all languages are described in Section~\ref{sec:rtteval}.

\subsection{Effects of Large Scale Data Augmentation}
\label{sec:mt:augmentation}
\insertmtfinetuning
In order to understand the limits of quality with our zero-resource approach, we select a subset of 30-languages for large-scale data augmentation. We continue training our 1000-language model on this subset of 30-languages with the MASS, translation and online back-translation objectives. To leverage the full power of data augmentation, we translate all the available monolingual data (up to 10 million sentences) for these languages into English, and sample around 10 million English web sentences which are translated into each of these languages. The model is then continued training with MASS, translation and offline back-translation and self-training with this synthetic data. This process is repeated twice since quality improvements with subsequent stages were observed to be incremental. This model is then compared against our vanilla 1000-language model that wasn't trained with large-scale augmented data.

The results of this comparison are depicted in Table~\ref{tab:mtfinetuning}. We notice that fine-tuning with augmented data shows substantial quality improvements across the board, with especially large improvements for \enxx translation for several languages. Excepting our Native American subset, most languages reach greater than 40 \chrf{} on \xxen translation and greater than 35 \chrf{} in the reverse direction.

\subsection{Effect of Filtering Synthetic Data}
\label{sec:mt:filtering}
\insertmtfiltering
One challenge with using the model's own predictions for improving model quality is the presence of positive feedback loops which can magnify any problems with the model outputs, as elaborated in Section~\ref{sec:errormagnification}. To reduce these effects we compare our fine-tuned model against a version that was fine-tuned with a round-trip translation and LangID filtered version of our synthetic data for training. 
The results of this comparison are depicted in Table~\ref{tab:mtfiltering}. For most languages this additional filtering has no major impact on model performance, often performing within 0.5 \chrf{} of the non-filtered model. However, for certain languages like Nigerian Pidgin (\texttt{pcm}) and Kalaallisut (\texttt{kl}) we observe large improvements in quality. Closer inspection reveals that our \texttt{pcm} data suffered from mixing with African-American Vernacular English (AAVE), which gets magnified as the model trained on its own outputs. Similarly, our \texttt{kl} monolingual set was polluted with a small fraction of Danish (\texttt{da}) data, which gets magnified with self-training. Filtering with round-trip-translation and LangID reduces the instances of data pollution improving the correctness of the model outputs. Section \ref{sec:errormagnification} explores this in greater depth.

\subsection{Distillation}
\label{sec:distill}
We next describe our approach for distilling our best 6B parameter Transformer model fine-tuned on $30$ languages into smaller, more efficient, architectures. This process also yields further quality gains.

\subsubsection{Data Generation for Distillation}
We follow the sequence-distillation approach~\citep{kim2016sequence} to distill our teacher model into smaller students. To this end, we generate large amounts of synthetic forward- and backward-translated data with the teacher model, which is then used to train a smaller student model.
We start with the $30$-language fine-tuned model and translate roughly $15$-$25$ million English sentences into each of the $30$ languages. We translate our entire corpus in these $30$ languages (up to a maximum of $20$ million sentences) into English. This synthetic data is then filtered using the same round-trip and LangID filtering approach described in Section~\ref{sec:mt:filtering}. We also applied a few manual regex-based filters for specific languages where we observed particular data pollution and noise issues, as further elaborated in Section \ref{sec:errormagnification}.

\subsubsection{Distillation Approach and Hyper-parameters}
\label{sec:distill:arch}

We looked at two candidate student architectures with increasing encoder depth; we refer to them as \emph{shallow encoder} (330M parameters) and \emph{deep encoder} (850M parameters). All our models are sequence-to-sequence models with attention~\citep{bahdanau2015neural}, using Transformer encoders and LSTM~\citep{hochreiter1997long} decoders, as described in~\citet{chen2018best}. All our student models are multilingual, with separate models for \xxen and \enxx translation.

\textbf{Effect of student model capacity:} Although there are some quality improvements from optimizing hyperparameters, we found that most distilled models performed similarly, having little sensitivity to the hyperparameters we experimented with. Nonetheless, one important take-away is that trends that appeared to hold for the shallow encoder model, for instance the impact of increased amounts of back-translated data, were often erased when experimenting with the deeper model.

In all cases, the shallow encoder model was noticeably worse. For 30-language models, the deep encoder saw gains of about $+0.3$ and $+1.0$ \chrf{} in the \enxx and \xxen directions respectively. With respect to multilinguality, we found that increasing the multilinguality of the students from 6 languages to 30 yielded small quality losses of $0.1-0.3$ median \chrf.

\textbf{Effect of amounts of synthetic \enxx data used for distillation:} Another important hyperparameter was the amount of English-original synthetic data used for distillation (the non-English datasets were small enough that we could just translate the entire dataset). In the \enxx direction, where English-original data is forward-translated data, we varied the number of forward-translated sentences from 1M to 8M, but found no significant differences in model performance. In the \xxen direction, where English-original data is back-translated data, we saw consistent but small gains across all languages, with \chrf{} rising by about +0.6 when increasing from 1M to 2M synthetic sentences. Increasing past 2M back-translated sentences saw minimal gains. However, these experiments were carried out on the lower-capacity shallow encoder models (with 14 languages each), so more gains from higher quantities of back-translated data may be seen on a higher-capacity model. 

Although the teacher model may be trained with back-translation (BT), sequence-level distillation is typically only conducted with forward-translated data (FT). However, in this case there are some interesting implications that arise from a) the very small data sizes, and b) the asymmetrical teacher model quality, where \xxen quality tends to be better than \enxx quality. For \enxx, FT (aka English-original) data is more abundant, but also lower quality; for \xxen, the FT data is higher quality but rather scarce. Therefore, we experiment with using different proportions of FT and BT data for distillation.

For \enxx, we saw very small differences in performance between different ratios, and preferred using 80\% FT and 20\% BT data.

For an initial experiment on a 7-language shallow encoder model in the \xxen direction, we saw noticeable losses with under 50\% FT data, losing about 1.7 \chrf{} when going to 33\% FT, and losing a further 3.7 \chrf{} when going to 20\% FT. Values above 50\% FT were not significantly different. However, when observing the \chrf{} curve over time, we saw that the models with more BT data were learning more slowly and probably underfitting. Replicating this experiment with the deep encoder model, the performance on all languages increases, and the aggregate differences between different ratios of synthetic data are minimal. However, there is a slight trend on a per-language basis, with higher-resource languages benefiting from more back-translated data, with a Kendall tau of 0.28 between the number of monolingual training sentences and the difference between the 70\% FT and 20\% model.
Since the differences are very slight, we favor models with a smaller percentage of FT data (20\%-50\%), motivated by the intuition that the increased amount of natural target side English compared to the small number of natural source-side sentences may have benefits we can't measure on our eval sets and metrics.

\subsubsection{Comparison against teacher model}

A comparison of our best deep encoder student models against the teacher on all $30$ languages can be seen in Table \ref{tab:teacher_vs_student}. The student outperforms the teacher on the \enxx direction, with an average gain of +1.1 \chrf, and minor gains of +0.2 average \chrf{} on \xxen . These gains are probably in part due to the filtering applied to the RTT data. However, for the most part the differences are likely an artifact of the fact that the eval sets are English-Original, meaning that the English sentences are natural sentences, and the non-English sentences were produced via translation. Since distilled models tend to produce more translationese than their teachers, reference-based metrics like \chrf{} will tend to overestimate their performance in the source-original direction (\enxx{}), and underestimate it in the target-original direction (\xxen{}). This phenomenon is investigated in depth in \citet{freitag2019ape}.



\inserteachervsstudent

For the 30 languages in the distilled model, we looked for correlations between the amount of monolingual data, the number of speakers, the percentage of data removed by deduplication, the harmonic data quality score (Section \ref{sec:data:anomalousness}), and the \chrf{} score and human-rated scores in \enxx and \xxen directions. The correlations were all quite low, with Kendell's Tau under 0.3. The only exception to this is that there was a larger correlation ($\tau$ = 0.59) between \chrf{} in the \xxen direction and a heuristic measure of closeness to supervised languages (e.g. Bhojpuri/\texttt{bho}, being very close to Hindi, gets a score of 0.9; Aymara/\texttt{ay}, with some loanwords from Spanish but otherwise entirely unrelated, gets a score of 0.1).

\clearpage

\section{Evaluation}
\label{sec:eval}

Traditional metrics like \bleu, which have enough problems with higher-resource languages~\citep{freitag2019ape,freitag2020bleu}, have even more problems with the languages studied in the present work. Reference translations are hard to come by, and tail languages are often less standardized with respect to dialect, orthography, and even sometimes Unicode encoding. Furthermore, the frequent presence of close varieties complicates evaluation: automatic metrics like \chrf{} can give very high scores to outputs which are entirely in the wrong variety. Finally, as \citet{marchisio2021on} find, outputs of unsupervised NMT are less monotonic and more natural than outputs from supervised NMT, which, like the findings on paraphrased references from \citet{freitag2020bleu}, produce \bleu{} scores that are much lower --- although sometimes better correlated with human judgements of quality. Model-based metrics like \textsc{YiSi}~\citep{lo-2019-yisi}, \textsc{BLEURT}~\citep{sellam-etal-2020-bleurt} or \textsc{COMET}~\citep{rei-etal-2020-comet} cannot be used for these languages due to a lack of human ratings and pretrained models in these languages.

The following sections analyze performance along a variety of axes. First we describe the evaluation sets we collected (Section \ref{sec:evalsets}). We analyze the models starting with more quantitative methods, including the performance of \chrf{} versus \bleu{} (\ref{sec:bleuvschrf}) and human evaluations (\ref{sec:human}). We explore \rttlangidchrf{}, a reference-free metric supervised for very low-resource languages, which shows reasonable correlation with \chrf{}. We next perform qualitative analysis of our model outputs, and highlight several patterns of errors including confusing between distributionally similar words and concepts like ``tiger'' and ``miniature crocodile'' (\ref{sec:miniaturecrocodile}), errors on single word inputs (\ref{sec:errorshort}), and investigation of magnification of error modes in distilled models (\ref{sec:errormagnification}).

\subsection{Evaluation Sets}
\label{sec:evalsets}
In order to measure the quality of translation for development and experiments, we collected reference translations for 38 languages. For ease of comparison we collected a multi-way parallel data set, with the same English side for all languages. The \xxen eval sets were made by reversing this dataset. Rather than opting for larger evaluation sets for a small number of languages, we decided to collect relatively small evaluation sets containing $1200$ sentences in 38 linguistically and geographically diverse languages. $50$\% of the English sentences were drawn from a corpus of simpler and more colloquial language (average length: 12.0 tokens; $\sigma = 4.45$), and the remaining $50$\% from more technical web content (average length: 20.1 tokens; $\sigma = 11.6$). The resulting dataset was shuffled and then split into a 600-sentence development set and a 600-sentence test set. The tables in this paper report the score across the combined set except where otherwise mentioned. The average absolute difference between the \chrf{} scores on both sets was about 0.5.

\subsection{Evaluation Metrics}
\label{sec:bleuvschrf}
For this work we deem token-level metrics like \bleu{} unsuitable, and rely almost entirely on the character level \chrf{} \footnote{sacrebleu signature \texttt{nrefs:1|case:mixed|eff:yes|nc:6|nw:0|space:no|version:2.0.0}}~\citep{popovic2015chrf}, reported on a scale from 0 to 100. This section explains this decision, and gives examples of where the two metrics differ on the languages we studied.

Many of the languages studied in this work have complex morphologies, including the agglutinating Bantu languages and the polysynthetic Native American languages. This means that they can inflect to form very long tokens. An extreme example can be seen in Table \ref{tab:inflecting}, which shows how full sentences in English translate to only a few tokens in Kalaallisut (\texttt{kl}) . For such highly inflecting languages, it is to be expected that a character-based metric, like \chrf, would correlate better with quality than a token-based metric like \bleu. This expectation is borne out by our observations. These observations align with~\citet{mirzakhalov2021evaluating}, who observed similar trends for the agglutinating Turkic languages.

It is difficult to make any sort of direct comparison between \bleu{} and \chrf. Not only do they measure different quantities and have different score distributions (e.g. \chrf{} = 0 is exceedingly unlikely), but they also are influenced by different artifacts. The \bleu{} score will be affected by infection, and a slightly wrong infection will nullify all affected n-grams. On the other hand, \chrf{} is affected by the writing system -- for languages with abugidas or abjads, for instance, it will behave differently than for languages with alphabets, as there will be fewer characters to match. (On languages with writing systems like Chinese or Japanese, it will function very similarly to \bleu.) 

Nonetheless, in order to make some attempt to demonstrate how and where the two metrics differ, we have defined a very simplistic conversion metric based on performance of the less-inflecting low-resourced languages we studied: \scaledchrf{} = 0.75$*$\chrf{} - 0.15. The purpose of this metric is solely to have some way of flagging which languages lead to very different performance under the two metrics. Table \ref{tab:chrfvsbleu} shows the language pairs in our distilled models, along with their \bleu{} score, their \chrf{} score (unscaled) and the ratio between \scaledchrf{} and \bleu{}. The results correspond with intuitions: The languages with the highest ratio tend to be polysynthetic (Quechua/\texttt{qu}, Aymara/\texttt{ay}, Kalaallisut/\texttt{kl}), agglutinative (Luganda/\texttt{lg}, Lingala/\texttt{ln}), or otherwise highly fusional (Sanskrit/\texttt{sa} ). Oromo (\texttt{om}) is worth a special mention, as its orthography seems to have higher character usage per morpheme (because of the many doubled letters), which may inflate \chrf{}. It is not clear why Dhivehi (\texttt{dv}) has such a high ratio. The performance of both metrics on Tigrinya (\texttt{ti}) is also something of a mystery, given that the model translations were rated very highly by humans, and native speakers validated that the reference translations were also high quality.

\insertbleuvschrf
\insertinflecting

\subsection{RTT LangID ChrF}
\label{sec:rtteval}
Because of the infeasibility of collecting reference translations for 1000+ languages, some sort of reference-free evaluation is increasingly important. Round-trip translations have been utilized as a metric for MT Quality Evaluation several times over the last few decades~\citep{huang1990machine,aiken2010efficacy,moon2020revisiting}. We experimented with a simple modification we call \rttlangidchrf{}. To compute this score, we simply round-trip-translate a corpus of English sentences through some language, and compute the \chrf{} score of these translations with respect to the original sentences. However, since this metric is trivially fooled by error modes like copying or translating into the wrong language, we omit any round-trip translation where the intermediate translation is not in the correct language according to our SSLID model. If fewer than 10\% of the intermediate translations receive the correct LangID score, we consider the score invalid, as this may also be a result of errors in the LangID model.

We computed correlation between \chrf{} and \rttlangidchrf{} over 30 languages, and found it to correlate moderately well both in the \enxx direction ($\rho=0.46, \tau=0.34$) and the \xxen direction ($\rho=0.30, \tau=0.28$). When we recalculated only on the scores from the distilled models, the correlations were much better in the \xxen direction ($\rho=0.80, \tau=0.69$), but similar in the \enxx direction ($\rho=0.36, \tau=0.22$).

\begin{figure}[t!]
\begin{center}
\includegraphics[scale=0.315]{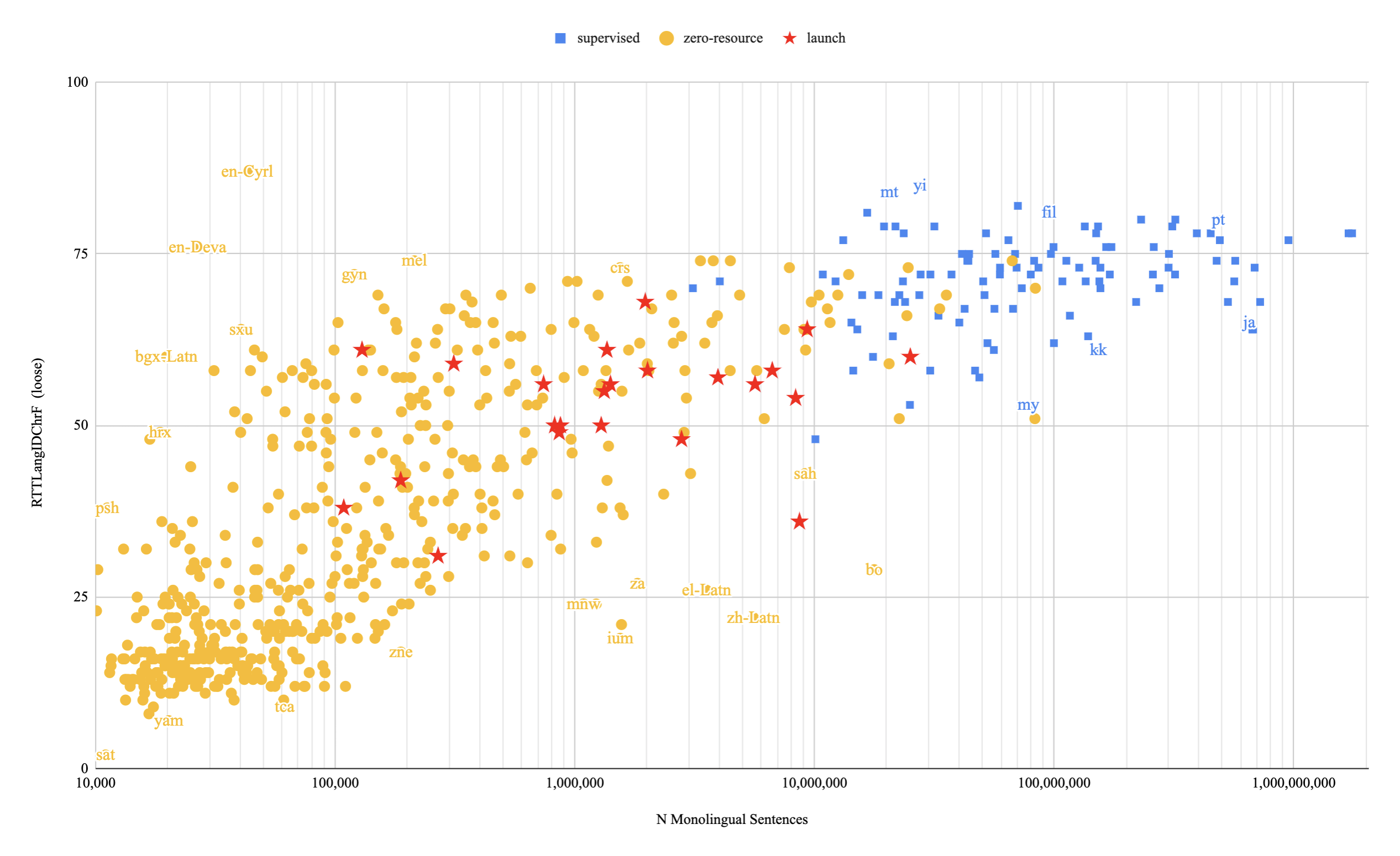}
\caption{Plot of \rttlangidchrf{} scores (loose) for languages as a function of log monolingual data size. With over 100,000 sentences, almost any language does reasonably well. Outliers are labeled with their language code. The largest outliers are English in Cyrillic script (\texttt{en-Cyrl}), which has excellent \rttlangidchrf{} score but very little monolingual data, and Tibetan (\texttt{bo}), which has plenty monolingual text but very poor performance. In general, the languages above the trend line are close to high-resource languages (where the metric may also be fooled), and the languages below the trend line are linguistically distant from other languages in the model or have poor-quality data. Languages added to Google Translate as part of this effort (all unsupervised except Sorani Kurdish (\texttt{ckb})) are marked with stars.}
\label{fig:rttlangidchrf}
\end{center}
\end{figure}

In addition to this, we computed this score over all languages in the 1000-language model, of which 630 passed the LangID > 10\% threshold. Figure \ref{fig:rttlangidchrf} shows these scores as a function of log data size. There is a relatively clear trend ($\rho = 0.80, \tau=0.63$), and the large majority of languages with over 100,000 monolingual sentences have relatively high scores.  In general, the languages above the trend line are close dialects to high-resource languages, most notably variants of English written in different scripts. Languages below the trend line tend to be unrelated to high-resource languages or have poor-quality data according to our data audit.

Of the 630 languages with valid \rttlangidchrf{} scores, 268 have a \rttlangidchrf{} score of over 30.0, which we tentatively deem of ``hopeful quality'', and 147 have \rttlangidchrf{} $\ge 50$, which is the minimum score from any of our supervised languages. Interestingly enough there is actually fairly low correlation between \rttlangidchrf{} and the percent of intermediate translations that were assigned the correct LangID score ($\rho=0.18, \tau=0.10$). One possible explanation is that frequent error modes are outputs in the wrong language and copying, phenomena that we observed in Section \ref{sec:errormagnification}.

We include \rttlangidchrf{} for all 630 languages in Appendix Table \ref{tab:datasize_full}, as an approximate measure of translation quality of the model. However, since we have done only cursory analysis of how effective this score is at measuring quality, we advise readers to view it with appropriate skepticism. While a low \rttlangidchrf{} probably means low translation quality, a high value may well mean something other than high quality.

The version of this score described above we call the \textit{loose} version of \rttlangidchrf{}, since it does not penalize the model for intermediate sentences in the wrong language. To get the \textit{strict} \rttlangidchrf{}, we multiply the loose \rttlangidchrf{} by the percent of intermediate translations assigned the correct LangID score, thereby penalizing wrong-language translations. This version does not correlate well with \chrf{} on the 30 languages where we have evaluation sets, and in fact has negative correlation in the \xxen direction ($\rho=-0.28, \tau=-0.20$), and only a weak correlation in the \enxx direction ($\rho=0.12, \tau=0.09$), likely due to noise from the LangID model. It also has weaker correlation with the size of the monolingual dataset over all 630 applicable languages ($\rho=0.43, \tau=0.55$), and is noticeably messed up by close dialects, e.g. assigning Bosnian (\texttt{bs}) a low score because intermediate translations were frequently LangID'd as Croatian (\texttt{hr}), a mistake that should not be penalized as the two languages are frequently indistinguishable. Despite these failings, it has the attractive property that it penalizes the common error mode of wrong-language outputs; and furthermore, the supervised and zero-shot languages appear more clearly separated when plotting them versus monolingual data size. The graph is therefore included in Appendix \ref{appendix:rtteval}.

\subsection{Human evaluations}
\label{sec:human}

Any decision on translation quality ultimately cannot by made with automatic metrics like \chrf{} alone. In order to understand the quality of our distilled models, we asked human raters to rate the quality of the translations from our test set on a scale from 0 (nonsense or wrong language) to 6 (perfect). Full results may be seen in Appendix Table \ref{tab:distilledperf}.

Although we made an attempt to calibrate raters and explain each point in the scale very clearly, each rater will naturally have a different understanding of ``a good translation''. For this reason, it is very difficult to interpret these results in any sort of holistic way. However, with some diving into the results together with native speakers, a few things stood out.

The biggest takeaway is that automatic metrics overestimate performance on related dialects. Nigerian Pidgin (\texttt{pcm}), a dialect of English, had very high \bleu{} and \chrf{} scores, of around 35 and 60 respectively. However, humans rated the translations very harshly, with a full 20\% judged as ``Nonsense/Wrong Language'', with trusted native speakers confirming that the translations were unusable. Krio (\texttt{kri} -- close to English), Maithili (\texttt{mai} --close to Hindi), and Bhojpuri (\texttt{bho} --close to Hindi) were in a similar boat, though trusted native speakers agreed that the translation quality, though borderline, was usable. What's happening here that the model translates into (a corrupted version of ) the wrong dialect, but it is close enough on a character n-gram level that the \chrf{} is still high. In our cases, this is the result of a data pollution problem. Since these languages are so close to other much more common languages on the web -- in this case, English and Hindi -- the training data is much more likely to be mixed with either corrupted versions of the higher-resource language, or other varieties. As a result, many model outputs that were supposed to be in Dogri (\texttt{doi}) were actually in misspelled or ungrammatical Hindi (\texttt{hi}), outputs supposed to be in Nigerian Pidgin (\texttt{pcm}) were sometimes in other English dialects like AAVE, and so on.

\subsection{Mistakes on distributionally similar words}
\label{sec:miniaturecrocodile}
\insertunmterrorsmain

We observe that our zero-resource models exhibit some characteristic error modes. The most common one relates to translating nouns that occur in distributionally similar contexts in the training data. This occurs even for relatively common nouns like ``tiger'' -- which is often translated as another kind of animal, showing that the model learned the distributional context in which this noun occurs, but was unable to acquire the exact mappings from one language to another with enough detail within this category. This may be related to the relatively small amounts of training data that were used, alongside the unsupervised nature of training. Common nouns suffering from these mistakes include animal names, colors, and times of day. This was also an issue with adjectives, but we observed few such errors with verbs. Sometimes, words were translated into sentences that might be considered culturally analogous concepts -- for example, translating ``cheese and butter'' into ``curd and yogurt'' when translating from Sanskrit (\texttt{sa}). Surprisingly, the model hypotheses were often strings that would probably yield a high perplexity under most language models. Table~\ref{tab:unmt_mistakes_full} provides a variety of examples of mistakes from these models. These examples are from the 6B parameter 1000-language Transformer model, described in Section~\ref{sec:mt:multilinguality}. 

A good example of what it means for words to have different meanings but to be distributionally similar given the usage of the language is the translation of the string ``English Language''. Around 25\% of the languages translated the string ``English Language'' into the name of their \textit{own} language, e.g. into Tsonga as \textit{ririmi ra xitsonga} or into Oromo as \textit{Afaan Oromoo}.

Another interesting phenomenon evident from Table~\ref{tab:unmt_mistakes_full} is that models seem to be good at translating ``red'', mediocre at translating ``yellow'', and poor at translating ``orange''. This observation is consistent with cross-lingual hierarchy of color terms described in \citet{berlin1969basic, saunders2002the}, which find that terms for color generally arise in specific orders across the world's cultures, with words for ``red'' occurring in stage II of color development, ``yellow'' in stage III/IV, and ``orange'' in stage VII.

An interesting parallel can be found between this problem and an issue of ``unsupervised'' translation in the real world -- the case of the words for \textit{four} and \textit{six} in Etruscan, a language that went extinct about 2,000 years ago \citep{etruscan}. There does not currently exist any surviving parallel text with Etruscan, excepting a single 37-word tablet, but there do exist some 13,000 monolingual inscriptions. Etruscan does not seem obviously related to any living languages, so meanings usually cannot be discovered by their similarity to cognates. As a result, modern scholars have to use a sort of ``unsupervied translation'' called the \textit{Combinatorial Method}, a first-principles approach to discovering word sense and grammar. And in the case of Etruscan, there are some words whose meaning cannot be teased apart with existing context. As an example, the words for all numerals from one to ten have been established with some certainty --  with the exception of \textit{four} and \textit{six}. For these two there is no clear contextual clue to separate them, and there is no academic consensus of which of \textit{hu}$\theta$ and \textit{{\'s}a} mean \textit{four} and \textit{six} \citep{artioli2011gambling}.

\subsection{Performance on tokens by token frequency}
\label{sec:tokfreq}

In order to quantitatively measure the distributional token errors from Section \ref{sec:miniaturecrocodile}, we decided to look at the accuracy of our model at translating particular tokens as a function of their frequency in an open-domain English web corpus.

First, we collected a list of the 8000 most common tokens in a large, web-crawled, monolingual English corpus. Then, we separated these into exponentially-spaced bins, based on the exponential distribution of token frequencies in natural text \citep{zipf}. Therefore, each bin corresponds to a set of English tokens in a certain frequency band. Then, for each of these bins, we made a evaluation set for this bin composed of all sentences in our standard \xxen evaluation set containing these target tokens at least once in their references.

To score a model on a bin-specific evaluation set, we looked at a simple hit-rate metric. For a given reference sentence containing $k$ tokens in the set of target sentence, the model got one point for each token it produced in its output that was among these $k$ tokens (a ``hit''), for a maximum of $k$ points. The hit-rate score for that eval set is then simply the number of hits the model got divided by the total number of possible hits. Formally, for a given set of tokens $\mathcal{B}$ (in our case a frequency bin), a list of reference translations $r_i \in R$ and a corresponding list of model hypotheses $h_i \in H$, the hit-rate $\Xi(H, R, \mathcal{B})$ is defined as follows:

$$\Xi(H, R, \mathcal{B}) = \frac{ \sum_i \sum_{t \in r_i} \mathbbm{1} \big\{ t \in h_i, t \in \mathcal{B} \big\} }{\sum_i \sum_{t \in r_i} \mathbbm{1} \big\{ t \in r_i, t \in \mathcal{B} \big\}}$$

Table \ref{table:grounding_tokens} shows the result of this analysis, sorted by the number of monolingual sentences seen by the model. Hit-rate is reported on five bins, starting on the most frequent tokens (tokens \#0 - \#125) and ending with the most infrequent tokens (tokens \#8000 - \#12800). For each bin the number of sentences in this eval set is included in the column labeled \texttt{sents}. The most frequent bin includes 1191 of the 1197 sentences in the full eval set; the lest frequent bin includes only 958 of them. For each bin the total number of possible points (i.e. hits) is also reported in the column labeled \texttt{pts}. Whereas the first bin has a total of 9682 possible points -- averaging eight per sentence, and constituting mostly function words -- the least frequent bin had only 2420 possible points, so slightly above two per sentence. It is also worth mentioning that the least frequent bin had 6,000 possible tokens in it, so only about 30\% of them actually occurred in our eval set.

\inserttablegroundingtokens

For many languages, higher \chrf{} means higher hit-rate across the board, and for others, like K'iche' (\texttt{quc}) low \chrf{} corresponds with generally lower hit-rate. Things get more interesting for languages that have higher hit-rates for the first several bins of higher frequency tokens, but lower hit-rates for less common tokens. These are the ones that have higher \chrf{} scores, corresponding to the ability to translate the top 500 tokens or so very well, but that also make frequent mistakes on less common tokens. Perhaps the best example of this is Fulfulde (\texttt{ff}), which had the relatively high \chrf{} of 40.7, but sees a large drop-off in hit-rate from 41\% on bin 2 (token \#125 - \#500) to 26\% on bin 4  (token \#2000 - \#8000). This is also the language where, anecdotally, we observed several of these sorts of mistakes.  Mizo (\texttt{lus}) and Bambara (\texttt{bm}) exhibit similar patterns.

\inserterrorssinglewordsshort

\subsection{Errors on short inputs}
\label{sec:errorshort}

Another category of errors we encountered was with single word inputs to the model. The translations tended to be longer, and the model would frequently give alternate translations or append frequent tokens (Table~\ref{tab:errors_singlewords_short}). Outputs also frequently had duplicates among the other outputs, suggesting hallucination, or copied inputs. A breakdown of types of errors can be found in Table \ref{tab:singlewordserrorcats}. 

This was an issue we observed mainly for lower-resource languages. For higher-resource languages, like Ilocano, the model tended to provide a succinct and correct single-word translation. Ensuring that the MASS training data covers the lower end of the length distribution would likely remedy these issues. However, this is also an inherently difficult problem since we do not provide the model any source language information. For the \xxen direction, the translation task needs to solve LangID and translation simultaneously, which could often be ambiguous and quite challenging for short queries. This can potentially be addressed by providing the model source language information along with the input, and is left to future work.

\insertsinglewordserrorcats

\subsection{Magnification of error modes in student models}
\label{sec:errormagnification}

After training distilled models, we noticed a variety of unexpected error modes when analyzing the output translations:

\begin{itemize}
    \item The translations to Nigerian Pidgin (\texttt{pcm}) frequently instead translated to (often offensive) US slang. For instance, the English sentence \textit{``She said to herself''} was translated to the unacceptable string \textit{``da b***** say ta da b*****self.''}
    \item Many of the translations to Kalaallisut (\texttt{kl})  were actually Danish  (\texttt{da}) 
    \item Many translations to Sanskrit  (\texttt{asa})  were actually Hindi  (\texttt{hi}) 
\end{itemize}

We developed filters to remove this content from the forward translated data and distilled the models again. We observed that these problems were more prevalent in the synthetic data used for distillation (generated by the teacher model) than in the monolingual data that had originally been used to train these models, and that the issues were more severe for synthetic text produced by translating a noisier source corpus. The changes in noise level are illustrated in Table \ref{tab:error_magnification}. We hypothesize that this error magnification could either be an artifact of a positive feedback loop arising from training the model on its own prediction (self-training), or due to a difference in domains between the training and distillation datasets.

This table also includes one entry about distillation from Japanese to English. In this case, we found that one particular Amharic string was often hallucinated. This string occurred occasionally in the target side of the original training data, and then occurred much more often in the distilled data.

\insertmagnification

\subsection{Comparison on Flores benchmark}
\label{sec:flores}

Since these models are not trained on the same data as public benchmarks, a comparison on public benchmarks is not necessarily very meaningful. Nonetheless, in Table \ref{tab:flores} we provide a comparison between the \textsc{spBLEU} results from our method (on distilled models) versus the Flores-101 benchmark scores~\citep{goyal2021flores} reported for the massively multilingual M2M-124~\citep{fan2021beyond} for overlapping languages. Given the higher language coverage in our monolingual dataset, our models yield higher \textsc{spBLEU} for all language pairs.

\begin{table*}[t]
\centering
\small
\begin{tabular}{c|cc||c|cc}
\hline
\hline
LP & this method  & M2M-100 & LP & this method  & M2M-100    \\
\hline

en$\rightarrow$as &\textbf{29.1} & 1.22 & as$\rightarrow$en & \textbf{34.5} & 3.76\\
en$\rightarrow$ckb & \textbf{28.5} & 0.23 & ckb$\rightarrow$en & \textbf{37.5} & 7.65 \\
en$\rightarrow$ff & \textbf{2.5} & 0.68 & ff$\rightarrow$en & \textbf{11.2} & 2.4 \\
en$\rightarrow$lg & \textbf{17.3} & 0.61 & lg$\rightarrow$en & \textbf{29.3} & 4.45\\
en$\rightarrow$ln & \textbf{24.7} & 1.03 & ln$\rightarrow$en & \textbf{30.2} & 4.57\\
en$\rightarrow$nso & \textbf{32.5} & 1.54 & nso$\rightarrow$en & \textbf{45.0} & 6.76\\
en$\rightarrow$om & \textbf{17.2} & 0.4 & om$\rightarrow$en & \textbf{30.8} & 3.33\\
\hline
\end{tabular}
\caption{Flores dev-test: comparing \textsc{spBleu} between this method and \citet{goyal2021flores}.
 \label{tab:flores}}
\end{table*}

\clearpage
\section{Additional Experiments and Notes}
\label{sec:additional}

\subsection{Non-English-centric bridging}
\label{sec:direct}

There is nothing inherently English-centric about the approach to zero-resource translation put forth in this paper. Nonetheless, the model has only seen translated text between English and other languages, so it would be a reasonable hypothesis that it is just inherently better at translating to and from English, even in the zero-shot scenario. 

We test this hypothesis by evaluating the model on bridged translations. We first translate the English source sentences to other languages using bilingual supervised models on these language pairs. Then we use the model proposed in this paper to translate these translations directly into the desired target language. For each desired target language we pick 1) the closest mid- or high-resource language (HRL) that we expect our bilingual models to do well on; and 2) if applicable, a lower-resource language (LRL) that may be closer to the desired target language, but have lower quality supervised models. Please note that the definition of close is somewhat approximate. For instance, for Native American languages we choose colonial languages as the ``close" languages, because they may share a certain amount of vocabulary, even if the grammar is entirely divergent. Furthermore, for some of these languages the ``close'' languages are in fact not very close at all, as with the Sino-Tibetan languages of North-East India (Bodo (\texttt{brx}), Meiteilon (\texttt{mni-Mtei}), Mizo (\texttt{lus})), which are only somewhat related to the ``closer LRL'' of Myanmar/Burmese (\texttt{my}).

\begin{table}[htb]
    \begin{subtable}[t]{.5\textwidth}
        \raggedright
            \begin{tabular}{l|l|ll|ll}
\hline
 lang. & direct & \multicolumn{2}{c}{close HRL} & \multicolumn{2}{|c}{closer LRL} \\
 
\hline
\multicolumn{6}{c}{}\\
\multicolumn{6}{c}{Native American Languages}\\
\hline
 ay & 33.1 & es & \textbf{34.2} & - &  - \\
 gn & \textbf{31.5} & es & 28.9 & - &  - \\
 kl & \textbf{35.5} & da & 27.5 & - &  - \\
 qu & \textbf{35.3} & es & 29.5 & - &  - \\
 quc & \textbf{24.1} & es & 22.5 & - &  - \\
 yua & \textbf{31.5} & es & 28.4 & - &  - \\
 
\multicolumn{6}{c}{}\\
\multicolumn{6}{c}{Indian Languages (Indo-European)}\\
\hline
 as & 39.2 & hi & \textbf{39.8} & bn & 36.3 \\
 bho & 42.0 & hi & \textbf{43.4} & - &  - \\
 doi & 36.3 & hi & \textbf{39.5} & pa & 33.3 \\
 dv & \textbf{44.4} & hi & 42.0 & si & 39.7 \\
 gom & \textbf{40.2} & hi & 39.3 & mr & 39.7 \\
 ks & 21.9 & hi & 25.7 & ur & \textbf{28.5} \\
 mai & 38.1 & hi & \textbf{44.3 }& - &  - \\
 sa & \textbf{30.5} & hi & 27.3 & - &  - \\
 \multicolumn{6}{c}{}\\
 \multicolumn{6}{c}{}\\
            \end{tabular}
    \end{subtable}%
   \begin{subtable}[t]{.5\textwidth}
        \raggedleft
        \begin{tabular}{l|l|ll|ll}
        \hline
         lang. & direct & \multicolumn{2}{c}{close HRL} & \multicolumn{2}{|c}{closer LRL} \\
 
\hline
\multicolumn{6}{c}{}\\
\multicolumn{6}{c}{Indian Languages (Not Indo-European)}\\
\hline
 brx-Beng & 4.6 &  hi & \textbf{11.5} & my & 3.4 \\
 lus & \textbf{38.6} &      hi & \textbf{38.2} & my & 34.1 \\
 mni & \textbf{40.7} &      hi & 35.5 & my & 29.4 \\
 sat-Latn & \textbf{20.9} & hi & \textbf{20.8} & km & 18.2 \\
 
\multicolumn{6}{c}{}\\
\multicolumn{6}{c}{Bantu Languages}\\
\hline
 lg & \textbf{38.7} &  sw & 34.1 & rw & 33.6 \\
 ln & \textbf{34.4} &  sw & 31.9 & fr & 25.5 \\
 nso & \textbf{45.7} & sw & 33.3 & st & 29.1 \\
 ts & \textbf{46.2} &  sw & 40.0 & zu & 44.0 \\
 
\multicolumn{6}{c}{}\\
\multicolumn{6}{c}{Other}\\
\hline
 bm & \textbf{34.3} & fr & 26.6 & - &  - \\
 ff & \textbf{34.7} & sw & 26.7 & - &  - \\
 ilo & \textbf{54.0} & id & 48.4 & fil & 51.2 \\
 om & \textbf{39.2} & - &  - & so & 30.2 \\
 ti & 21.4 & - &  - & am & \textbf{22.6} \\
 yue & 20.3 & zh & \textbf{21.6} & - &  - \\
        \end{tabular}
    \end{subtable}
    
\caption{Results for bridged translation from English (\chrf{}) on 1200 sentences/language. Bridging seems to improve quality only when the intermediate language is both higher-resource and close to the target language.} 
\label{tab:direct}
\end{table}

Table~\ref{tab:direct} shows the results of this investigation, along with which languages were chosen as ``close high-resource languages''\footnote{For this experiment, we considered Swahili  (\texttt{sw}) to be a high resource language. This is not precisely accurate, but it was the closest mid-resource language available.} and ``closer low-resource languages''.

We find that in a substantial number of cases, bridged translation scores better on automatic metrics. This is true especially for the languages of India, with bridged translation drawing even or improving even for the non-Indo-European languages. Overall, the largest improvements are seen on close dialects, for instance Maithili (\texttt{mai}: +6.2 \chrf{}), Bhojpuri (\texttt{bho}: +1.4 \chrf{}), Kashmiri (\texttt{ks}: +7.4 \chrf{}), and Cantonese (\texttt{yue}: +1.5 \chrf{}). Aymara (\texttt{ay}) and Dogri (\texttt{doi})  also saw noticeable gains. However, most Native American languages, Bantu languages, and languages without close relatives saw large losses by bridging.

Nonetheless, in 19 out of 28 cases, we find that direct translation from English produces the highest \chrf{} score. This may be due in part to the model being inherently better at translating from English, and in part from errors compounding from the two-step process. Evidence for the second hypothesis may be seen in the fact that only two languages (Kashmiri (\texttt{ks}) and Tigrinya (\texttt{ti})) see any gain from bridging through a lower-resource language. Overall, it seems that bridging only improves quality when the intermediate model is already relatively high quality, and additionally when the intermediate language is close to the target language.

The fact that bridging appears to work relatively well is a good sign. In practice, translation to and from English is not the major use case for many low resource languages, and direct models to local languages (e.g. Hindi for India, Spanish for Latin America, etc.) would likely provide more utility to local communities.

\subsection{Zero-shot transliteration}
\label{sec:translit}

Many of the world's languages are written in multiple scripts, whether because of historical and national changes, informal online usage, non-standardized writing systems, or different ethnic or religious populations. For this reason, especially with under-resourced languages where this is more common, it is especially important to be able to support these languages in their many different writing systems.

For our data collection efforts, we crawled data in a variety of different scripts for several languages, e.g. both Malayalam in Malayalam script (\texttt{ml}) and Malayalam in Latin script (\texttt{ml-Latn}). We treated script-variants of the same language as separate languages, with their distinct \texttt{<2xx>} tags. To transliterate, we simply asked the model to provide ``translations'' from text in one script to the same language in another script.

We applied this approach to transliterate from Latinized variants of Indian languages to their native scripts, and found that it worked very well out of the box, appearing to be more robust than existing transliteration libraries to informal or nonstandard spellings. One example is the common abbreviation \textit{kr} in Latinized Hindi (\texttt{hi-Latn}), as in \textit{kya kr rhe ho} which our model correctly transliterated as ``kar'' (in Devanagari script), whereas the rule-based model incorrectly rendered as ``ke''. Similarly the model was able to do some spelling correction, e.g. on the the misspelled Konkani \textit{swpnatat}, which the model correctly transliterated to ``swapnaat''. However, the model also had a tendency to change small parts of the input sentence, as well as occasionally hallucinating extra content. This also made it difficult to compare this model to rule-based approaches in a rigorous way, because the lack of guaranteed monotonic alignment made word error rate inapplicable.

One interesting example highlighting issues this approach has is the Konkani sentence \textit{Xet-camot ani kneddeam- gauncho vaur vo dondo.} This was transliterated mostly correctly to Devanagari, but the the dialect was changed: the original text is Goan Roman Catholic Konkani, but the transliteration was in Goan or Maharashtrian Konkani, changing ``camot'' to ``camat'' and ``vo dondo'' to ``vaa dando''.

Future work is necessary to coax this technique into a form that does not take any liberties with the input. One promising direction is to separate the \texttt{<2xx>} tokens~\citep{johnson2016google} into a language subtag and a script subtag. For instance, \texttt{<2ms>} becomes \texttt{<2ms> <2Latn>}, and \texttt{<2ms-Arab>} becomes \texttt{<2ms> <2Arab>}. This enables 0-shot transliteration between any scripts supported by the model, simply by using the desired \texttt{<2Script>} tag at inference time. In order to prevent the model editorializing, it would also be advisable to consider adding synthetic transliterated parallel text with a \texttt{<2transliteration>} tag, to teach the model that this task only involves substituting letters/sounds.

\subsection{The ``Period Trick''}
\label{sec:period}
Even after having been distilled on a mix of clean and noisy data, we observed that some of these languages still had lower performance on inputs that lacked terminal punctuations. To study this, we compared performance on the evaluation sets with and without terminal punctuation. Table \ref{tab:period} illustrates the results of this experiment on the distilled models, though we noticed similar trends in the teacher models. The gain is small but consistent, and in the \xxen direction, 100\% of the language pairs benefit. We noticed that sentences without terminal punctuation sometimes triggered common error modes, e.g. decoding into Danish  (\texttt{da}) instead of Kalaallisut (\texttt{kl}), or misspelled Hindi  (\texttt{hi})  instead of Dogri  (\texttt{doi}). We hypothesize that the presence of terminal punctuations might provide a ``domain'' signal to the model and trigger different translation qualities.

\insertperiod

\subsection{Robustness to non-standard glyph usage} %
\label{sec:glyphs}

There are many different ways to write certain letters, especially those where a Unicode standard was introduced after a population was already active online, or before keyboards using this standard were widely available. Common cases include the many Unicode points for the ``open o'' (\AS{\m{o}}) and ``open e'' (\AS{\m{e}}) used in many African languages; the Palochka (resembling the letter I), used in many Caucasian languages; the apostrophe or `Okina, used around the world but especially in many American languages; and many other examples that can be found in the Unicode Confusables list~\citep{davis:2021}. Table \ref{tab:chechen} gives an example of this phenomenon in the wild, showing the variation of Unicode points used for the Chechen and Chuvash languages in our webcrawled data. We refer the reader to \citet{prasad2018mining} for many more examples of this phenomenon. 

We conducted a simple experiment to determine how robust our model was to these different usages. Looking at translations into English, we decoded our evaluation sets (with the non-finetuned, 1000-language teacher model) using each of a variety of different ways of representing each letter. We then compared the \chrf{} scores on the output. Results can be seen in Table \ref{tab:glyph_bleu}. We found that in most cases, this made very little difference in \chrf, even when using rare glyphs like the capital Greek Iota in place of the Palochka in Caucasian languages, indicating that our models were quite robust to these variations. However, we did notice a relatively large change in \chrf{} for West African languages when using nonstandard glyphs, including the ``chatspeak'' ASCII characters used when texting or writing very informally -- for instance, writing ``\AS{aho)f3}'' for ``\AS{aho\m{o}f\m{e}}''.

\begin{table*}[ht]
\small
\centering
\begin{tabular}{lr}
Unicode character name &	Percent of Data \\
\hline
LATIN CAPITAL LETTER I &	33.6\% \\
DIGIT ONE &	30.4\% \\
\textbf{CYRILLIC LETTER PALOCHKA} &	3.8\% \\
CYRILLIC CAPITAL LETTER BYELORUSSIAN-UKRAINIAN I &	4.2\% \\
LATIN SMALL LETTER L &	2.8\% \\
GREEK CAPITAL LETTER IOTA &	0.4\% \\
\hdashline
LATIN SMALL LETTER A WITH BREVE	&	52.4\% \\
LATIN SMALL LETTER E WITH BREVE	&	55.0\% \\
LATIN SMALL LETTER C WITH CEDILLA	&	43.4\% \\
\textbf{CYRILLIC SMALL LETTER IE WITH BREVE}	&	10.2\% \\
\textbf{CYRILLIC SMALL LETTER A WITH BREVE}	&	9.4\% \\
\textbf{CYRILLIC SMALL LETTER ES WITH DESCENDER}	&	8.8\% \\
\end{tabular}
\caption{Examples of the different Unicode points used to encode the Palochka character in the Chechen language (above) and letters with diacritics in the Chuvash language (below), along with their prevalence on web-crawled data. In both cases, the ``correct'' Unicode point (\textbf{bolded}) is much less common.
 \label{tab:chechen}}
\end{table*}

\begin{table*}[ht]
\small
\centering
\begin{tabular}{llr}
Unicode point & 	name & 	avg. \chrf{} \\ 
\hline		
\multicolumn{3}{c}{}\\		
\multicolumn{3}{c}{gn, luo, quc, yua $\rightarrow$ en}\\		
\hdashline		
U+2019 & 	Right quote & 	25.7 \\ 
U+0060 & 	Grave accent & 	25.5 \\ 
U+0027 & 	Apostrophe & 	25.5 \\ 
U+02BB & 	'Okina & 	25.5 \\ 
U+00B4 & 	Acute accent & 	25.4 \\ 
 & 	None & 	24.5 \\ 
 
\multicolumn{3}{c}{}\\		
\multicolumn{3}{c}{ady, av, ce $\rightarrow$ en}\\		
\hdashline		
U+04C0 & 	Palochka & 	49.6 \\ 
U+0031 & 	ASCII 1 & 	49.4 \\ 
U+0399 & 	Greek Iota & 	49.4 \\ 
U+0406 & 	Byelorusian/Ukrainian I & 	49.3 \\ 
		
\multicolumn{3}{c}{}\\		
\multicolumn{3}{c}{ak, bm, dyu, ee $\rightarrow$ en}\\		
\hdashline		
U+025B, U+0254 & 	Latin open e/o & 	33.6 \\ 
U+03B5, U+1D10 & 	Greek epsilon; small capital O & 	25.8 \\ 
U+0033, U+0029 & 	3 and ) (chatspeak) & 	25.3 \\ 
\multicolumn{3}{c}{}\\		
\multicolumn{3}{c}{cv $\rightarrow$ en}\\		
\hdashline		
U+04D7, U+04D1, U+04AB & 	Cyrillic codepoints & 	42.3 \\ 
U+0115, U+0103, U+00E7 & 	Latin Codepoints & 	43.0 \\ 
 \end{tabular}
\caption{\chrf{} score where the source uses different versions of common Unicode points. The top line of each bloc represents the ``correct'' codepoints, whereas the lower lines are other ways of representing the same letter. In many cases there is very little difference in performance, but the African languages are affected by nonstandard \AS{\m{o}} and \AS{\m{e}}. When the apostrophe is removed entirely, the performance also drops noticeably.
 \label{tab:glyph_bleu}}
\end{table*}

\subsection{Non-Unicode fonts}
\label{sec:nonuni}

Before easy access to keyboards using the correct Unicode points, or before the Unicode standard itself, it was often not clear how to represent alphabets or characters that were not in the ASCII range. We already saw one consequence of this in Section \ref{sec:glyphs}. A more difficult consequence is the existence of non-Unicode fonts. The way this often works is that one types in ASCII characters, but downloads a special font that provides glyphs for these the ASCII code points that give the desired visual rendering -- for instance, one types ``l72is4wo'', and it renders as the Ewe word \AS{l{\~a}\M{d}is\m{o}wo}. Other non-Unicode fonts even assign different values to code points beyond the ASCSII block: one well-known example is the Zawgyi encoding for Myanmar (Burmese)~\citep{liao:2017}. In the course of this work, we discovered that a wide variety of languages still use non-Unicode fonts. We ran into such fonts for Ewe (\texttt{ee}), Kashmiri (\texttt{ks}), Meiteilon (\texttt{mni-Mtei}), Moor{\'e} (\texttt{mos}), Navajo (\texttt{nv}), and Tamazight (\texttt{ber-Latn}), usually in the case of deliveries from professional translators. It is likely that there exists a large, hidden portion of data for some languages using these or other fonts. Our attempted reconstruction of some of these fonts can be seen in Appendix Section \ref{appendix:nonuni}.

\clearpage

\section{Importance of Native Speakers}
\label{sec:nativespeakers}
Since the advent of statistical and then neural machine translation, large data sets and improved modeling techniques have driven progress in translation quality. It has sometimes been difficult for the expertise of native speakers and linguists to continue helping to improve models in this environment. However, the participation of speakers of languages, and of members of affected communities, is nonetheless still vital. In the very-low-resource domain, there are more errors in models and data -- and consequently, more opportunities for native speakers to help improve the quality.

We stress that where possible, it is important to try to build relationships with native speakers and members of these communities, rather than simply interacting with them as crowd-workers at a distance. For this work, the authors reached out to members of as many communities as we could, having conversations with over 100 members of these communities, many of whom were active in this project (See Acknowledgements).

Here is an incomplete list of ways in which speakers of these languages helped us extend and improve machine translation to their languages:

\begin{enumerate}
    \item \textbf{Understanding Data:} As in \citet{kreutzer2022quality}, we conducted an extensive review of the quality of our datasets (see Appendix Table \ref{tab:audit}). In addition to simply giving us an idea of which languages had higher or lower quality data, this also gave us valuable insights about other uses and aspects of the corpora that were useful beyond this project -- for instance, which corpora had more colloquial (or religious) text, and which dialects were mixed with other languages.
    \item \textbf{Understanding errors in reference translations:} A variety of languages had extremely low automatic metrics (e.g. \bleu{} under 1.0), despite having large and clean corpora. For two such cases, native speakers helped us identify quality and fluency issues with our reference translations -- and that the model outputs often looked better than the references. 
    \item \textbf{Specialized Filters:} Our corpora for a few languages were polluted with related high-resource languages that had somehow passed all the previous rounds of filtering. For two cases, natives speakers helped design custom filters to remove the unwanted content, and for a third, they helped remove sensitive content. 
    \item \textbf{Transliteration and political sensitivity around script:} For one language, we were initially unaware that we were using the wrong script. We were using a script that was associated with colonial times, and had since been replaced in the entire region, and was a matter of political sensitivity. Native speakers both pointed out this issue and helped us transliterate the text to the appropriate script. (See Appendix Section \ref{sec:mtei}) 
    \item \textbf{Understanding crowd-worker annotations:} when we sent translations to crowd workers to rate their quality, several languages showed unusual rating patterns, or patterns that did not line up with our expectations. We were fortunate to have native speakers who helped us understand the ratings differentiate cases where raters were mis-calibrated against real quality issues. \citet{freitag-etal-2021-experts} explores some of these phenomena more. 
    \item \textbf{Clarifying utility for Community:} Even if one \textit{can} build a translation model for a language, should one? 
    For some groups, community desires may differ from what many in the machine translation community might expect~\citep{mapuche,maori,hawaiian}. And if the translation model is of imperfect quality, is that still helpful for the community -- or is it perhaps offensive? These are questions that can only be answered by members of the community. In our interviews, we generally found that native speakers of the languages we spoke to were very enthusiastic for even lower-quality translation offerings. This said, no one person can represent the entire community, and there is much to learn in how to handle situations where opinions and desires differ within a community.
    \item \textbf{Commenting on Dialects:} Many ``languages'' have a wide variety of dialects, sometimes hardly mutually intelligible. Native speakers helped us understand when our models were producing a particular dialect, or mixing and matching them.
    \item \textbf{The correct name to use for a language:} Whereas a language like French has a fairly unambiguous name, many languages have multiple names, some of which may be offensive or exclusive. Frequently there exists a colonial name (e.g. ``Oriya'' or ``Lushai''), which may be more widely known, but may also be disliked by the members of the community, as well as a native name (e.g. ``Odia'' and ``Mizo'') that is lesser known but preferred. Similarly, there may be names which may feel exclusive to some sub populations -- for instance using the name ``Manipuri'' for the language of the Meitei ethnic group. Although the most common way to refer to the Meitei language is indeed ``Manipuri'', this is exclusive to the many other ethnic groups in the state of Manipur with their own languages.
\end{enumerate}

\clearpage

\section{Conclusions and Open Problems}

\subsection{Main Findings} 
Starting with an initial seed dataset of monolingual sentences spanning over $1500$ languages, we demonstrate that it is possible to build relatively clean web-mined monolingual text datasets for over $1000$ languages. We highlight the importance of incorporating expressive semi-supervised LangID models, document-level consistency signals, and several word-based and custom filtering techniques to identify and filter web text in long-tail languages (Section~\ref{sec:data:dummy}). Using this approach we are able to build a multilingual unlabeled text dataset containing over $1$ million sentences for more than $200$ languages and over $100$ thousand sentences in more than $400$ languages (Section~\ref{sec:data:data}).

Training on this dataset and a parallel corpus spanning $112$ languages, we build massively multilingual models capable of translating across $1000$ languages (Section~\ref{sec:mt}). We highlight the importance of model capacity when training highly multilingual translation models and the positive effect on zero-resource quality when increasing the number of languages in the model. We also describe the significant quality improvements achievable by incorporating large-scale back-translation and self-training, and share our findings towards developing practical, inference-friendly models for long-tail languages.

We evaluate our models on evaluation sets collected for $38$ languages, and highlight the need for choosing the right automatic metric (\chrf{}) when evaluating long-tail languages (Section~\ref{sec:bleuvschrf}). Apart from automatic metrics on evaluation sets, we additionally release approximate reference-free quality scores of our $1000$-language MT model to provide an indicator of web-trained multilingual model quality on hundreds of previously under-studied languages (Section~\ref{sec:rtteval}). We perform human evaluations on a subset of the languages in distilled models, and highlight that it is possible to build high quality, practical MT models for long-tail languages utilizing the approach described in this work (Section~\ref{sec:human}).

Through qualitative and quantitative analysis of the model outputs, we reveal a few characteristic error modes of our models; including confusing distributionally similar and infrequent tokens, and also producing verbose and inaccurate translations for short or single word queries arising from the extreme data sparsity of the zero-resource setting (Sections~\ref{sec:miniaturecrocodile} and~\ref{sec:errorshort}). We furthermore highlight several other observations from our studies, including non-English-centric direct translation, zero-shot transliteration, the effect of terminal punctuation on translation quality, and the robustness of the model to the non-standard glyph usages that are common for many languages (Section~\ref{sec:additional}).

Finally, we highlight several indispensable contributions of native speakers who helped us evaluate, understand, filter and improve our datasets and models; and helped us understand the overall context of how these models should fit in with their communities (Section~\ref{sec:nativespeakers}).

\subsection{Related Work}
\label{sec:related_work}

There is a considerable wealth of literature on building highly multilingual text corpora, LangID models, and MT models. Our work differs largely in the scale, quality, and number of languages covered, together with the integration of many moving parts in the entire data-to-translation-model pipeline.

Access to multilingual datasets for NLP research has vastly improved over the past years. Since 2006, the \textit{Web as Corpus} workshops have focused on the challenges around identifying relevant pages, extracting clean text, content de-duplication, and many other relevant topics~\citep{wac-2020-web,jakubicek-etal-2020-current}. A variety of web-derived collections for hundreds of languages is available for anyone to download, such as the Corpora Collection at Leipzig University \citep{goldhahn-etal-2012-building}, the Corpus of Global Language Use \citep{Dunn2020}, ParaCrawl~\citep{espla-etal-2019-paracrawl, banon-etal-2020-paracrawl}, WikiMatrix~\citep{schwenk2019wikimatrix}, CCNET~\citep{wenzek-etal-2020-ccnet} and CCAligned~\citep{elkishky_ccaligned_2020}, \mbox{OSCAR}~\citep{ortiz-suarez-etal-2019-asynchronous, ortiz-suarez-etal-2020-monolingual,oscar2022}, and several others; all of which have between 100 and 300 languages.  The largest language coverage is probably An Crúbadán, which does not leverage LangID, and found (small amounts of) web data in about 2,000 languages \citep{Scannell-2007}. 
These corpora have in turn enabled a variety of highly multilingual models, like mT5 \citep{xue2020mt5}, M2M\nobreakdash-100 \citep{fan2020englishcentric}, and M4 \citep{arivazhagan2019massively,siddhant2022towards}. 

Curating such datasets relies on the websites giving clues about the language of their contents (e.g. a language identifier in the URL) and on automatic language classification (LangID).
It is commonly known that these automatically crawled and filtered datasets tend to have overall lower quality than hand-curated collections~\citep{koehn-etal-2020-findings}, but their quality is rarely measured directly, and is rather judged through the improvements they bring to downstream applications~\citep{schwenk2019wikimatrix}. Therefore, many of these multilingual web corpora suffer from serious quality issues, especially for low-resource languages. A recent audit conducted by \citet{kreutzer2022quality} on five public, multilingual datasets found pervasive issues. Many corpora claiming to be in one particular language in fact contained zero percent in-language content --- and sometimes zero percent linguistic content entirely. Of the many issues contributing to this phenomenon, a fundamental one is the poor efficacy of LangID on low-resource languages.

Several works have investigated LangID at the level of multilinguality studied in this work. One relevant LangID implementation is \citet{Dunn2020}, achieving an F1 above 0.95 for 464 languages, and offering a thorough evaluation on different data sources and domains. The only LangID systems with higher coverage that we are aware of are those developed by \citet{brown2012finding,brown2013selecting,brown-2014-non}, with the most recent version covering as many as 1,366 language varieties, with accuracy above 99\%. Finally, \citet{caswell2020language} trains LangID models on 1,629 languages, and demonstrates that although they appear to have very high scores on held-out evaluation sets, in practice, when applied to web text, they produce datasets of almost unusable noisiness. Various error pathologies are detailed, and a few novel filtering techniques are proposed to counteract them, including \tfiif{} filtering and semi supervised LangID (SSLID). The present work can be viewed as an extension of \citet{caswell2020language}, and a fusion of it with translation technology.

Our MT modeling approach builds on several previous works on massively multilingual, zero-resource and self-supervised MT; differing primarily in terms of the scale of multilinguality, model capacity and the extreme data sparsity in our experimental setting. We call these approaches that combine different aspects of scale as M4: massively multilingual, massive machine translation\footnote{\url{https://ai.googleblog.com/2019/10/exploring-massively-multilingual.html}}.

Multilingual Neural Machine Translation models were first introduced during the last decade~\citep{firat-etal-2016-multi,johnson-etal-2017-googles}, but the initial versions of these models were limited to a few languages (10--12). Over the last few years, there has been an explosion of work focusing on massively multilingual models that could translate between around $100$ languages~\citep{neubig-hu-2018-rapid,aharoni2019massively,arivazhagan2019massively,zhang2020improving,tang-etal-2021-multilingual,fan2021beyond}. However, most work on massively multilingual MT has focused on the purely supervised setting. There are a few works that have ventured beyond the limitations of large-scale multilingual corpora, and trained MT models spanning over $1000$ languages~\citep{mueller-etal-2020-analysis}, usually limited to narrow-domain (often religious) corpora.

Another stream of research on unsupervised MT developed modeling approaches to train MT models using monolingual datasets only~\citep{lample2017unsupervised,artetxe2017unsupervised,DBLP:journals/corr/abs-1905-02450}. With the advent of multilingual pre-training, with models like multilingual BERT~\citep{devlin2019bert}, XLM~\citep{lample2019cross}, mBART~\citep{liu2020multilingual} and others, the focus shifted towards fine-tuning pre-trained models with paired data in a sub-set of the pre-training languages to enable zero-resource translation in the remaining languages. These approaches are often complemented with large-scale back-translation~\citep{sennrich-etal-2016-improving, edunov-etal-2018-understanding} to continue improving the model beyond its initial zero-shot performance. 

Our work builds on~\citet{siddhant2020leveraging,garcia-etal-2021-harnessing,siddhant2022towards} that combines together multilingual supervised MT, zero-resource MT~\citep{firat-etal-2016-zero} and self-supervised learning within a single model. We extend the work in~\citet{siddhant2022towards} by scaling to larger models, a more multilingual dataset, utilizing self-training and a novel filtering technique based on round-trip translation consistency and LangID predictions.

Apart from efforts focused on building highly multilingual web-mined corpora and MT models, another line of NLP research has focused on building datasets and NLP technologies for specific languages, not necessarily from web content. Many of these are grassroots, bottom-up efforts from the affected communities, organized through research collectives like Masakhane \citep{masakhane}, Turkish Interlingua \citep{mirzakhalov2021large, mirzakhalov2021evaluating}, and GhanaNLP~\citep{ghananlp};
and conferences and workshops like AfricaNLP\footnote{\url{https://africanlp.masakhane.io/}}, AmericasNLP~\citep{mager-etal-2021-findings} and ArabNLP\footnote{\url{http://www.arabic-nlp.net/}}. These efforts, in addition to providing datasets, frequently provide models and baselines, or even public interfaces, like the Khaya Translator Web App\footnote{\url{https://ghananlp.org/project/translator-webapp/}} by GhanaNLP for West African languages, and the lesan.ai\footnote{\url{https://lesan.ai/translate}} translation website for Ethiopian languages.

Participation is especially strong from the African continent, including corpora and models for pan East-African languages \citep{babirye2022building}, languages from the Horn of Africa \citep{hornmt}, Ethiopian languages \citep{abate-etal-2018-parallel,gezmu2021extended}, Ugandan languages \citep{akera2022machine}, Emakhuwa \citep{felermino2021towards},  South-African languages \citep{eiselen-puttkammer-2014-developing}, Setswana and Sepedi \citep{marivate-etal-2020-investigating}, Yorùbá \citep{adelani-etal-2021-effect,adelani2021menyo},  Oshiwambo \citep{nekoto2022participatory}, Igbo \citep{ezeani2020igbo}, 
Zulu~\citep{rooweither_mabuya_2021_5035171},
Twi \citep{azunre2021english}, Gbe \citep{hacheme2021english2gbe}, Bambara \citep{tapo2021domain}, and Fon \citep{emezue-dossou-2020-ffr}. Outside of Africa, corpora have been created for languages of the Americas, including for four indigenous languages of Peru in \citet{bustamante-etal-2020-data}, the numerous results on the largely South- and Central American languages from the first AmericasNLP conference \citep{mager-etal-2021-findings}, and the Inuktitut language of Canada \citep{joanis-etal-2020-nunavut}. Datasets for lower-resourced languages of India have also sprung up, including the 13-language PMIndia \citep{haddow2020pmindia}, and datasets focused on languages of the Northeast like Mizo \citep{thihlum2020mizo}, Khasi \citep{laskar-etal-2021-enkhcorp1} and Assamese \citep{laskar-etal-2020-enascorp1}. Finally, a variety of such datasets and models are available for public use on HuggingFace\footnote{\url{https://huggingface.co/datasets?multilinguality=multilinguality:translation&task_categories=task_categories:translation&sort=downloads}} or Zenodo\footnote{\url{https://zenodo.org/communities/africanlp/}}. We believe language-specific efforts to be orthogonal and complementary to massively multilingual approaches for corpora building and modeling, as we elaborate further in Section~\ref{sec:future}.

\subsection{Future Work}
\label{sec:future}

Barring a dramatic increase in the amount of web text available for long-tail languages, the types of errors produced by our zero-resource models (Section \ref{sec:miniaturecrocodile}) are likely to persist. We highlight a few potential directions for future research that could help address the data sparsity that underlies the quality limitations of these models.

\textbf{Utilizing dictionaries to ground distributionally similar words:} One approach to address errors with distributionally similar words could involve helping ground the model's translations by utilizing bilingual dictionaries or similar resources. Dictionaries are relatively widely available and have already yielded promising results for individual language pairs~\citep{xia-etal-2019-generalized, duan-etal-2020-bilingual, karamanolakis-etal-2020-cross, reid-etal-2021-afromt}. For languages where dictionaries do not exist or coverage is low, dictionaries are much cheaper to build as compared to a dataset of parallel sentences. Efforts to develop high-coverage dictionaries and modeling approaches to incorporate them in massively multilingual MT models could nicely complement a corpus of monolingual web text and massively multilingual MT.

\textbf{Complementing massively multilingual modeling with language-specific efforts:} The quality of web-mined datasets is unlikely to match that of language-specific, hand-curated datasets; and building hand-curated datasets might be the only way forward to build text datasets for languages with limited presence on the web. However, we believe the two approaches to be orthogonal and complementary. Leveraging highly multilingual web-mined datasets and models significantly reduces the amount of data and research efforts needed to build practical NLP technologies for these languages~\citep{wang-etal-2020-extending,emezue-dossou-2021-mmtafrica,adelani2022afew,alabi2022maft,nekoto2022participatory}, and research efforts could be more efficient by building resources and modeling approaches that complement the weaknesses of web-based massively multilingual models. Furthermore, community-based contributions could yield other useful language-specific tools, like specialized data filters as in Section~\ref{sec:errormagnification}, tools to normalize orthography or script (like those described in Appendix \ref{sec:mtei}) or pre- and post-processors to correct certain mistakes, improve diacritization, etc.

\textbf{Leveraging multimodal datasets and models:} A large proportion of the $7000+$ languages spoken in the world have no written forms or standardized orthographic conventions. For a large majority there is limited amounts of text data available on the web. Being able to train models that can learn from and represent speech and text jointly~\citep{Zheng2021FusedAA,mslam,Chen2022MAESTROMS,Bai2022A3TAA,Tang2022UnifiedSP} is essential to alleviate data sparsity and building more robust language technologies for such languages.

In future work we plan to investigate the above-mentioned and related threads of research, hopefully making progress towards building and supporting language and speech technologies for more languages.

\subsubsection*{Acknowledgments}
We would like to extend our deepest gratitude to the following native speakers and members of the affected communities, who helped us in a wide variety of ways:

Yasser Salah Eddine Bouchareb (\textbf{Algerian Arabic}); Mfoniso Ukwak (\textbf{Anaang}); Bhaskar Borthakur, Janu Nelakanti, Kishor Barman, Rasika Saikia, Suraj Bharech (\textbf{Assamese}); Annette David, William Merza (\textbf{Assyrian Neo Aramaic}), Ruben Hilare Quispe (\textbf{Aymara}); Devina Suyanto, Puspa Si Pinguin (\textbf{Balinese}); Allahserix Auguste Tapo, Bakary Diarrassouba, Maimouna Siby, Moussa Doumbouya (\textbf{Bambara}); Mohammad Jahangir (\textbf{Baluchi}); Subhajit Naskar (\textbf{Bengali}); Animesh Pathak, Ankur Bapna, Anup Mohan, Chaitanya Joshi, Chandan Dubey, Kapil Kumar, Manish Katiyar, Mayank Srivastava, Neeharika, Saumya Pathak, Tanya Sinha, Vikas Singh (\textbf{Bhojpuri}); Bowen Liang, Ellie Chio, Eric Dong, Frank Tang, Jeff Pitman, John Wong, Kenneth Chang, Manish Goregaokar, Mingfei Lau, Ryan Li, Yiwen Luo (\textbf{Cantonese}); Monang Setyawan (\textbf{Caribbean Javanese}); Craig Cornelius (\textbf{Cherokee}); Anton Prokopyev (\textbf{Chuvash}); Nash Rafeeq (\textbf{Dhivehi}); Rajat Dogra, Sid Dogra (\textbf{Dogri});  Mohamed Kamagate (\textbf{Dyula}); Chris Assigbe, Dan Ameme, Emeafa Doe, Irene Nyavor, Thierry Gnanih, Yvonne Dumor (\textbf{Ewe}); Abdoulaye Barry, Adama Diallo, Fauzia van der Leeuw, Ibrahima Barry (\textbf{Fulfulde}); Isabel Papadimitriou (\textbf{Greek}); Alex Rudnick (\textbf{Guarani}); Mohammad Khdeir (\textbf{Gulf Arabic}); Paul Remollata (\textbf{Hiligaynon}); Ankur Bapna (\textbf{Hindi}); Mfoniso Ukwak (\textbf{Ibibio}); Nze Lawson (\textbf{Igbo}); D.J. Abuy, Miami Cabansay (\textbf{Ilocano}); Archana Koul, Shashwat Razdan, Sujeet Akula (\textbf{Kashmiri}); Jatin Kulkarni, Salil Rajadhyaksha, Sanjeet Hegde Desai, Sharayu Shenoy, Shashank Shanbhag, Shashi Shenoy (\textbf{Konkani}); Ryan Michael, Terrence Taylor (\textbf{Krio}); Bokan Jaff, Medya Ghazizadeh, Roshna Omer Abdulrahman, Saman Vaisipour, Sarchia Khursheed (\textbf{Kurdish (Sorani)}); Suphian Tweel (\textbf{Libyan Arabic}); Espoir Murhabazi, Doudou Kisabaka, Marie-Celeste Bagniakana (\textbf{Lingala}); Colleen Mallahan, John Quinn (\textbf{Luganda}); Cynthia Mboli (\textbf{Luyia}); Abhishek Kumar, Neeraj Mishra, Priyaranjan Jha, Saket Kumar, Snehal Bhilare (\textbf{Maithili}); Lisa Wang (\textbf{Mandarin Chinese}); Cibu Johny (\textbf{Malayalam}); Viresh Ratnakar (\textbf{Marathi}); Abhi Sanoujam, Gautam Thockchom, Pritam Pebam, Sam Chaomai, Shangkar Mayanglambam, Thangjam Hindustani Devi (\textbf{Meiteilon (Manipuri)}); Hala Ajil (\textbf{Mesopotamian Arabic}); Hamdanil Rasyid (\textbf{Minangkabau}); Elizabeth John, Remi Ralte, S Lallienkawl Gangte, Vaiphei Thatsing, Vanlalzami Vanlalzami (\textbf{Mizo}); Ahmed Kachkach, Hanaa ElAzizi (\textbf{Moroccan Arabic});  George Ouais (\textbf{MSA}); Ujjwal Rajbhandari (\textbf{Newari}); Ebuka Ufere, Gabriel Fynecontry, Onome Ofoman, Titi Akinsanmi (\textbf{Nigerian Pidgin}); Marwa Khost Jarkas (\textbf{North Levantine Arabic}); Abduselam Shaltu, Ace Patterson, Adel Kassem, Mo Ali, Yonas Hambissa (\textbf{Oromo}); Carlos Molina Vital, Elvia Andia Grageda, Helvia Taina, Marisol Necochea (\textbf{Quechua}); Rowena Marin (\textbf{Romani}), AbdelKarim Mardini (\textbf{Saidi Arabic}); Ishank Saxena, Manasa Harish, Manish Godara, Mayank Agrawal, Nitin Kashyap, Ranjani Padmanabhan, Ruchi Lohani, Shilpa Jindal, Shreevatsa Rajagopalan, Vaibhav Agarwal, Vinod Krishnan (\textbf{Sanskrit}); Nabil Shahid (\textbf{Saraiki}); Ayanda Mnyakeni (\textbf{Sesotho, Sepedi}); Landis Baker (\textbf{Seychellois Creole}); Taps Matangira (\textbf{Shona}); Ashraf Elsharif (\textbf{Sudanese Arabic}); Sakhile Dlamini (\textbf{Swati}); Hakim Sidahmed (\textbf{Tamazight}); Melvin Johnson (\textbf{Tamil}); Sneha Kudugunta (\textbf{Telugu}); Alexander Tekle, Bserat Ghebremicael, Nami Russom, Naud Ghebre (\textbf{Tigrinya}); Abigail Annkah, Diana Akron, Maame Ofori, Monica Opoku-Geren, Seth Duodu-baah, Yvonne Dumor (\textbf{Twi}); Ousmane Loum (\textbf{Wolof}), and Daniel Virtheim (\textbf{Yiddish}).

The authors also thank Raiomond Doctor and Cibu Johny for their invaluable help with the Meiteilon (Manipuri) transliteration project, and Colin Cherry, Markus Freitag and Johan Schalkwyk for their invaluable comments that helped improve the paper.

\FloatBarrier

\newpage
\bibliography{iclr2022_conference.bbl}
\bibliographystyle{iclr2022_conference}

\newpage

\appendix

\section{Complete Human Evaluation and \chrf{} Results for Distilled Models}

Table \ref{tab:distilledperf} gives the performance of the distilled models, as measured both by humans on a seven point scale (nonsense to perfect), and in \chrf{}. Please keep in mind that human evaluation numbers are quite hard to compare across languages.

\insertdistilledperf

\section{Complete Audit Results}
\label{appendix:audit}
We rated samples of 100 sentences for \howmanylaguagesaudited{} of the languages present in the dataset we crawled. The error metrics used are described in Table \ref{tab:audit_instructions}. The breakdown can be seen in Table \ref{tab:audit}.

\insertauditinstructions
\insertntlaudit

\subsection{Strict RTT LangID ChrF}
\label{appendix:rtteval}
In the main section of this paper (\ref{sec:rtteval}), we reported the \textit{loose} version of \rttlangidchrf{}, which seems to correlate better with \chrf{} and data size. One issue with this score is that it doesn't penalize the model for producing content in the wrong language, whereas the \textit{strict} version does.

\begin{figure}[t!]
\begin{center}
\includegraphics[scale=0.315]{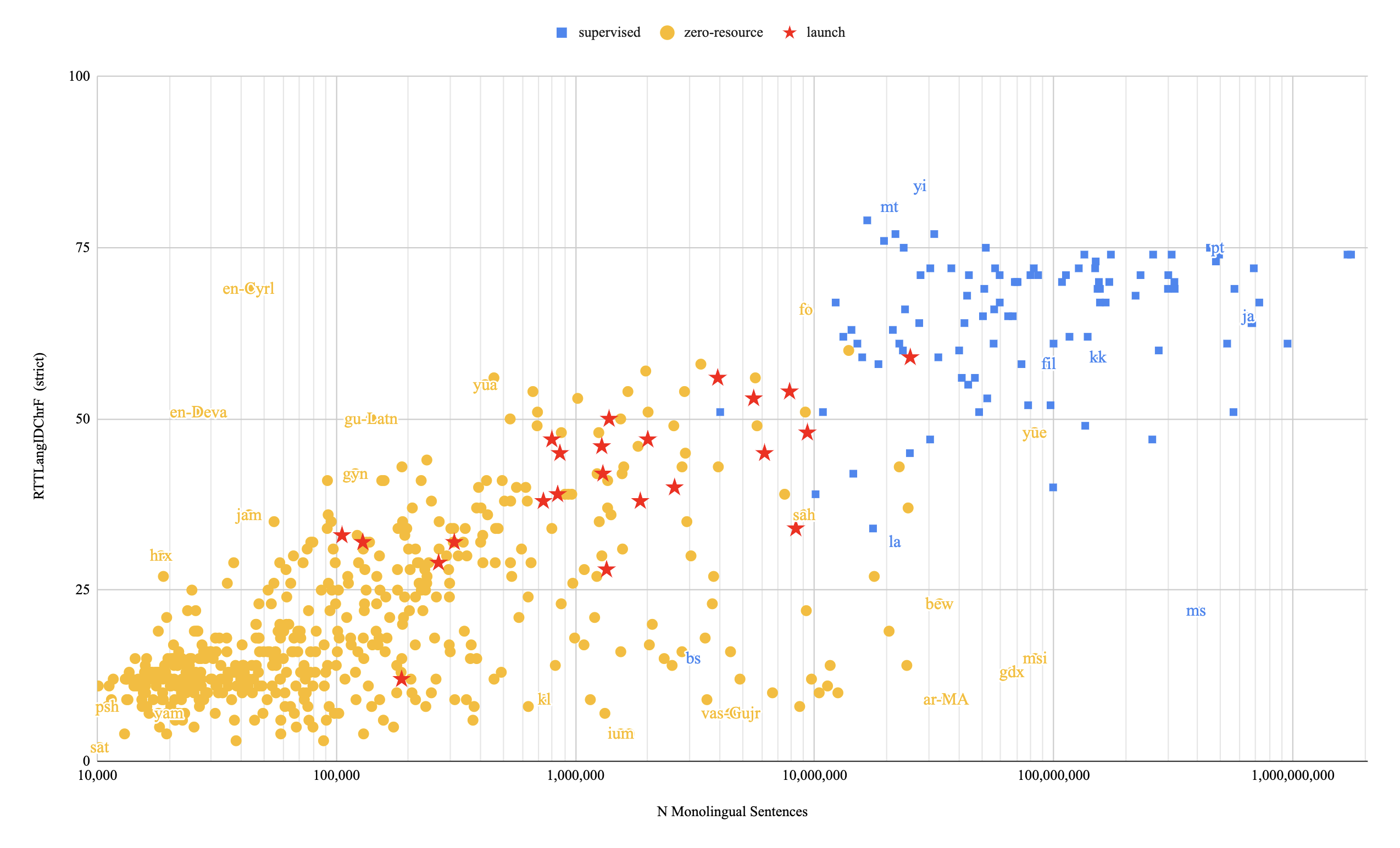}
\caption{Plot of \rttlangidchrf{} scores (strict) for languages as a function of log monolingual data size. This score has worse correlation with metrics like \chrf{} than the loose veriosn of \rttlangidchrf{}, but shows an interesting trend when plotted versus data size like this. Compared to the loose version, languages much below the trend line on the right-hand side are often close to high-resource languages (E.g. Betawi/\texttt{bew}, Sabah Malay/\texttt{msi}, Godwari/\texttt{gdx}, Darija/\texttt{ar-MA}), indicating that their apparently large monolingual datasets are actually a result of over-triggering with a higher-resource language like Indonesian, and there are many wrong-language intermediate translations.}
\label{fig:rttlangidchrfappendix}
\end{center}
\end{figure}

Figure \ref{fig:rttlangidchrfappendix} shows strict \rttlangidchrf{} as a function of log data size. Although the correlation is worse than with the loose version, there is a clear trend, and there appears to be some sort of upper bound in quality as a function of data size.

\section{Transliteration for Meiteilon}
\label{sec:mtei}
The large majority of the text we found online for Meiteilon (Manipuri) was in the Bengali script. Finding almost no Meiteilon in its native script, Meetei Mayek (\texttt{mni-Mtei}), we initially erroneously assumed that this script was archaic or only used in rare contexts. However, conversations with Meiteilon speakers quickly disillusioned us of this notion -- not only is it used, but it is now on its way to being the primary script used in the state. Most likely, the reason our \texttt{mni-Mtei} corpus was so small was because the available text online was largely in non-Unicode fonts, and therefore inaccessible to our LangID model (which had not been trained to detect such non-Unicode data). Therefore, we needed to convert our Bengali-script corpus to Meetei Mayek. However, this is not a straightforward task for such a non-injective script as Bengali.

Meiteilon is a tonal Tibeto-Burman language that is one of the
scheduled languages of India and a lingua franca of the Manipur state~\citep{chelliah:1997,singh:2011}. The Meetei Mayek script is
an indigenous script that was used to record Meiteilon until the 18th
century when it was mostly superseded by the Bengali script.
Despite recent Meetei Mayek gradual revitalization efforts by the Indian
government, Meetei Mayek literacy is still quite low~\citep{singh:2007}
and the available online Meiteilon data is mostly in
Bengali script~\citep{achom:2015,moirangthem:2021}.

Meetei Mayek belongs to the Tibetan family of Brahmic scripts and is
well suited for Meiteilon phonology representing a near bijective mapping
between graphemes and phonemes of a language~\citep{singh:2007}. Unlike the major
Brahmic scripts, this script uses a special class of explicit silent final
consonants (\emph{lonsum iyek}) in closed syllable codas, but these
consonants are represented as full letters rather than combining signs.
In modern Meetei Mayek orthography, the falling tone is often unmarked or
sometimes marked with full stop punctuation, whereas in the traditional
literature a special \emph{lum iyek} sign was used~\citep{everson:2007}.

Unlike Meetei Mayek, the orthographic conventions for Meiteilon in Bengali
script are ambiguous due to its larger letter inventory, where more than one
Bengali letter or clusters of letters may map to a single Meiteilon sound~\citep{singh:2007,khanganba:2014}.
This implies that any Bengali to Meeitei Mayek transliteration mechanism
needs to implement a many-to-one relation.

\begin{figure}
\centering%
\begin{adjustbox}{width=0.50\linewidth,center}
\begin{tikzpicture}
  \tikzset{
    mynode/.style={rectangle,rounded corners, draw=gray, top color=white,
      bottom color=white!90!gray,very thick, inner sep=1em,
      minimum size=1em, text centered, drop shadow,
      minimum width=2.4cm, text width=2.3cm},
    myline/.style={draw, -latex'},
  }
  \node[] (input) {\texttt{Beng}};
  \node[mynode, right=0.65cm of input] (visual) {$\mathcal{N}$ \\\texttt{Beng}};
  \node[mynode, right=1.3cm of visual] (romanize) {$\mathcal{R}'$ \\\texttt{Beng}};
  \node[mynode, below=1.3cm of romanize] (postprocess) {$\mathcal{P}$ \\\texttt{Latn}};
  \node[mynode, left=1.3cm of postprocess] (invromanize) {$\mathcal{R}^{-1}$ \\\texttt{Latn}};
  \node[left=0.65cm of invromanize] (output) {\texttt{Mtei}};
  
  \path[myline] (input) -- (visual) node[midway,above] {$\circ$};
  \path[myline] (visual) -- (romanize) node[midway,above] {$\circ$} node[midway,below] {\texttt{Beng}};
  \path[myline] (romanize) -- (postprocess) node[midway,right] {$\circ$} node[midway,left] {\texttt{Latn}};
  \path[myline] (postprocess) -- (invromanize) node[midway,above] {$\circ$} node[midway,below] {\texttt{Latn}};
  \path[myline] (invromanize) -- (output) node[midway,above] {$\circ$};

\end{tikzpicture}
\end{adjustbox}
\caption{Bengali (\texttt{Beng}) to Meetei Mayek (\texttt{Mtei}) transliteration components. The script codes are denoted according to ISO 15924~\citep{iso:15924}.}
\label{fig:beng-mni-overview}
\end{figure}
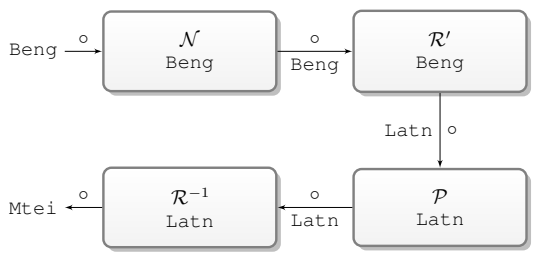

Our transliterator uses the open-source Nisaba library of finite-state script 
normalization and processing utilities~\citep{johny-etal-2021-finite}.\footnote{\url{https://github.com/google-research/nisaba}} The script operations in Nisaba are efficiently and succinctly represented
as weighted finite-state transducers (WFSTs) using Pynini finite-state grammars~\citep{gorman-2016-pynini,gorman:2021}. The main components of transliteration
workflow are shown in Figure~\ref{fig:beng-mni-overview}. The four component WFSTs
are compiled into the final transliteration WFST $\mathcal{T} = \mathcal{N} \circ \mathcal{R'} \circ \mathcal{P} \circ \mathcal{R}^{-1}$, where ``$\circ$'' denotes FST composition operation~\citep{mohri:2009}. The first component transducer $\mathcal{N}$ implements visual normalization of the Bengali script input that consists of visually invariant normalization transformations including NFC~\citep{johny-etal-2021-finite}.
This is followed by the Meiteilon-specific Bengali to Latin script many-to-one
mapping $\mathcal{R}'$ that produces Latin script output in ISO 15919 format~\citep{iso:15919} augmented with some placeholder markers required for the next processing stage.\footnote{This Nisaba operation is denoted $\mathcal{R}'$ to
distinguish it from the more general \emph{reversible} romanization $\mathcal{R}$
provided by Nisaba for Bengali and Assamese.}

The third stage implements post-processing transformations $\mathcal{P}$ that are
required to resolve ambiguities represented by the placeholder markers (introduced
by $\mathcal{R}'$) based on the orthographic context. One example of such
transformation is the resolution of the Bengali \emph{virama} sign, whose original
purpose in the Bengali script is to mark silent consonants pronounced without an
inherent vowel in consonant clusters. Its Meetei Mayek counterpart, the \emph{apun iyek}
mark, functions differently --- it only applies to a non-silent subset of consonants
(i.e., all consonants excluding the set of \emph{lonsum iyek} mentioned above).
Hence, given the \emph{virama} placeholder in the input, two finite-state
context-dependent rewrites are required for the resolution: convert the preceding
consonant to \emph{lonsum iyek} representation and remove the placeholder if
the preceding consonant sound is covered by the \emph{lonsum iyek} set, otherwise
simply convert the \emph{virama} placeholder to \emph{apun iyek} for all other cases.

The final transducer $\mathcal{R}^{-1}$ implements reverse romanization transliterating
unambiguous Latin script input in ISO 15919 format into corresponding representation
in Meetei Mayek.

\section{Tables of non-Unicode fonts}
\label{appendix:nonuni}

Table \ref{tab:nonuni} shows the mappings of ASCII characters to Unicode points for a few fonts we ran into. We discovered these mappings by copy-pasting text from an environment where it rendered (e.g. a PDF) to one where the font wasn't installed (e.g. a text editor), and finding the Unicode character that looked like the way it rendered.

\inserttablenonuniappendix

\FloatBarrier 
\section{List of languages with regions, approximate speaker counts and data sizes}
\label{appendix:onetabletorulethemall} 
Information about the languages and datasets we found on the web, including the name, number of speakers, continent, script (writing system), cluster (see Section \ref{sec:data:cluster}), langID F1 score for the SSLID model, and \rttlangidchrf{} (loose) score. The \rttlangidchrf{} score provides an approximate measure of quality of the model, but it should not be trusted as a reliable measure of translation quality (Section \ref{sec:rtteval}). 
Number of speakers refers to the estimated number of L1 speakers, following the estimates from~\citet{51306}.

\newcommand{\insertgianttable}{
\begin{footnotesize}
\begin{longtable}{p{0.09\linewidth}|p{0.07\linewidth}p{0.17\linewidth}p{0.07\linewidth}p{0.09\linewidth}lp{0.09\linewidth}p{0.03\linewidth}l}

BCP-47  & 	Mono  & 	Language Name  & 	Speak.  & Cont.  &  Script  & 	Clust.  & 	F1  & 	RTT  \\
\hline
\endhead
\hline
en  &	7388M  &	English  &	550M  &	Europe  &	Latn  &	en  &	98.2  &	NA \\
es  &	1751M  &	Spanish  &	490M  &	Europe  &	Latn  &	es  &	98.8  &	77.8 \\
de  &	1693M  &	German  &	83M  &	Europe  &	Latn  &	de  &	98.4  &	78.2 \\
id  &	950M  &	Indonesian  &	200M  &	Asia  &	Latn  &	id  &	97.3  &	76.9 \\
hu  &	724M  &	Hungarian  &	13M  &	Europe  &	Latn  &	hu  &	98.8  &	68.1 \\
pl  &	687M  &	Polish  &	38M  &	Europe  &	Latn  &	pl  &	98.0  &	72.8 \\
zh  &	676M  &	Chinese  &	1000M  &	Asia  &	Hans  &	zh  &	99.1  &	64.0 \\
ko  &	672M  &	Korean  &	52M  &	Asia  &	Kore  &	ko  &	99.9  &	64.0 \\
ja  &	652M  &	Japanese  &	130M  &	Asia  &	Jpan  &	ja  &	99.6  &	65.2 \\
ru  &	570M  &	Russian  &	150M  &	Europe  &	Cyrl  &	ru  &	97.3  &	74.0 \\
tr  &	565M  &	Turkish  &	82M  &	Europe  &	Latn  &	tr  &	99.0  &	70.9 \\
th  &	531M  &	Thai  &	20M  &	Asia  &	Thai  &	th  &	98.9  &	67.9 \\
it  &	491M  &	Italian  &	61M  &	Europe  &	Latn  &	it  &	98.9  &	77.3 \\
pt  &	485M  &	Portuguese  &	220M  &	Europe  &	Latn  &	pt  &	98.0  &	80.0 \\
vi  &	477M  &	Vietnamese  &	96M  &	Asia  &	Latn  &	vi  &	99.7  &	73.8 \\
fr  &	450M  &	French  &	75M  &	Europe  &	Latn  &	fr  &	97.6  &	77.7 \\
ms  &	394M  &	Malay  &	80M  &	Asia  &	Latn  &	ms  &	92.3  &	77.5 \\
nl  &	320M  &	Dutch  &	21M  &	Europe  &	Latn  &	nl  &	99.2  &	79.8 \\
uk  &	320M  &	Ukrainian  &	32M  &	Europe  &	Cyrl  &	uk  &	99.2  &	72.1 \\
sv  &	311M  &	Swedish  &	8M  &	Europe  &	Latn  &	sv  &	98.7  &	79.0 \\
ro  &	301M  &	Romanian  &	19M  &	Europe  &	Latn  &	ro  &	96.8  &	75.1 \\
cs  &	300M  &	Czech  &	10M  &	Europe  &	Latn  &	cs  &	98.9  &	73.0 \\
fa  &	275M  &	Persian  &	40M  &	Asia  &	Arab  &	fa  &	95.0  &	70.3 \\
lv  &	260M  &	Standard Latvian  &	2M  &	Europe  &	Latn  &	lv  &	98.5  &	75.5 \\
ar  &	258M  &	Arabic  &	180M  &	Africa  &	Arab  &	ar  &	97.9  &	71.8 \\
no  &	231M  &	Norwegian  &	5M  &	Europe  &	Latn  &	no  &	87.1  &	79.9 \\
ta  &	220M  &	Tamil  &	76M  &	Asia  &	Taml  &	ta  &	100  &	67.9 \\
el  &	173M  &	Modern Greek  &	13M  &	Europe  &	Grek  &	el  &	99.4  &	75.8 \\
fi  &	171M  &	Finnish  &	5M  &	Europe  &	Latn  &	fi  &	97.0  &	72.3 \\
hi  &	165M  &	Hindi  &	320M  &	Asia  &	Deva  &	hi  &	97.9  &	75.5 \\
mr  &	156M  &	Marathi  &	83M  &	Asia  &	Deva  &	mr  &	99.5  &	69.8 \\
sk  &	156M  &	Slovak  &	4M  &	Europe  &	Latn  &	sk  &	99.5  &	72.6 \\
hy  &	154M  &	Eastern Armenian  &	6M  &	Europe  &	Armn  &	hy  &	100  &	70.5 \\
kk  &	153M  &	Kazakh  &	11M  &	Europe  &	Cyrl  &	kk  &	99.6  &	60.6 \\
da  &	152M  &	Danish  &	5M  &	Europe  &	Latn  &	da  &	97.0  &	79.3 \\
mk  &	150M  &	Macedonian  &	1M  &	Europe  &	Cyrl  &	mk  &	98.8  &	77.6 \\
bg  &	149M  &	Bulgarian  &	7M  &	Europe  &	Cyrl  &	bg  &	98.3  &	74.4 \\
sr  &	141M  &	Serbian Standard  &	6M  &	Europe  &	Cyrl  &	sr  &	98.4  &	- \\
ml  &	139M  &	Malayalam  &	35M  &	Asia  &	Mlym  &	ml  &	100  &	62.5 \\
az  &	135M  &	Azerbaijani  &	18M  &	Asia  &	Latn  &	az  &	98.3  &	70.9 \\
is  &	134M  &	Icelandic  &	300K  &	Europe  &	Latn  &	is  &	97.5  &	78.6 \\
te  &	127M  &	Telugu  &	82M  &	Asia  &	Telu  &	te  &	100  &	72.8 \\
ne  &	116M  &	Nepali  &	16M  &	Asia  &	Deva  &	ne  &	99.3  &	65.6 \\
et  &	112M  &	Estonian  &	1M  &	Europe  &	Latn  &	et  &	99.4  &	73.5 \\
he  &	108M  &	Modern Hebrew  &	4M  &	Europe  &	Hebr  &	he  &	100  &	- \\
mn  &	100M  &	Mongolian  &	5M  &	Asia  &	Cyrl  &	mn  &	97.3  &	61.5 \\
ur  &	99M  &	Urdu  &	120M  &	Asia  &	Arab  &	ur  &	98.4  &	75.5 \\
hr  &	97M  &	Croatian Standard  &	4M  &	Europe  &	Latn  &	hr  &	79.4  &	75.3 \\
fil  &	95M  &	Filipino  &	22M  &	Asia  &	Latn  &	fil  &	97.3  &	80.6 \\
lt  &	86M  &	Lithuanian  &	2M  &	Europe  &	Latn  &	lt  &	99.9  &	72.8 \\
msi  &	84M  &	Brunei-Sabah Malay  &	3M  &	Asia  &	Latn  &	\scriptsize{ace-12l}  &	94.3  &	69.6 \\
be  &	83M  &	Belarusian  &	3M  &	Europe  &	Cyrl  &	be  &	99.2  &	73.9 \\
kn  &	80M  &	Kannada  &	43M  &	Asia  &	Knda  &	kn  &	100  &	71.5 \\
my  &	78M  &	Burmese  &	33M  &	Asia  &	Mymr  &	my  &	99.7  &	53.3 \\
bn  &	73M  &	Bengali  &	260M  &	Asia  &	Beng  &	bn  &	98.0  &	70.4 \\
af  &	71M  &	Afrikaans  &	17M  &	Africa  &	Latn  &	af  &	97.7  &	82.1 \\
eu  &	70M  &	Basque  &	500K  &	Europe  &	Latn  &	eu  &	99.9  &	72.9 \\
gu  &	69M  &	Gujarati  &	56M  &	Asia  &	Gujr  &	gu  &	98.0  &	75.2 \\
mi  &	67M  &	Maori  &	200K  &	Oceania  &	Latn  &	\scriptsize{mi-4l}  &	99.6  &	67.2 \\
gdx  &	67M  &	Godwari  &	3M  &	Asia  &	Deva  &	\scriptsize{bfz-15l}  &	70.5  &	74.2 \\
gl  &	64M  &	Galician  &	2M  &	Europe  &	Latn  &	gl  &	94.2  &	76.9 \\
sl  &	59M  &	Slovenian  &	2M  &	Europe  &	Latn  &	sl  &	97.9  &	72.7 \\
lo  &	59M  &	Lao  &	20M  &	Asia  &	Laoo  &	\scriptsize{kjg-2l}  &	100  &	72.0 \\
gd  &	57M  &	Scottish Gaelic  &	50K  &	Europe  &	Latn  &	\scriptsize{alq-4l}  &	99.4  &	74.5 \\
so  &	56M  &	Somali  &	24M  &	Africa  &	Latn  &	\scriptsize{aa-13l}  &	99.7  &	67.4 \\
si  &	56M  &	Sinhala  &	16M  &	Asia  &	Sinh  &	si  &	100  &	61.3 \\
uz  &	53M  &	Uzbek  &	57M  &	Europe  &	Latn  &	uz  &	99.8  &	62.2 \\
sw  &	52M  &	Swahili  &	24M  &	Africa  &	Latn  &	sw  &	99.3  &	78.1 \\
km  &	51M  &	Central Khmer  &	16M  &	Asia  &	Khmr  &	km  &	99.9  &	69.4 \\
ha  &	51M  &	Hausa  &	70M  &	Africa  &	Latn  &	\scriptsize{amo-7l}  &	99.4  &	70.5 \\
wry  &	50M  &	Merwari  &	8M  &	Asia  &	Deva  &	\scriptsize{bfz-15l}  &	68.2  &	- \\
ky  &	49M  &	Kirghiz  &	8M  &	Europe  &	Cyrl  &	ky  &	99.8  &	56.7 \\
am  &	47M  &	Amharic  &	22M  &	Africa  &	Ethi  &	am  &	95.1  &	58.0 \\
ca  &	44M  &	Catalan  &	4M  &	Europe  &	Latn  &	ca  &	99.5  &	75.0 \\
ceb  &	44M  &	Cebuano  &	16M  &	Asia  &	Latn  &	\scriptsize{abx-11l}  &	99.1  &	73.9 \\
yo  &	43M  &	Yoruba  &	21M  &	Africa  &	Latn  &	\scriptsize{bcq-Latn-12l}  &	99.6  &	73.7 \\
sd  &	42M  &	Sindhi  &	68M  &	Asia  &	Arab  &	\scriptsize{bgq-Arab-4l}  &	99.0  &	67.3 \\
gju-Deva-IN  &	41M  &	Gujari  &	1M  &	Asia  &	Deva  &	\scriptsize{bfz-15l}  &	68.6  &	- \\
co  &	41M  &	Corsican  &	50K  &	Europe  &	Latn  &	\scriptsize{cjk-14l}  &	97.5  &	75.2 \\
mg  &	40M  &	Malagasy  &	18M  &	Africa  &	Latn  &	\scriptsize{apy-7l}  &	99.9  &	65.4 \\
pa  &	37M  &	Eastern Punjabi  &	28M  &	Asia  &	Guru  &	pa  &	99.7  &	71.9 \\
ar-MA  &	36M  &	Arabic  &	180M  &	Africa  &	Arab  &	\scriptsize{acm-9l}  &	46.1  &	69.0 \\
bew  &	33M  &	Betawi  &	5M  &	Asia  &	Latn  &	\scriptsize{abs-13l}  &	93.8  &	67.2 \\
ny  &	33M  &	Chichewa  &	12M  &	Africa  &	Latn  &	\scriptsize{cjk-14l}  &	98.8  &	66.1 \\
sq  &	32M  &	Albanian  &	5M  &	Europe  &	Latn  &	sq  &	99.3  &	79.0 \\
ka  &	30M  &	Georgian  &	4M  &	Europe  &	Geor  &	ka  &	99.9  &	71.9 \\
haw  &	30M  &	Hawaiian  &	50K  &	Americas  &	Latn  &	\scriptsize{gah-11l}  &	99.7  &	58.3 \\
hmn  &	28M  &	Hmong  &	3M  &	Europe  &	Latn  &	\scriptsize{hmn-3l}  &	99.7  &	71.5 \\
yi  &	28M  &	Yiddish  &	2M  &	Europe  &	Hebr  &	\scriptsize{gv-3l}  &	100  &	84.8 \\
ig  &	27M  &	Igbo  &	18M  &	Africa  &	Latn  &	\scriptsize{dzg-5l}  &	99.0  &	68.6 \\
ckb  &	25M  &	Central Kurdish  &	7M  &	Europe  &	Arab  &	\scriptsize{ckb-2l}  &	99.3  &	60.0 \\
tt  &	25M  &	Tatar  &	4M  &	Europe  &	Cyrl  &	tt  &	97.0  &	52.5 \\
gsw  &	25M  &	Central Alemannic  &	5M  &	Europe  &	Latn  &	\scriptsize{ang-7l}  &	93.2  &	73.3 \\
arq  &	24M  &	Algerian Arabic  &	27M  &	Africa  &	Arab  &	\scriptsize{acm-9l}  &	79.0  &	66.0 \\
sm  &	24M  &	Samoan  &	500K  &	Oceania  &	Latn  &	\scriptsize{mi-4l}  &	99.9  &	67.6 \\
eo  &	24M  &	Esperanto  &	2M  &	Europe  &	Latn  &	eo  &	99.7  &	78.4 \\
zu  &	23M  &	Zulu  &	11M  &	Africa  &	Latn  &	\scriptsize{gup-8l}  &	96.3  &	71.3 \\
st  &	23M  &	Southern Sotho  &	14M  &	Africa  &	Latn  &	\scriptsize{cce-8l}  &	98.8  &	69.2 \\
ru-Latn  &	23M  &	Russian  &	150M  &	Europe  &	Latn  &	\scriptsize{bg-Latn-5l}  &	97.2  &	51.4 \\
cy  &	22M  &	Welsh  &	700K  &	Europe  &	Latn  &	cy  &	99.7  &	79.4 \\
la  &	22M  &	Latin  &	0K  &	Europe  &	Latn  &	la  &	93.2  &	67.9 \\
or  &	21M  &	Odia  &	35M  &	Asia  &	Orya  &	or  &	100  &	63.2 \\
mt  &	21M  &	Maltese  &	500K  &	Europe  &	Latn  &	mt  &	99.5  &	84.2 \\
rkt  &	20M  &	Rangpuri  &	15M  &	Asia  &	Beng  &	\scriptsize{bpy-7l}  &	98.8  &	59.2 \\
ht  &	20M  &	Haitian  &	10M  &	Americas  &	Latn  &	ht  &	99.9  &	79.3 \\
sn  &	18M  &	Shona  &	9M  &	Africa  &	Latn  &	\scriptsize{cjk-14l}  &	99.6  &	68.6 \\
bo  &	18M  &	Tibetan  &	1M  &	Asia  &	Tibt  &	\scriptsize{bo-2l}  &	99.9  &	28.5 \\
ps  &	18M  &	Nuclear Pashto  &	13M  &	Asia  &	Arab  &	ps  &	98.9  &	59.7 \\
ga  &	17M  &	Irish  &	200K  &	Europe  &	Latn  &	ga  &	98.8  &	80.9 \\
shu  &	16M  &	Chadian Arabic  &	1M  &	Africa  &	Arab  &	\scriptsize{acm-9l}  &	89.5  &	- \\
xh  &	16M  &	Xhosa  &	19M  &	Africa  &	Latn  &	\scriptsize{gup-8l}  &	97.8  &	69.0 \\
lb  &	15M  &	Moselle Franconian  &	400K  &	Europe  &	Latn  &	lb  &	96.7  &	63.7 \\
ku  &	15M  &	Kurdish  &	32M  &	Europe  &	Latn  &	ku  &	98.2  &	57.7 \\
tg  &	14M  &	Tajik  &	8M  &	Europe  &	Cyrl  &	tg  &	99.9  &	64.9 \\
hi-Latn  &	14M  &	Hindi  &	320M  &	Asia  &	Latn  &	\scriptsize{bgp-13l}  &	98.1  &	71.7 \\
jv  &	13M  &	Javanese  &	85M  &	Asia  &	Latn  &	jv  &	99.4  &	77.0 \\
apd-SD  &	13M  &	Sudanese Arabic  &	17M  &	Africa  &	Arab  &	\scriptsize{acm-9l}  &	74.2  &	69.1 \\
su  &	12M  &	Sundanese  &	39M  &	Asia  &	Latn  &	su  &	99.1  &	71.1 \\
mzn  &	12M  &	Mazanderani  &	6M  &	Asia  &	Arab  &	\scriptsize{apr-8l}  &	52.4  &	64.8 \\
meo  &	11M  &	Kedah Malay  &	3M  &	Asia  &	Latn  &	\scriptsize{ace-12l}  &	94.4  &	66.7 \\
fy  &	11M  &	Western Frisian  &	800K  &	Europe  &	Latn  &	fy  &	99.7  &	71.7 \\
mrw  &	10M  &	Maranao  &	800K  &	Asia  &	Latn  &	\scriptsize{agn-14l}  &	98.0  &	69.2 \\
rw  &	10M  &	Kinyarwanda  &	11M  &	Africa  &	Latn  &	rw  &	98.6  &	47.7 \\
apc  &	10M  &	North Levantine Arabic  &	15M  &	Europe  &	Arab  &	\scriptsize{abv-10l}  &	40.4  &	67.7 \\
as  &	9M  &	Assamese  &	13M  &	Asia  &	Beng  &	as  &	99.7  &	63.7 \\
aec  &	9M  &	Saidi Arabic  &	19M  &	Africa  &	Arab  &	\scriptsize{abv-10l}  &	78.0  &	64.3 \\
fo  &	9M  &	Faroese  &	50K  &	Europe  &	Latn  &	\scriptsize{bn-Latn-6l}  &	98.0  &	71.6 \\
pap  &	9M  &	Papiamento  &	600K  &	Americas  &	Latn  &	\scriptsize{ach-10l}  &	94.8  &	61.4 \\
sah  &	9M  &	Sakha  &	500K  &	Europe  &	Cyrl  &	\scriptsize{az-RU-13l}  &	97.7  &	43.4 \\
acm  &	9M  &	Gilit Mesopotamian Arabic  &	26M  &	Europe  &	Arab  &	\scriptsize{acm-9l}  &	35.2  &	64.2 \\
lus  &	8M  &	Mizo  &	700K  &	Asia  &	Latn  &	\scriptsize{bgr-6l}  &	100  &	35.9 \\
grc  &	8M  &	Ancient Greek (to 1453)  &	0K  &	Europe  &	Grek  &	grc  &	99.9  &	- \\
dv  &	8M  &	Dhivehi  &	300K  &	Asia  &	Thaa  &	\scriptsize{bpy-7l}  &	100  &	53.6 \\
oc  &	7M  &	Occitan  &	500K  &	Europe  &	Latn  &	\scriptsize{acf-10l}  &	96.7  &	73.4 \\
luz  &	7M  &	Southern Luri  &	900K  &	Asia  &	Arab  &	\scriptsize{fay-14l}  &	68.4  &	64.2 \\
sa  &	6M  &	Sanskrit  &	100K  &	Asia  &	Deva  &	\scriptsize{bho-13l}  &	97.8  &	57.6 \\
bgp-Arab  &	6M  &	Eastern Balochi  &	2M  &	Asia  &	Arab  &	\scriptsize{bal-13l}  &	55.1  &	- \\
ug  &	6M  &	Uyghur  &	10M  &	Asia  &	Arab  &	\scriptsize{ckb-2l}  &	100  &	50.5 \\
ba  &	6M  &	Bashkir  &	1M  &	Europe  &	Cyrl  &	\scriptsize{az-RU-13l}  &	98.0  &	58.2 \\
om  &	6M  &	Oromo  &	24M  &	Africa  &	Latn  &	\scriptsize{amf-6l}  &	99.7  &	56.0 \\
zza  &	5M  &	Zaza  &	2M  &	Europe  &	Latn  &	\scriptsize{aeb-Latn-11l}  &	71.7  &	- \\
zh-Latn  &	5M  &	Chinese  &	1000M  &	Asia  &	Latn  &	\scriptsize{cqd-8l}  &	99.2  &	22.2 \\
vas-Gujr  &	4M  &	Vasavi  &	1M  &	Asia  &	Gujr  &	\scriptsize{bhb-Gujr-3l}  &	78.7  &	68.5 \\
vec  &	4M  &	Venetian  &	4M  &	Europe  &	Latn  &	\scriptsize{bi-6l}  &	94.5  &	73.5 \\
tk  &	4M  &	Turkmen  &	8M  &	Europe  &	Latn  &	tk  &	99.9  &	57.5 \\
hil  &	4M  &	Hiligaynon  &	6M  &	Asia  &	Latn  &	\scriptsize{abx-11l}  &	99.3  &	70.8 \\
ti  &	4M  &	Tigrinya  &	9M  &	Africa  &	Ethi  &	\scriptsize{gez-8l}  &	99.9  &	57.0 \\
syl  &	4M  &	Sylheti  &	23M  &	Asia  &	Beng  &	\scriptsize{bpy-7l}  &	86.9  &	65.8 \\
phr  &	4M  &	Pahari Potwari  &	2M  &	Asia  &	Arab  &	\scriptsize{bal-13l}  &	54.5  &	73.6 \\
ayl  &	4M  &	Libyan Arabic  &	5M  &	Africa  &	Arab  &	\scriptsize{acm-9l}  &	86.3  &	65.0 \\
el-Latn  &	3M  &	Modern Greek  &	13M  &	Europe  &	Latn  &	\scriptsize{bg-Latn-5l}  &	98.9  &	26.0 \\
kaa  &	3M  &	Kara-Kalpak  &	600K  &	Europe  &	Cyrl  &	\scriptsize{az-RU-13l}  &	99.3  &	61.8 \\
bs  &	3M  &	Bosnian Standard  &	9M  &	Europe  &	Cyrl  &	bs  &	70.8  &	74.1 \\
bgp  &	3M  &	Eastern Balochi  &	2M  &	Asia  &	Latn  &	\scriptsize{bgp-13l}  &	91.1  &	70.0 \\
br  &	3M  &	Breton  &	200K  &	Europe  &	Latn  &	\scriptsize{br-4l}  &	82.8  &	42.8 \\
gdq  &	3M  &	Mehri  &	100K  &	Europe  &	Arab  &	\scriptsize{acm-9l}  &	51.7  &	- \\
tet  &	3M  &	Tetum  &	500K  &	Asia  &	Latn  &	\scriptsize{ceg-5l}  &	98.4  &	53.5 \\
to  &	3M  &	Tonga (Tonga Islands)  &	200K  &	Oceania  &	Latn  &	\scriptsize{mi-4l}  &	100  &	57.5 \\
cv  &	3M  &	Chuvash  &	1M  &	Europe  &	Cyrl  &	\scriptsize{az-RU-13l}  &	92.9  &	48.6 \\
kfr  &	3M  &	Kachchi  &	900K  &	Asia  &	Gujr  &	\scriptsize{bhb-Gujr-3l}  &	82.5  &	63.4 \\
ilo  &	3M  &	Ilocano  &	9M  &	Asia  &	Latn  &	\scriptsize{agn-14l}  &	93.4  &	47.6 \\
bg-Latn  &	3M  &	Bulgarian  &	7M  &	Europe  &	Latn  &	\scriptsize{bg-Latn-5l}  &	97.2  &	65.0 \\
bal  &	3M  &	Balochi  &	8M  &	Asia  &	Arab  &	\scriptsize{bal-13l}  &	23.0  &	61.8 \\
hif  &	2M  &	Fiji Hindi  &	500K  &	Oceania  &	Latn  &	\scriptsize{bgp-13l}  &	99.6  &	69.4 \\
mey  &	2M  &	Hassaniyya  &	3M  &	Africa  &	Arab  &	\scriptsize{acm-9l}  &	69.5  &	- \\
dz  &	2M  &	Dzongkha  &	200K  &	Asia  &	Tibt  &	\scriptsize{bo-2l}  &	98.2  &	39.5 \\
tn  &	2M  &	Tswana  &	8M  &	Africa  &	Latn  &	\scriptsize{cce-8l}  &	90.5  &	66.5 \\
bn-Latn  &	2M  &	Bengali  &	260M  &	Asia  &	Latn  &	\scriptsize{bn-Latn-6l}  &	98.5  &	57.5 \\
lg  &	2M  &	Ganda  &	4M  &	Africa  &	Latn  &	\scriptsize{cgg-15l}  &	98.7  &	58.2 \\
te-Latn  &	2M  &	Telugu  &	82M  &	Asia  &	Latn  &	\scriptsize{bgp-13l}  &	98.6  &	59.3 \\
mde  &	2M  &	Maba (Chad)  &	300K  &	Africa  &	Arab  &	\scriptsize{acm-9l}  &	48.5  &	- \\
nso  &	2M  &	Pedi  &	14M  &	Africa  &	Latn  &	\scriptsize{cce-8l}  &	98.6  &	68.0 \\
rm  &	2M  &	Romansh  &	50K  &	Europe  &	Latn  &	rm  &	99.9  &	61.7 \\
za  &	2M  &	Northern Daic  &	15M  &	Asia  &	Latn  &	\scriptsize{cqd-8l}  &	98.7  &	27.3 \\
tpi  &	2M  &	Tok Pisin  &	4M  &	Oceania  &	Latn  &	\scriptsize{bi-6l}  &	99.3  &	61.3 \\
pcm  &	2M  &	Nigerian Pidgin  &	30M  &	Africa  &	Latn  &	pcm  &	97.9  &	71.1 \\
mtq  &	2M  &	Muong  &	1M  &	Asia  &	Latn  &	\scriptsize{blt-Latn-4l}  &	96.3  &	- \\
cnh  &	2M  &	Haka Chin  &	100K  &	Asia  &	Latn  &	\scriptsize{bgr-6l}  &	99.9  &	37.2 \\
iba  &	2M  &	Iban  &	2M  &	Asia  &	Latn  &	\scriptsize{amo-7l}  &	99.8  &	54.8 \\
ja-Latn  &	2M  &	Japanese  &	130M  &	Asia  &	Latn  &	\scriptsize{dgz-5l}  &	98.6  &	20.9 \\
ium  &	2M  &	Iu Mien  &	800K  &	Asia  &	Latn  &	\scriptsize{cqd-8l}  &	98.9  &	18.9 \\
crs  &	2M  &	Seselwa Creole French  &	50K  &	Africa  &	Latn  &	\scriptsize{acf-10l}  &	99.8  &	72.9 \\
se  &	1.5M  &	North Saami  &	50K  &	Europe  &	Latn  &	\scriptsize{bn-Latn-6l}  &	90.2  &	- \\
kha  &	1.4M  &	Khasi  &	800K  &	Asia  &	Latn  &	\scriptsize{ace-12l}  &	99.9  &	38.3 \\
ln  &	1.4M  &	Kinshasa Lingala  &	15M  &	Africa  &	Latn  &	\scriptsize{jbu-8l}  &	96.4  &	56.1 \\
tyv  &	1.4M  &	Tuvinian  &	300K  &	Europe  &	Cyrl  &	\scriptsize{sty-2l}  &	99.4  &	47.4 \\
ce  &	1.4M  &	Chechen  &	1M  &	Europe  &	Cyrl  &	\scriptsize{ce-3l}  &	99.9  &	41.6 \\
mai  &	1.3M  &	Maithili  &	17M  &	Asia  &	Deva  &	\scriptsize{awa-14l}  &	98.6  &	60.8 \\
nr  &	1.3M  &	South Transvaal Ndebele  &	1M  &	Africa  &	Latn  &	\scriptsize{gup-8l}  &	86.3  &	- \\
dty  &	1.3M  &	Dotyali  &	800K  &	Asia  &	Deva  &	\scriptsize{bho-13l}  &	77.6  &	57.8 \\
ts  &	1.3M  &	Tsonga  &	13M  &	Africa  &	Latn  &	\scriptsize{cjk-14l}  &	72.9  &	55.0 \\
fj  &	1.3M  &	Fijian  &	300K  &	Oceania  &	Latn  &	\scriptsize{fj-4l}  &	98.8  &	37.5 \\
ak  &	1.3M  &	Akan  &	13M  &	Africa  &	Latn  &	\scriptsize{abr-5l}  &	93.8  &	49.5 \\
kaa-Latn  &	1.3M  &	Kara-Kalpak  &	600K  &	Europe  &	Latn  &	\scriptsize{bcq-Latn-12l}  &	86.7  &	55.9 \\
os  &	1.3M  &	Iron Ossetian  &	600K  &	Europe  &	Cyrl  &	\scriptsize{gaw-8l}  &	93.5  &	55.2 \\
bik  &	1.2M  &	Bikol  &	2M  &	Asia  &	Latn  &	\scriptsize{abx-11l}  &	94.4  &	68.7 \\
ber-Latn  &	1.2M  &	Berber  &	2M  &	Africa  &	Latn  &	\scriptsize{alq-4l}  &	99.7  &	32.9 \\
ctd-Latn  &	1.2M  &	Tedim Chin  &	300K  &	Asia  &	Latn  &	\scriptsize{bgs-9l}  &	99.5  &	23.5 \\
tly-IR  &	1.2M  &	North-Central Talysh  &	900K  &	Europe  &	Arab  &	\scriptsize{fay-14l}  &	88.7  &	63.1 \\
plk  &	1.1M  &	Kohistani Shina  &	200K  &	Asia  &	Arab  &	\scriptsize{bal-13l}  &	65.1  &	- \\
tks  &	1.1M  &	Takestani  &	200K  &	Asia  &	Arab  &	\scriptsize{fay-14l}  &	88.9  &	- \\
kmz  &	1.1M  &	Khorasan Turkic  &	400K  &	Asia  &	Arab  &	\scriptsize{fay-14l}  &	60.7  &	64.4 \\
mnw  &	1.1M  &	Mon  &	900K  &	Asia  &	Mymr  &	\scriptsize{ksw-3l}  &	99.9  &	24.4 \\
scl  &	1M  &	Shina  &	400K  &	Asia  &	Arab  &	\scriptsize{bal-13l}  &	30.1  &	- \\
ml-Latn  &	1M  &	Malayalam  &	35M  &	Asia  &	Latn  &	\scriptsize{bgp-13l}  &	98.7  &	58.2 \\
el-CY  &	1M  &	Modern Greek  &	13M  &	Europe  &	Grek  &	el-CY  &	99.6  &	71.0 \\
skr  &	1M  &	Saraiki  &	26M  &	Asia  &	Arab  &	\scriptsize{doi-Arab-5l}  &	99.8  &	65.1 \\
udm  &	1M  &	Udmurt  &	300K  &	Europe  &	Cyrl  &	\scriptsize{ady-9l}  &	98.1  &	45.6 \\
bbc  &	900K  &	Batak Toba  &	2M  &	Asia  &	Latn  &	\scriptsize{ace-12l}  &	98.3  &	47.9 \\
war  &	900K  &	Waray (Philippines)  &	3M  &	Asia  &	Latn  &	\scriptsize{bgs-9l}  &	98.3  &	70.8 \\
mki  &	900K  &	Dhatki  &	200K  &	Asia  &	Arab  &	\scriptsize{bal-13l}  &	26.8  &	- \\
srn  &	900K  &	Sranan Tongo  &	100K  &	Americas  &	Latn  &	\scriptsize{djk-3l}  &	98.4  &	56.9 \\
pck  &	900K  &	Paite Chin  &	50K  &	Asia  &	Latn  &	\scriptsize{bgs-9l}  &	99.5  &	32.3 \\
gn  &	900K  &	Guarani  &	5M  &	Americas  &	Latn  &	\scriptsize{gn-6l}  &	98.8  &	50.3 \\
qu  &	800K  &	Quechua  &	9M  &	Americas  &	Latn  &	\scriptsize{ae-Latn-10l}  &	98.3  &	48.8 \\
kj  &	800K  &	Kuanyama  &	600K  &	Africa  &	Latn  &	\scriptsize{kj-5l}  &	98.5  &	39.9 \\
bug  &	800K  &	Buginese  &	5M  &	Asia  &	Latn  &	\scriptsize{amo-7l}  &	83.5  &	- \\
ee  &	800K  &	Ewe  &	4M  &	Africa  &	Latn  &	\scriptsize{avn-3l}  &	99.5  &	49.6 \\
ltg  &	800K  &	East Latvian  &	200K  &	Europe  &	Latn  &	\scriptsize{gaw-8l}  &	94.9  &	63.7 \\
hsb  &	800K  &	Upper Sorbian  &	5K  &	Europe  &	Latn  &	\scriptsize{ach-10l}  &	59.6  &	- \\
kl  &	700K  &	Kalaallisut  &	50K  &	Americas  &	Latn  &	\scriptsize{bi-6l}  &	87.7  &	34.1 \\
bho  &	700K  &	Bhojpuri  &	51M  &	Asia  &	Deva  &	\scriptsize{bho-13l}  &	96.6  &	56.2 \\
ta-Latn  &	700K  &	Tamil  &	76M  &	Asia  &	Latn  &	\scriptsize{bgp-13l}  &	99.1  &	53.8 \\
ve  &	700K  &	Venda  &	2M  &	Africa  &	Latn  &	\scriptsize{cce-8l}  &	99.9  &	52.9 \\
bi  &	700K  &	Bislama  &	200K  &	Oceania  &	Latn  &	\scriptsize{bi-6l}  &	99.1  &	- \\
az-RU  &	700K  &	Azerbaijani  &	18M  &	Europe  &	Cyrl  &	\scriptsize{az-RU-13l}  &	99.7  &	58.1 \\
kbd  &	700K  &	Kabardian  &	2M  &	Europe  &	Cyrl  &	\scriptsize{ady-9l}  &	87.7  &	46.1 \\
nut  &	600K  &	Nung (Viet Nam)  &	1M  &	Asia  &	Latn  &	\scriptsize{blt-Latn-4l}  &	97.9  &	- \\
hif-Deva  &	600K  &	Fiji Hindi  &	500K  &	Oceania  &	Deva  &	\scriptsize{awa-14l}  &	89.5  &	69.9 \\
ar-Latn  &	600K  &	Arabic  &	180M  &	Europe  &	Latn  &	\scriptsize{aeb-Latn-11l}  &	95.2  &	29.5 \\
krc  &	600K  &	Karachay-Balkar  &	300K  &	Europe  &	Cyrl  &	\scriptsize{az-RU-13l}  &	98.5  &	52.7 \\
chm  &	600K  &	Mari  &	500K  &	Europe  &	Cyrl  &	\scriptsize{az-RU-13l}  &	99.5  &	44.5 \\
pag  &	600K  &	Pangasinan  &	1M  &	Asia  &	Latn  &	\scriptsize{agn-14l}  &	92.0  &	48.9 \\
mui  &	600K  &	Musi  &	3M  &	Asia  &	Latn  &	\scriptsize{abs-13l}  &	90.4  &	62.5 \\
shn  &	600K  &	Shan  &	3M  &	Asia  &	Mymr  &	\scriptsize{ksw-3l}  &	100  &	40.3 \\
gom-Latn  &	500K  &	Goan Konkani  &	2M  &	Asia  &	Latn  &	\scriptsize{bgp-13l}  &	99.0  &	56.0 \\
noe  &	500K  &	Nimadi  &	2M  &	Asia  &	Deva  &	\scriptsize{ahr-13l}  &	99.2  &	62.8 \\
mh  &	500K  &	Marshallese  &	50K  &	Oceania  &	Latn  &	\scriptsize{kpj-5l}  &	100  &	31.4 \\
min  &	500K  &	Minangkabau  &	6M  &	Asia  &	Latn  &	\scriptsize{ace-12l}  &	100  &	59.1 \\
yue  &	500K  &	Yue Chinese  &	84M  &	Asia  &	Hant  &	yue  &	99.1  &	55.4 \\
new  &	500K  &	Kathmandu Valley Newari  &	900K  &	Asia  &	Deva  &	\scriptsize{ahr-13l}  &	98.3  &	51.4 \\
cjk  &	500K  &	Chokwe  &	1M  &	Africa  &	Latn  &	\scriptsize{cjk-14l}  &	98.7  &	43.7 \\
jax  &	500K  &	Jambi Malay  &	1M  &	Asia  &	Latn  &	\scriptsize{abs-13l}  &	85.8  &	69.1 \\
bft  &	500K  &	Balti  &	300K  &	Asia  &	Arab  &	\scriptsize{bal-13l}  &	13.7  &	- \\
dov  &	500K  &	Dombe  &	2M  &	Africa  &	Latn  &	\scriptsize{dov-9l}  &	100  &	44.8 \\
kv  &	500K  &	Komi  &	200K  &	Europe  &	Cyrl  &	\scriptsize{ady-9l}  &	95.2  &	44.0 \\
cfm  &	500K  &	Falam Chin  &	100K  &	Asia  &	Latn  &	\scriptsize{bgr-6l}  &	99.9  &	36.7 \\
fay  &	500K  &	Fars Dialects  &	50K  &	Asia  &	Arab  &	\scriptsize{fay-14l}  &	60.9  &	62.3 \\
mr-Latn  &	500K  &	Marathi  &	83M  &	Asia  &	Latn  &	\scriptsize{bgp-13l}  &	98.0  &	64.5 \\
lu  &	400K  &	Luba-Katanga  &	2M  &	Africa  &	Latn  &	\scriptsize{jbu-8l}  &	98.3  &	39.1 \\
wa  &	400K  &	Walloon  &	600K  &	Europe  &	Latn  &	\scriptsize{acf-10l}  &	91.2  &	54.1 \\
yua  &	400K  &	Yucatec Maya  &	800K  &	Americas  &	Latn  &	\scriptsize{chd-6l}  &	99.3  &	57.9 \\
abv  &	400K  &	Baharna Arabic  &	300K  &	Europe  &	Arab  &	\scriptsize{abv-10l}  &	8.1  &	- \\
sg  &	400K  &	Sango  &	2M  &	Africa  &	Latn  &	\scriptsize{enx-4l}  &	99.7  &	31.0 \\
iso  &	400K  &	Isoko  &	400K  &	Africa  &	Latn  &	\scriptsize{ajg-8l}  &	99.9  &	35.1 \\
kac  &	400K  &	Southern Jinghpaw  &	900K  &	Asia  &	Latn  &	\scriptsize{ace-12l}  &	100  &	38.4 \\
syr  &	400K  &	Syriac  &	1M  &	Europe  &	Syrc  &	\scriptsize{ach-10l}  &	99.9  &	40.4 \\
pam  &	400K  &	Pampanga  &	2M  &	Asia  &	Latn  &	\scriptsize{agn-14l}  &	99.5  &	52.5 \\
afb  &	400K  &	Gulf Arabic  &	5M  &	Europe  &	Arab  &	\scriptsize{abv-10l}  &	4.1  &	- \\
qxq  &	400K  &	Qashqa'i  &	2M  &	Asia  &	Arab  &	\scriptsize{fay-14l}  &	61.8  &	61.0 \\
ada  &	400K  &	Adangme  &	800K  &	Africa  &	Latn  &	\scriptsize{ada-13l}  &	99.7  &	44.1 \\
ctg  &	400K  &	Chittagonian  &	15M  &	Asia  &	Beng  &	\scriptsize{bpy-7l}  &	90.9  &	65.2 \\
kg  &	400K  &	Kongo  &	6M  &	Africa  &	Latn  &	\scriptsize{jbu-8l}  &	93.8  &	- \\
mtr  &	400K  &	Mewari  &	5M  &	Asia  &	Deva  &	\scriptsize{bfz-15l}  &	89.2  &	44.8 \\
io  &	400K  &	Ido  &	5K  &	Europe  &	Latn  &	\scriptsize{acf-10l}  &	84.6  &	- \\
pmy  &	400K  &	Papuan Malay  &	500K  &	Asia  &	Latn  &	\scriptsize{abs-13l}  &	88.3  &	68.3 \\
ss-SZ  &	400K  &	Swati  &	3M  &	Africa  &	Latn  &	\scriptsize{gup-8l}  &	87.3  &	64.7 \\
mwr  &	400K  &	Marwari  &	8M  &	Asia  &	Deva  &	\scriptsize{bfz-15l}  &	97.3  &	- \\
ktu  &	400K  &	Kituba (DRC)  &	5M  &	Africa  &	Latn  &	\scriptsize{jbu-8l}  &	99.8  &	43.6 \\
nd  &	300K  &	Zimbabwean Ndebele  &	2M  &	Africa  &	Latn  &	\scriptsize{gup-8l}  &	98.6  &	69.1 \\
kek  &	300K  &	Kekchí  &	800K  &	Americas  &	Latn  &	\scriptsize{kek-2l}  &	100  &	35.1 \\
mfe  &	300K  &	Morisyen  &	1M  &	Africa  &	Latn  &	\scriptsize{acf-10l}  &	100  &	66.0 \\
bua  &	300K  &	Buriat  &	400K  &	Europe  &	Cyrl  &	\scriptsize{bua-2l}  &	96.9  &	44.5 \\
tvl  &	300K  &	Tuvalu  &	50K  &	Oceania  &	Latn  &	\scriptsize{mi-4l}  &	99.6  &	- \\
chk  &	300K  &	Chuukese  &	50K  &	Oceania  &	Latn  &	\scriptsize{chk-2l}  &	100  &	34.4 \\
dcc  &	300K  &	Deccan  &	13M  &	Asia  &	Arab  &	\scriptsize{bal-13l}  &	89.2  &	61.2 \\
gom  &	300K  &	Goan Konkani  &	2M  &	Asia  &	Deva  &	\scriptsize{awa-14l}  &	85.6  &	59.2 \\
pon  &	300K  &	Pohnpeian  &	50K  &	Oceania  &	Latn  &	\scriptsize{bgs-9l}  &	99.7  &	39.9 \\
av  &	300K  &	Avar  &	800K  &	Europe  &	Cyrl  &	\scriptsize{ady-9l}  &	99.1  &	45.8 \\
mfb  &	300K  &	Bangka  &	300K  &	Asia  &	Latn  &	\scriptsize{abs-13l}  &	91.5  &	66.6 \\
sjp  &	300K  &	Surjapuri  &	1M  &	Asia  &	Deva  &	\scriptsize{bho-13l}  &	96.3  &	55.4 \\
tiv  &	300K  &	Tiv  &	2M  &	Africa  &	Latn  &	\scriptsize{bim-3l}  &	100  &	27.7 \\
ady  &	300K  &	Adyghe  &	600K  &	Europe  &	Cyrl  &	\scriptsize{ady-9l}  &	89.2  &	43.2 \\
btx  &	300K  &	Batak Karo  &	600K  &	Asia  &	Latn  &	\scriptsize{ban-3l}  &	100  &	39.1 \\
wo  &	300K  &	Wolof  &	4M  &	Africa  &	Latn  &	\scriptsize{aeb-Latn-11l}  &	99.3  &	50.1 \\
tsc  &	300K  &	Tswa  &	1M  &	Africa  &	Latn  &	\scriptsize{cce-8l}  &	98.6  &	66.9 \\
hne  &	300K  &	Chhattisgarhi  &	18M  &	Asia  &	Deva  &	\scriptsize{bho-13l}  &	98.9  &	56.9 \\
ay  &	300K  &	Central-Southern Aymara  &	3M  &	Americas  &	Latn  &	\scriptsize{auc-6l}  &	99.6  &	31.0 \\
pa-Arab  &	300K  &	Eastern Punjabi  &	28M  &	Asia  &	Arab  &	\scriptsize{doi-Arab-5l}  &	96.4  &	64.3 \\
gag  &	300K  &	Gagauz  &	600K  &	Europe  &	Latn  &	\scriptsize{aeb-Latn-11l}  &	99.0  &	61.9 \\
raj  &	300K  &	Rajasthani  &	18M  &	Asia  &	Deva  &	\scriptsize{bfz-15l}  &	88.0  &	47.9 \\
quc  &	200K  &	K'iche'  &	2M  &	Americas  &	Latn  &	\scriptsize{acr-4l}  &	99.9  &	38.8 \\
zyj  &	200K  &	Youjiang Zhuang  &	900K  &	Asia  &	Latn  &	\scriptsize{cqd-8l}  &	99.1  &	26.4 \\
bhb-Gujr  &	200K  &	Bhili  &	3M  &	Asia  &	Gujr  &	\scriptsize{bhb-Gujr-3l}  &	98.2  &	- \\
ho  &	200K  &	Hiri Motu  &	100K  &	Oceania  &	Latn  &	\scriptsize{gah-11l}  &	99.4  &	32.6 \\
zap  &	200K  &	Zapotec  &	400K  &	Americas  &	Latn  &	\scriptsize{zaa-5l}  &	100  &	32.1 \\
crh  &	200K  &	Crimean Tatar  &	500K  &	Europe  &	Cyrl  &	\scriptsize{az-RU-13l}  &	99.7  &	53.1 \\
mam  &	200K  &	Mam  &	500K  &	Americas  &	Latn  &	\scriptsize{chd-6l}  &	99.8  &	28.3 \\
rhg-Latn  &	200K  &	Rohingya  &	2M  &	Asia  &	Latn  &	\scriptsize{bn-Latn-6l}  &	98.7  &	50.0 \\
rom  &	200K  &	Romani  &	4M  &	Europe  &	Latn  &	\scriptsize{rmc-3l}  &	99.1  &	44.0 \\
mgh  &	200K  &	Makhuwa-Meetto  &	7M  &	Africa  &	Latn  &	\scriptsize{apu-13l}  &	99.6  &	30.0 \\
lrc  &	200K  &	Northern Luri  &	2M  &	Asia  &	Arab  &	\scriptsize{apr-8l}  &	67.2  &	- \\
kmz-Latn  &	200K  &	Khorasan Turkic  &	400K  &	Asia  &	Latn  &	\scriptsize{aeb-Latn-11l}  &	93.0  &	54.9 \\
myv  &	200K  &	Erzya  &	400K  &	Europe  &	Cyrl  &	\scriptsize{ady-9l}  &	98.0  &	36.3 \\
ace  &	200K  &	Acehnese  &	4M  &	Asia  &	Latn  &	\scriptsize{ace-12l}  &	99.7  &	50.1 \\
bns  &	200K  &	Bundeli  &	3M  &	Asia  &	Deva  &	\scriptsize{awa-14l}  &	87.8  &	- \\
acw  &	200K  &	Hijazi Arabic  &	14M  &	Europe  &	Arab  &	\scriptsize{abv-10l}  &	6.4  &	- \\
nan-Latn-TW  &	200K  &	Min Nan Chinese  &	59M  &	Asia  &	Latn  &	\scriptsize{cdo-5l}  &	100  &	27.1 \\
quh  &	200K  &	South Bolivian Quechua  &	3M  &	Americas  &	Latn  &	\scriptsize{ae-Latn-10l}  &	99.9  &	39.2 \\
gbm  &	200K  &	Garhwali  &	3M  &	Asia  &	Deva  &	\scriptsize{bfz-15l}  &	95.1  &	53.9 \\
ctu  &	200K  &	Chol  &	200K  &	Americas  &	Latn  &	\scriptsize{chz-9l}  &	100  &	29.8 \\
acq  &	200K  &	Ta'izzi-Adeni Arabic  &	7M  &	Europe  &	Arab  &	\scriptsize{abv-10l}  &	12.3  &	- \\
bfy  &	200K  &	Bagheli  &	8M  &	Asia  &	Deva  &	\scriptsize{awa-14l}  &	91.5  &	62.0 \\
gv  &	200K  &	Manx  &	5K  &	Europe  &	Latn  &	\scriptsize{gv-3l}  &	98.3  &	36.6 \\
xmm  &	200K  &	Manado Malay  &	800K  &	Asia  &	Latn  &	\scriptsize{abs-13l}  &	94.5  &	59.7 \\
mel  &	200K  &	Central Melanau  &	100K  &	Asia  &	Latn  &	\scriptsize{abx-11l}  &	89.7  &	74.2 \\
ksw  &	200K  &	S'gaw Karen  &	1M  &	Asia  &	Mymr  &	\scriptsize{ksw-3l}  &	100  &	38.0 \\
seh  &	200K  &	Sena  &	2M  &	Africa  &	Latn  &	\scriptsize{cjk-14l}  &	99.5  &	52.5 \\
msb  &	200K  &	Masbatenyo  &	400K  &	Asia  &	Latn  &	\scriptsize{abx-11l}  &	99.4  &	57.3 \\
tsg  &	200K  &	Tausug  &	1M  &	Asia  &	Latn  &	\scriptsize{abx-11l}  &	98.4  &	54.0 \\
dtp  &	200K  &	Central Dusun  &	100K  &	Asia  &	Latn  &	\scriptsize{dtb-4l}  &	100  &	23.9 \\
yap  &	200K  &	Yapese  &	5K  &	Oceania  &	Latn  &	\scriptsize{bru-6l}  &	99.3  &	- \\
kum  &	200K  &	Kumyk  &	400K  &	Europe  &	Cyrl  &	\scriptsize{az-RU-13l}  &	99.9  &	47.9 \\
fon  &	200K  &	Fon  &	1M  &	Africa  &	Latn  &	\scriptsize{ada-13l}  &	99.9  &	40.9 \\
alt  &	200K  &	Southern Altai  &	50K  &	Europe  &	Cyrl  &	alt  &	100  &	43.4 \\
wal  &	200K  &	Wolaytta  &	2M  &	Africa  &	Latn  &	\scriptsize{drs-Latn-8l}  &	99.7  &	29.6 \\
en-Arab  &	200K  &	English  &	550M  &	Europe  &	Arab  &	\scriptsize{abv-10l}  &	97.4  &	56.5 \\
nzi  &	200K  &	Nzima  &	300K  &	Africa  &	Latn  &	\scriptsize{bov-4l}  &	100  &	40.6 \\
cak  &	200K  &	Kaqchikel  &	400K  &	Americas  &	Latn  &	\scriptsize{acr-4l}  &	100  &	24.1 \\
ban  &	200K  &	Balinese  &	3M  &	Asia  &	Latn  &	\scriptsize{ban-3l}  &	100  &	51.7 \\
zne  &	200K  &	Zande  &	1M  &	Africa  &	Latn  &	\scriptsize{agt-6l}  &	99.5  &	16.9 \\
bm  &	200K  &	Bambara  &	14M  &	Africa  &	Latn  &	\scriptsize{ada-13l}  &	99.1  &	42.1 \\
mdh  &	200K  &	Maguindanao  &	1M  &	Asia  &	Latn  &	\scriptsize{agn-14l}  &	96.4  &	43.7 \\
xal  &	200K  &	Oirad-Kalmyk-Darkhat  &	400K  &	Europe  &	Cyrl  &	\scriptsize{az-RU-13l}  &	91.5  &	42.8 \\
abs  &	200K  &	Ambonese Malay  &	2M  &	Asia  &	Latn  &	\scriptsize{abs-13l}  &	88.8  &	63.7 \\
tzo  &	200K  &	Tzotzil  &	200K  &	Americas  &	Latn  &	\scriptsize{aeb-Latn-11l}  &	99.9  &	29.5 \\
bgc  &	200K  &	Haryanvi  &	13M  &	Asia  &	Deva  &	\scriptsize{ahr-13l}  &	91.7  &	56.9 \\
vkt  &	200K  &	Tenggarong Kutai Malay  &	200K  &	Asia  &	Latn  &	\scriptsize{abs-13l}  &	93.6  &	64.8 \\
doi  &	200K  &	Kangri-Dogri  &	2M  &	Asia  &	Deva  &	\scriptsize{bfz-15l}  &	93.5  &	- \\
gez  &	200K  &	Geez  &	0K  &	Africa  &	Ethi  &	\scriptsize{gez-8l}  &	87.6  &	- \\
rwr  &	200K  &	Marwari (India)  &	8M  &	Asia  &	Deva  &	\scriptsize{bfz-15l}  &	84.9  &	44.9 \\
tum  &	200K  &	Tumbuka  &	2M  &	Africa  &	Latn  &	\scriptsize{cjk-14l}  &	47.9  &	- \\
ang  &	200K  &	Old English (ca. 450-1100)  &	0K  &	Europe  &	Latn  &	\scriptsize{ang-7l}  &	97.6  &	- \\
nia  &	200K  &	Nias  &	800K  &	Asia  &	Latn  &	\scriptsize{ceg-5l}  &	100  &	23.1 \\
bts  &	200K  &	Batak Simalungun  &	1M  &	Asia  &	Latn  &	\scriptsize{agn-14l}  &	100  &	34.1 \\
ng  &	200K  &	Ndonga  &	800K  &	Africa  &	Latn  &	\scriptsize{kj-5l}  &	95.0  &	35.0 \\
ach  &	200K  &	Acoli  &	1M  &	Africa  &	Latn  &	\scriptsize{ach-10l}  &	99.6  &	21.0 \\
rcf  &	200K  &	Réunion Creole French  &	600K  &	Africa  &	Latn  &	\scriptsize{acf-10l}  &	99.3  &	66.9 \\
kos  &	200K  &	Kosraean  &	5K  &	Oceania  &	Latn  &	\scriptsize{ceg-5l}  &	98.8  &	- \\
skg  &	200K  &	West Malagasy Sakalava  &	1M  &	Africa  &	Latn  &	\scriptsize{apy-7l}  &	93.4  &	57.6 \\
mrj  &	200K  &	Western Mari  &	5K  &	Europe  &	Cyrl  &	\scriptsize{az-RU-13l}  &	94.3  &	45.8 \\
bci  &	200K  &	Baoulé  &	2M  &	Africa  &	Latn  &	\scriptsize{ada-13l}  &	98.9  &	32.1 \\
ch  &	200K  &	Chamorro  &	50K  &	Oceania  &	Latn  &	\scriptsize{ayo-4l}  &	98.9  &	31.9 \\
qvi  &	200K  &	Imbabura Highland Quichua  &	300K  &	Americas  &	Latn  &	\scriptsize{quw-5l}  &	100  &	38.7 \\
gym  &	200K  &	Ngäbere  &	200K  &	Americas  &	Latn  &	gym  &	100  &	19.5 \\
cyo  &	150K  &	Cuyonon  &	100K  &	Asia  &	Latn  &	\scriptsize{agn-14l}  &	85.0  &	69.1 \\
stq  &	150K  &	Ems-Weser Frisian  &	5K  &	Europe  &	Latn  &	\scriptsize{apr-8l}  &	89.6  &	- \\
tcy  &	150K  &	Tulu  &	2M  &	Asia  &	Knda  &	\scriptsize{lmn-Knda-2l}  &	100  &	49.4 \\
meu  &	150K  &	Motu  &	50K  &	Oceania  &	Latn  &	\scriptsize{gah-11l}  &	99.1  &	27.4 \\
pau  &	150K  &	Palauan  &	50K  &	Oceania  &	Latn  &	pau  &	100  &	- \\
bas  &	150K  &	Basa (Cameroon)  &	300K  &	Africa  &	Latn  &	\scriptsize{bas-11l}  &	99.4  &	21.0 \\
guc  &	140K  &	Wayuu  &	300K  &	Americas  &	Latn  &	guc  &	100  &	18.5 \\
quy  &	140K  &	Ayacucho Quechua  &	900K  &	Americas  &	Latn  &	\scriptsize{ae-Latn-10l}  &	99.7  &	29.8 \\
gu-Latn  &	140K  &	Gujarati  &	56M  &	Asia  &	Latn  &	\scriptsize{bgp-13l}  &	98.1  &	60.9 \\
mad  &	140K  &	Madurese  &	7M  &	Asia  &	Latn  &	\scriptsize{ade-4l}  &	99.8  &	44.7 \\
cbk  &	140K  &	Chavacano  &	700K  &	Asia  &	Latn  &	cbk  &	100  &	- \\
awa  &	140K  &	Awadhi  &	4M  &	Asia  &	Deva  &	\scriptsize{awa-14l}  &	98.1  &	61.4 \\
brx  &	130K  &	Bodo-Mech  &	1M  &	Asia  &	Deva  &	\scriptsize{ahr-13l}  &	100  &	32.8 \\
koi  &	130K  &	Komi-Permyak  &	50K  &	Europe  &	Cyrl  &	\scriptsize{ady-9l}  &	93.7  &	40.6 \\
arn  &	130K  &	Mapudungun  &	300K  &	Americas  &	Latn  &	\scriptsize{arn-6l}  &	99.6  &	33.6 \\
nv  &	130K  &	Navajo  &	200K  &	Americas  &	Latn  &	\scriptsize{nv-2l}  &	99.2  &	24.6 \\
nhe  &	130K  &	Eastern Huasteca Nahuatl  &	400K  &	Americas  &	Latn  &	\scriptsize{azz-7l}  &	99.7  &	29.4 \\
fip  &	130K  &	Fipa  &	200K  &	Africa  &	Latn  &	\scriptsize{dov-9l}  &	100  &	32.0 \\
dyu  &	130K  &	Dyula  &	16M  &	Africa  &	Latn  &	\scriptsize{ada-13l}  &	99.8  &	27.8 \\
xnr  &	130K  &	Kangri  &	2M  &	Asia  &	Deva  &	\scriptsize{bfz-15l}  &	96.2  &	57.7 \\
kbp  &	130K  &	Kabiyé  &	1M  &	Africa  &	Latn  &	\scriptsize{kbp-2l}  &	100  &	32.1 \\
kw  &	130K  &	Cornish  &	5K  &	Europe  &	Latn  &	\scriptsize{br-4l}  &	92.6  &	- \\
kri  &	130K  &	Krio  &	4M  &	Africa  &	Latn  &	\scriptsize{cjk-14l}  &	99.6  &	61.0 \\
rn  &	130K  &	Rundi  &	9M  &	Africa  &	Latn  &	\scriptsize{cgg-15l}  &	97.5  &	- \\
mdf  &	120K  &	Moksha  &	5K  &	Europe  &	Cyrl  &	\scriptsize{ady-9l}  &	84.8  &	- \\
tzh  &	120K  &	Tzeltal  &	400K  &	Americas  &	Latn  &	\scriptsize{arn-6l}  &	99.4  &	30.9 \\
cab  &	120K  &	Garifuna  &	200K  &	Americas  &	Latn  &	\scriptsize{bch-10l}  &	100  &	18.8 \\
trp  &	120K  &	Kok Borok  &	800K  &	Asia  &	Latn  &	\scriptsize{ace-12l}  &	99.8  &	38.3 \\
mkn  &	120K  &	Kupang Malay  &	200K  &	Asia  &	Latn  &	\scriptsize{abs-13l}  &	98.6  &	53.7 \\
rki  &	120K  &	Rakhine  &	2M  &	Asia  &	Mymr  &	rki  &	99.9  &	49.2 \\
gyn  &	120K  &	Guyanese Creole English  &	600K  &	Americas  &	Latn  &	\scriptsize{bzj-9l}  &	99.4  &	71.8 \\
syl-Latn  &	120K  &	Sylheti  &	23M  &	Asia  &	Latn  &	\scriptsize{bn-Latn-6l}  &	99.1  &	- \\
max  &	120K  &	North Moluccan Malay  &	700K  &	Asia  &	Latn  &	\scriptsize{abs-13l}  &	91.3  &	- \\
izz  &	120K  &	Izi  &	500K  &	Africa  &	Latn  &	\scriptsize{anw-8l}  &	99.7  &	26.6 \\
dje  &	110K  &	Zarma-Kaado  &	2M  &	Africa  &	Latn  &	\scriptsize{ang-7l}  &	99.6  &	22.2 \\
smt  &	110K  &	Simte  &	50K  &	Asia  &	Latn  &	\scriptsize{bgs-9l}  &	99.9  &	- \\
iu  &	110K  &	Inuktitut  &	50K  &	Americas  &	Cans  &	\scriptsize{crm-3l}  &	99.8  &	26.9 \\
bmm  &	110K  &	Northern Betsimisaraka Malagasy  &	1M  &	Africa  &	Latn  &	\scriptsize{apy-7l}  &	91.8  &	- \\
lhu  &	110K  &	Lahu  &	500K  &	Asia  &	Latn  &	\scriptsize{hni-8l}  &	99.9  &	29.3 \\
djk  &	110K  &	Aukan  &	50K  &	Americas  &	Latn  &	\scriptsize{djk-3l}  &	100  &	35.3 \\
cuk  &	110K  &	San Blas Kuna  &	50K  &	Americas  &	Latn  &	\scriptsize{bnj-6l}  &	100  &	12.3 \\
mni  &	110K  &	Manipuri  &	2M  &	Asia  &	Beng  &	mni  &	99.9  &	37.9 \\
mak  &	100K  &	Makasar  &	5M  &	Asia  &	Latn  &	mak  &	100  &	19.2 \\
bto  &	100K  &	Iriga Bicolano  &	200K  &	Asia  &	Latn  &	\scriptsize{agn-14l}  &	95.2  &	64.6 \\
tab  &	100K  &	Tabasaran  &	100K  &	Europe  &	Cyrl  &	\scriptsize{ce-3l}  &	100  &	33.4 \\
mps  &	100K  &	Dadibi  &	50K  &	Oceania  &	Latn  &	\scriptsize{mps-3l}  &	99.8  &	- \\
mbt  &	100K  &	Matigsalug Manobo  &	50K  &	Asia  &	Latn  &	mbt  &	100  &	21.5 \\
gui  &	100K  &	Eastern Bolivian Guaraní  &	50K  &	Americas  &	Latn  &	gui  &	100  &	20.6 \\
fan  &	100K  &	Fang (Equatorial Guinea)  &	1M  &	Africa  &	Latn  &	\scriptsize{bas-11l}  &	99.1  &	31.0 \\
ngu  &	100K  &	Central Guerrero Nahuatl  &	200K  &	Americas  &	Latn  &	\scriptsize{azz-7l}  &	99.7  &	28.3 \\
srm  &	100K  &	Saramaccan  &	50K  &	Americas  &	Latn  &	\scriptsize{djk-3l}  &	100  &	- \\
ms-Arab-BN  &	100K  &	Malay  &	80M  &	Asia  &	Arab  &	\scriptsize{mfa-3l}  &	92.7  &	53.9 \\
ndc-ZW  &	100K  &	Ndau  &	2M  &	Africa  &	Latn  &	\scriptsize{cjk-14l}  &	99.0  &	60.5 \\
nnb  &	100K  &	Nande  &	900K  &	Africa  &	Latn  &	\scriptsize{cgg-15l}  &	99.7  &	35.7 \\
ber  &	100K  &	Berber  &	2M  &	Africa  &	Tfng  &	ber  &	100  &	26.5 \\
ibb  &	100K  &	Ibibio  &	2M  &	Africa  &	Latn  &	\scriptsize{anw-8l}  &	98.6  &	48.3 \\
nyu  &	90K  &	Nyungwe  &	400K  &	Africa  &	Latn  &	\scriptsize{cjk-14l}  &	98.0  &	- \\
gub  &	90K  &	Guajajára  &	50K  &	Americas  &	Latn  &	\scriptsize{abr-5l}  &	100  &	- \\
kmb  &	90K  &	Kimbundu  &	4M  &	Africa  &	Latn  &	\scriptsize{dov-9l}  &	100  &	25.0 \\
trw  &	90K  &	Torwali  &	50K  &	Asia  &	Arab  &	\scriptsize{bal-13l}  &	88.4  &	43.8 \\
nij  &	90K  &	Ngaju  &	900K  &	Asia  &	Latn  &	\scriptsize{kje-6l}  &	100  &	38.9 \\
cgg  &	90K  &	Chiga  &	2M  &	Africa  &	Latn  &	\scriptsize{cgg-15l}  &	99.3  &	50.5 \\
twu  &	90K  &	Termanu  &	50K  &	Asia  &	Latn  &	\scriptsize{bhw-9l}  &	100  &	19.9 \\
kn-Latn  &	90K  &	Kannada  &	43M  &	Asia  &	Latn  &	\scriptsize{bgp-13l}  &	98.6  &	46.0 \\
ms-Arab  &	90K  &	Malay  &	80M  &	Asia  &	Arab  &	\scriptsize{mfa-3l}  &	88.9  &	55.7 \\
oj  &	90K  &	Ojibwe  &	50K  &	Americas  &	Latn  &	\scriptsize{agr-3l}  &	99.9  &	- \\
acf  &	90K  &	Saint Lucian Creole French  &	200K  &	Americas  &	Latn  &	\scriptsize{acf-10l}  &	100  &	48.9 \\
cac  &	90K  &	Chuj  &	50K  &	Americas  &	Latn  &	cac  &	100  &	13.8 \\
jiv  &	90K  &	Shuar  &	50K  &	Americas  &	Latn  &	\scriptsize{bn-Latn-6l}  &	100  &	12.0 \\
alz  &	90K  &	Alur  &	2M  &	Africa  &	Latn  &	\scriptsize{adh-5l}  &	99.6  &	21.3 \\
akb  &	90K  &	Batak Angkola  &	800K  &	Asia  &	Latn  &	\scriptsize{akb-4l}  &	100  &	15.4 \\
ff  &	90K  &	Fula  &	12M  &	Africa  &	Latn  &	\scriptsize{aa-13l}  &	90.6  &	40.6 \\
aa  &	80K  &	Afar  &	2M  &	Africa  &	Latn  &	\scriptsize{aa-13l}  &	99.7  &	19.8 \\
miq  &	80K  &	Mískito  &	200K  &	Americas  &	Latn  &	\scriptsize{miq-2l}  &	100  &	19.1 \\
bgz  &	80K  &	Banggai  &	100K  &	Asia  &	Latn  &	\scriptsize{amo-7l}  &	92.7  &	56.2 \\
nyn  &	80K  &	Nyankole  &	2M  &	Africa  &	Latn  &	\scriptsize{cgg-15l}  &	99.7  &	37.7 \\
adh  &	80K  &	Adhola  &	400K  &	Africa  &	Latn  &	\scriptsize{adh-5l}  &	100  &	18.5 \\
bzj  &	80K  &	Belize Kriol English  &	200K  &	Americas  &	Latn  &	\scriptsize{bzj-9l}  &	100  &	57.9 \\
nyo  &	80K  &	Nyoro  &	700K  &	Africa  &	Latn  &	\scriptsize{cgg-15l}  &	99.3  &	46.7 \\
kfy  &	80K  &	Kumaoni  &	2M  &	Asia  &	Deva  &	\scriptsize{bfz-15l}  &	93.8  &	50.7 \\
kjb  &	80K  &	Q'anjob'al  &	50K  &	Americas  &	Latn  &	\scriptsize{cbr-6l}  &	96.7  &	14.3 \\
frp  &	80K  &	Arpitan  &	100K  &	Europe  &	Latn  &	\scriptsize{acf-10l}  &	90.4  &	- \\
jvn  &	80K  &	Caribbean Javanese  &	50K  &	Americas  &	Latn  &	jvn  &	100  &	26.9 \\
mdy  &	80K  &	Male (Ethiopia)  &	100K  &	Africa  &	Ethi  &	mdy  &	100  &	22.7 \\
hae  &	80K  &	Eastern Oromo  &	4M  &	Africa  &	Latn  &	\scriptsize{amf-6l}  &	99.5  &	48.6 \\
qxr  &	80K  &	Cañar Highland Quichua  &	200K  &	Americas  &	Latn  &	\scriptsize{quw-5l}  &	99.7  &	37.7 \\
noa  &	80K  &	Woun Meu  &	50K  &	Americas  &	Latn  &	\scriptsize{cbv-6l}  &	99.9  &	- \\
krj  &	70K  &	Kinaray-a  &	400K  &	Asia  &	Latn  &	\scriptsize{abx-11l}  &	99.5  &	58.9 \\
hui  &	70K  &	Huli  &	200K  &	Oceania  &	Latn  &	\scriptsize{hui-3l}  &	99.9  &	12.0 \\
enq  &	70K  &	Enga  &	200K  &	Oceania  &	Latn  &	\scriptsize{avt-12l}  &	99.5  &	12.1 \\
sgj  &	70K  &	Surgujia  &	1M  &	Asia  &	Deva  &	\scriptsize{bho-13l}  &	97.5  &	56.8 \\
lrl  &	70K  &	Larestani  &	200K  &	Asia  &	Arab  &	\scriptsize{fay-14l}  &	56.8  &	- \\
gof  &	70K  &	Gofa  &	400K  &	Africa  &	Ethi  &	\scriptsize{drs-Latn-8l}  &	99.6  &	24.1 \\
sxn  &	70K  &	Sangir  &	200K  &	Asia  &	Latn  &	sxn  &	100  &	31.9 \\
hus  &	70K  &	Huastec  &	100K  &	Americas  &	Latn  &	\scriptsize{chd-6l}  &	100  &	19.6 \\
tuc  &	70K  &	Mutu  &	5K  &	Oceania  &	Latn  &	\scriptsize{bmu-6l}  &	100  &	- \\
maz  &	70K  &	Central Mazahua  &	50K  &	Americas  &	Latn  &	\scriptsize{jic-3l}  &	100  &	15.7 \\
hix  &	70K  &	Hixkaryána  &	5K  &	Americas  &	Latn  &	hix  &	99.6  &	- \\
ks  &	70K  &	Kashmiri  &	6M  &	Asia  &	Arab  &	\scriptsize{doi-Arab-5l}  &	96.5  &	47.0 \\
clu  &	70K  &	Caluyanun  &	50K  &	Asia  &	Latn  &	\scriptsize{abx-11l}  &	98.9  &	- \\
acx  &	70K  &	Omani Arabic  &	1M  &	Europe  &	Arab  &	\scriptsize{abv-10l}  &	4.9  &	- \\
prk  &	70K  &	South Wa  &	800K  &	Asia  &	Latn  &	\scriptsize{ace-12l}  &	100  &	26.3 \\
mas  &	70K  &	Masai  &	1M  &	Africa  &	Latn  &	\scriptsize{big-5l}  &	99.8  &	16.2 \\
pis  &	70K  &	Pijin  &	400K  &	Oceania  &	Latn  &	\scriptsize{bi-6l}  &	100  &	21.3 \\
amu  &	70K  &	Guerrero Amuzgo  &	50K  &	Americas  &	Latn  &	amu  &	100  &	12.0 \\
myu  &	70K  &	Mundurukú  &	5K  &	Americas  &	Latn  &	\scriptsize{bch-10l}  &	99.9  &	- \\
kjg  &	70K  &	Khmu  &	700K  &	Asia  &	Laoo  &	\scriptsize{kjg-2l}  &	100  &	37.1 \\
sda  &	70K  &	Toraja-Sa'dan  &	800K  &	Asia  &	Latn  &	\scriptsize{kje-6l}  &	98.4  &	20.1 \\
kpg  &	70K  &	Kapingamarangi  &	5K  &	Oceania  &	Latn  &	kpg  &	100  &	- \\
bus  &	70K  &	Bokobaru  &	50K  &	Africa  &	Latn  &	\scriptsize{akp-4l}  &	99.5  &	16.5 \\
mag  &	70K  &	Magahi  &	28M  &	Asia  &	Deva  &	\scriptsize{bho-13l}  &	99.6  &	57.7 \\
dwr  &	70K  &	Dawro  &	500K  &	Africa  &	Latn  &	\scriptsize{drs-Latn-8l}  &	100  &	19.8 \\
sja  &	70K  &	Epena  &	5K  &	Americas  &	Latn  &	\scriptsize{mpm-4l}  &	99.1  &	- \\
ahk  &	60K  &	Akha  &	600K  &	Asia  &	Latn  &	ahk  &	100  &	26.2 \\
ffm  &	60K  &	Maasina Fulfulde  &	1M  &	Africa  &	Latn  &	\scriptsize{aa-13l}  &	98.6  &	29.0 \\
xav  &	60K  &	Xavánte  &	5K  &	Americas  &	Latn  &	xav  &	87.5  &	- \\
chr  &	60K  &	Cherokee  &	50K  &	Americas  &	Cher  &	chr  &	100  &	19.7 \\
kyz  &	60K  &	Kayabí  &	5K  &	Americas  &	Latn  &	\scriptsize{avt-12l}  &	71.1  &	- \\
tll  &	60K  &	Tetela  &	800K  &	Africa  &	Latn  &	\scriptsize{kj-5l}  &	99.9  &	24.8 \\
tee  &	60K  &	Huehuetla Tepehua  &	5K  &	Americas  &	Latn  &	\scriptsize{tee-2l}  &	99.9  &	- \\
niq  &	60K  &	Nandi  &	1M  &	Africa  &	Latn  &	\scriptsize{awk-4l}  &	98.8  &	20.0 \\
din  &	60K  &	Dinka  &	1M  &	Africa  &	Latn  &	\scriptsize{din-3l}  &	99.7  &	27.6 \\
nog  &	60K  &	Nogai  &	50K  &	Europe  &	Cyrl  &	\scriptsize{bcq-Latn-12l}  &	100  &	51.6 \\
tca  &	60K  &	Ticuna  &	50K  &	Americas  &	Latn  &	tca  &	100  &	8.6 \\
abt  &	60K  &	Ambulas  &	50K  &	Oceania  &	Latn  &	\scriptsize{abt-3l}  &	99.8  &	10.3 \\
bhw  &	60K  &	Biak  &	50K  &	Asia  &	Latn  &	\scriptsize{bhw-9l}  &	100  &	20.5 \\
dln  &	60K  &	Darlong  &	5K  &	Asia  &	Latn  &	\scriptsize{bla-7l}  &	99.9  &	- \\
xog  &	60K  &	Soga  &	3M  &	Africa  &	Latn  &	\scriptsize{cgg-15l}  &	99.4  &	56.6 \\
cbt  &	60K  &	Shawi  &	5K  &	Americas  &	Latn  &	\scriptsize{cbt-2l}  &	100  &	- \\
vep  &	60K  &	Veps  &	5K  &	Europe  &	Latn  &	\scriptsize{gaw-8l}  &	90.9  &	- \\
gil  &	60K  &	Gilbertese  &	100K  &	Oceania  &	Latn  &	\scriptsize{aaz-6l}  &	100  &	14.3 \\
tyz  &	60K  &	Tày  &	3M  &	Asia  &	Latn  &	\scriptsize{blt-Latn-4l}  &	99.5  &	22.8 \\
bqc  &	60K  &	Boko (Benin)  &	200K  &	Africa  &	Latn  &	\scriptsize{akp-4l}  &	99.8  &	19.0 \\
ksd  &	60K  &	Kuanua  &	50K  &	Oceania  &	Latn  &	\scriptsize{apb-13l}  &	99.2  &	15.4 \\
tob  &	60K  &	Toba  &	50K  &	Americas  &	Latn  &	\scriptsize{hni-8l}  &	100  &	13.3 \\
bru  &	60K  &	Eastern Bru  &	50K  &	Asia  &	Latn  &	\scriptsize{bru-6l}  &	100  &	21.4 \\
sat-Latn  &	60K  &	Santali  &	6M  &	Asia  &	Latn  &	\scriptsize{aaz-6l}  &	98.9  &	40.0 \\
msy  &	60K  &	Aruamu  &	5K  &	Oceania  &	Latn  &	\scriptsize{msy-2l}  &	99.6  &	- \\
cbu  &	60K  &	Candoshi-Shapra  &	5K  &	Americas  &	Latn  &	\scriptsize{apu-13l}  &	99.1  &	- \\
shp  &	60K  &	Shipibo-Conibo  &	50K  &	Americas  &	Latn  &	\scriptsize{apu-13l}  &	100  &	- \\
qub  &	60K  &	Huallaga Huánuco Quechua  &	50K  &	Americas  &	Latn  &	\scriptsize{qub-7l}  &	99.5  &	26.0 \\
agr  &	60K  &	Aguaruna  &	50K  &	Americas  &	Latn  &	\scriptsize{agr-3l}  &	100  &	14.6 \\
emp  &	60K  &	Northern Emberá  &	50K  &	Americas  &	Latn  &	\scriptsize{cbv-6l}  &	99.6  &	12.3 \\
suz  &	60K  &	Sunwar  &	50K  &	Asia  &	Deva  &	suz  &	100  &	17.2 \\
maa  &	60K  &	San Jerónimo Tecóatl Mazatec  &	50K  &	Americas  &	Latn  &	\scriptsize{kcg-9l}  &	100  &	- \\
ifk  &	60K  &	Tuwali Ifugao  &	50K  &	Asia  &	Latn  &	\scriptsize{bnc-8l}  &	100  &	19.9 \\
arc  &	50K  &	Official Aramaic (700-300 BCE)  &	0K  &	Europe  &	Syrc  &	\scriptsize{ach-10l}  &	80.0  &	- \\
bim  &	50K  &	Bimoba  &	400K  &	Africa  &	Latn  &	\scriptsize{bim-3l}  &	100  &	15.6 \\
otq  &	50K  &	Querétaro Otomi  &	50K  &	Americas  &	Latn  &	\scriptsize{ann-3l}  &	99.9  &	47.0 \\
tdx  &	50K  &	Tandroy Malagasy  &	1M  &	Africa  &	Latn  &	\scriptsize{apy-7l}  &	99.6  &	48.2 \\
tav  &	50K  &	Tatuyo  &	5K  &	Americas  &	Latn  &	\scriptsize{des-4l}  &	100  &	- \\
pir  &	50K  &	Wa'ikhana  &	5K  &	Americas  &	Latn  &	\scriptsize{des-4l}  &	99.6  &	- \\
ded  &	50K  &	Dedua  &	5K  &	Oceania  &	Latn  &	\scriptsize{bru-6l}  &	99.7  &	- \\
txu  &	50K  &	Kayapó  &	5K  &	Americas  &	Latn  &	\scriptsize{apn-3l}  &	99.2  &	- \\
qvh  &	50K  &	Huamalíes-Dos de Mayo Huánuco Quechua  &	50K  &	Americas  &	Latn  &	\scriptsize{qub-7l}  &	100  &	19.7 \\
mav  &	50K  &	Sateré-Mawé  &	5K  &	Americas  &	Latn  &	\scriptsize{bwu-4l}  &	100  &	- \\
ubu  &	50K  &	Umbu-Ungu  &	50K  &	Oceania  &	Latn  &	\scriptsize{gbi-8l}  &	99.9  &	11.8 \\
toj  &	50K  &	Tojolabal  &	50K  &	Americas  &	Latn  &	\scriptsize{toj-4l}  &	100  &	27.2 \\
inb  &	50K  &	Inga  &	50K  &	Americas  &	Latn  &	inb  &	100  &	- \\
aoj  &	50K  &	Mufian  &	50K  &	Oceania  &	Latn  &	aoj  &	99.6  &	- \\
mbb  &	50K  &	Western Bukidnon Manobo  &	50K  &	Asia  &	Latn  &	\scriptsize{mbb-3l}  &	85.9  &	- \\
acr  &	50K  &	Achi  &	50K  &	Americas  &	Latn  &	\scriptsize{acr-4l}  &	99.9  &	21.1 \\
xtd  &	50K  &	Diuxi-Tilantongo Mixtec  &	5K  &	Americas  &	Latn  &	\scriptsize{mpm-4l}  &	99.9  &	- \\
kjh  &	50K  &	Khakas  &	50K  &	Europe  &	Cyrl  &	kjh  &	100  &	37.9 \\
dyi  &	50K  &	Djimini Senoufo  &	50K  &	Africa  &	Latn  &	dyi  &	100  &	18.5 \\
mpm  &	50K  &	Yosondúa Mixtec  &	50K  &	Americas  &	Latn  &	\scriptsize{mpm-4l}  &	99.7  &	- \\
kwi  &	50K  &	Awa-Cuaiquer  &	50K  &	Americas  &	Latn  &	\scriptsize{bla-7l}  &	100  &	- \\
bjj  &	50K  &	Kanauji  &	10M  &	Asia  &	Deva  &	\scriptsize{awa-14l}  &	97.8  &	55.4 \\
amn  &	50K  &	Amanab  &	5K  &	Oceania  &	Latn  &	\scriptsize{amn-2l}  &	99.9  &	- \\
plu  &	50K  &	Palikúr  &	5K  &	Americas  &	Latn  &	plu  &	98.7  &	- \\
ood  &	50K  &	Tohono O'odham  &	50K  &	Americas  &	Latn  &	\scriptsize{alq-4l}  &	100  &	- \\
czt  &	50K  &	Zotung Chin  &	50K  &	Asia  &	Latn  &	czt  &	100  &	19.9 \\
tbg  &	50K  &	North Tairora  &	50K  &	Oceania  &	Latn  &	\scriptsize{amp-12l}  &	100  &	- \\
gjk  &	50K  &	Kachi Koli  &	600K  &	Asia  &	Arab  &	\scriptsize{bal-13l}  &	77.3  &	59.6 \\
crn  &	50K  &	El Nayar Cora  &	50K  &	Americas  &	Latn  &	\scriptsize{cok-2l}  &	99.7  &	- \\
cni  &	50K  &	Asháninka  &	50K  &	Americas  &	Latn  &	\scriptsize{apu-13l}  &	100  &	13.0 \\
qup  &	50K  &	Southern Pastaza Quechua  &	1M  &	Americas  &	Latn  &	\scriptsize{qup-2l}  &	99.7  &	- \\
ify  &	50K  &	Keley-i Kallahan  &	50K  &	Asia  &	Latn  &	ify  &	99.9  &	- \\
srq  &	50K  &	Sirionó  &	5K  &	Americas  &	Latn  &	\scriptsize{srq-3l}  &	98.9  &	- \\
tbz  &	50K  &	Ditammari  &	200K  &	Africa  &	Latn  &	\scriptsize{gej-4l}  &	100  &	16.1 \\
aeb  &	50K  &	Tunisian Arabic  &	11M  &	Africa  &	Arab  &	\scriptsize{abv-10l}  &	13.1  &	- \\
sus  &	50K  &	Susu  &	1M  &	Africa  &	Latn  &	\scriptsize{bas-11l}  &	100  &	20.5 \\
itv  &	50K  &	Itawit  &	100K  &	Asia  &	Latn  &	\scriptsize{bnj-6l}  &	99.9  &	28.9 \\
cce  &	50K  &	Chopi  &	800K  &	Africa  &	Latn  &	\scriptsize{cce-8l}  &	99.4  &	33.3 \\
qwh  &	50K  &	Huaylas Ancash Quechua  &	300K  &	Americas  &	Latn  &	\scriptsize{nhg-3l}  &	100  &	25.4 \\
teo  &	50K  &	Teso  &	2M  &	Africa  &	Latn  &	\scriptsize{bla-7l}  &	100  &	14.4 \\
lon  &	50K  &	Malawi Lomwe  &	2M  &	Africa  &	Latn  &	\scriptsize{cjk-14l}  &	99.5  &	- \\
gun  &	50K  &	Mbyá Guaraní  &	50K  &	Americas  &	Latn  &	\scriptsize{gn-6l}  &	99.4  &	- \\
ppk  &	50K  &	Uma  &	50K  &	Asia  &	Latn  &	\scriptsize{pmf-2l}  &	100  &	- \\
ann  &	50K  &	Obolo  &	200K  &	Africa  &	Latn  &	\scriptsize{ann-3l}  &	100  &	26.3 \\
laj  &	50K  &	Lango (Uganda)  &	2M  &	Africa  &	Latn  &	\scriptsize{adh-5l}  &	99.7  &	25.7 \\
eza  &	50K  &	Ezaa  &	600K  &	Africa  &	Latn  &	\scriptsize{anw-8l}  &	99.7  &	29.2 \\
caf  &	50K  &	Southern Carrier  &	5K  &	Americas  &	Latn  &	\scriptsize{bpr-7l}  &	99.9  &	- \\
lex  &	50K  &	Luang  &	50K  &	Asia  &	Latn  &	\scriptsize{kje-6l}  &	100  &	- \\
msm  &	50K  &	Agusan Manobo  &	50K  &	Asia  &	Latn  &	\scriptsize{atd-4l}  &	100  &	24.6 \\
tuf  &	50K  &	Central Tunebo  &	5K  &	Americas  &	Latn  &	tuf  &	100  &	- \\
quf  &	50K  &	Lambayeque Quechua  &	50K  &	Americas  &	Latn  &	\scriptsize{ae-Latn-10l}  &	99.5  &	- \\
bzc  &	50K  &	Southern Betsimisaraka Malagasy  &	2M  &	Africa  &	Latn  &	\scriptsize{apy-7l}  &	96.4  &	61.4 \\
jbo  &	50K  &	Lojban  &	0K  &	Europe  &	Latn  &	\scriptsize{bi-6l}  &	98.0  &	- \\
yss  &	50K  &	Yessan-Mayo  &	5K  &	Oceania  &	Latn  &	\scriptsize{bom-5l}  &	99.9  &	- \\
tzj  &	50K  &	Tz'utujil  &	50K  &	Americas  &	Latn  &	tzj  &	100  &	13.4 \\
qvc  &	40K  &	Cajamarca Quechua  &	50K  &	Americas  &	Latn  &	qvc  &	100  &	16.4 \\
stp  &	40K  &	Southeastern Tepehuan  &	50K  &	Americas  &	Latn  &	\scriptsize{cbr-6l}  &	100  &	- \\
wsk  &	40K  &	Waskia  &	50K  &	Oceania  &	Latn  &	\scriptsize{aom-10l}  &	97.7  &	- \\
knj  &	40K  &	Akateko  &	50K  &	Americas  &	Latn  &	\scriptsize{knj-8l}  &	99.3  &	15.8 \\
qvz  &	40K  &	Northern Pastaza Quichua  &	50K  &	Americas  &	Latn  &	\scriptsize{quw-5l}  &	100  &	- \\
pad  &	40K  &	Paumari  &	5K  &	Americas  &	Latn  &	\scriptsize{anw-8l}  &	100  &	- \\
cbi  &	40K  &	Cha'palaa  &	5K  &	Americas  &	Latn  &	cbi  &	100  &	- \\
tac  &	40K  &	Lowland Tarahumara  &	50K  &	Americas  &	Latn  &	\scriptsize{alp-2l}  &	99.9  &	- \\
ngl  &	40K  &	Mozambique Lomwe  &	2M  &	Africa  &	Latn  &	\scriptsize{cjk-14l}  &	96.2  &	- \\
kaq  &	40K  &	Capanahua  &	50K  &	Americas  &	Latn  &	\scriptsize{bgs-9l}  &	100  &	- \\
jam  &	40K  &	Jamaican Creole English  &	3M  &	Americas  &	Latn  &	\scriptsize{bzj-9l}  &	99.9  &	57.9 \\
cpy  &	40K  &	South Ucayali Ashéninka  &	5K  &	Americas  &	Latn  &	\scriptsize{apu-13l}  &	99.2  &	15.0 \\
en-Cyrl  &	40K  &	English  &	550M  &	Europe  &	Cyrl  &	\scriptsize{az-RU-13l}  &	98.0  &	87.3 \\
car  &	40K  &	Galibi Carib  &	5K  &	Americas  &	Latn  &	car  &	100  &	- \\
gju  &	40K  &	Gujari  &	1M  &	Asia  &	Arab  &	\scriptsize{doi-Arab-5l}  &	98.9  &	51.3 \\
pab  &	40K  &	Parecís  &	5K  &	Americas  &	Latn  &	\scriptsize{pab-2l}  &	99.9  &	- \\
yuj  &	40K  &	Karkar-Yuri  &	5K  &	Oceania  &	Latn  &	\scriptsize{amp-12l}  &	98.0  &	- \\
jac  &	40K  &	Popti'  &	5K  &	Americas  &	Latn  &	\scriptsize{ayo-4l}  &	98.1  &	- \\
mjc  &	40K  &	San Juan Colorado Mixtec  &	50K  &	Americas  &	Latn  &	\scriptsize{cta-4l}  &	99.6  &	- \\
med  &	40K  &	Melpa  &	100K  &	Oceania  &	Latn  &	\scriptsize{cwt-4l}  &	100  &	13.6 \\
qvs  &	40K  &	San Martín Quechua  &	50K  &	Americas  &	Latn  &	\scriptsize{qup-2l}  &	99.7  &	- \\
urb  &	40K  &	Urubú-Kaapor  &	5K  &	Americas  &	Latn  &	\scriptsize{bhw-9l}  &	99.7  &	- \\
bnp  &	40K  &	Bola  &	50K  &	Oceania  &	Latn  &	\scriptsize{bch-10l}  &	99.0  &	- \\
yle  &	40K  &	Yele  &	5K  &	Oceania  &	Latn  &	\scriptsize{bib-2l}  &	100  &	- \\
acu  &	40K  &	Achuar-Shiwiar  &	5K  &	Americas  &	Latn  &	\scriptsize{acu-8l}  &	99.1  &	- \\
cof  &	40K  &	Tsafiki  &	5K  &	Americas  &	Latn  &	\scriptsize{bmr-3l}  &	100  &	- \\
poh  &	40K  &	Poqomchi'  &	50K  &	Americas  &	Latn  &	\scriptsize{hni-8l}  &	100  &	12.6 \\
gbi  &	40K  &	Galela  &	50K  &	Asia  &	Latn  &	\scriptsize{gbi-8l}  &	100  &	13.6 \\
kto  &	40K  &	Kuot  &	5K  &	Oceania  &	Latn  &	\scriptsize{bkv-4l}  &	99.9  &	- \\
bfz  &	40K  &	Mahasu Pahari  &	1M  &	Asia  &	Deva  &	\scriptsize{bfz-15l}  &	94.4  &	- \\
cao  &	40K  &	Chácobo  &	5K  &	Americas  &	Latn  &	\scriptsize{auc-6l}  &	99.7  &	- \\
myy  &	40K  &	Macuna  &	5K  &	Americas  &	Latn  &	\scriptsize{cub-2l}  &	100  &	- \\
aaz  &	40K  &	Amarasi  &	50K  &	Asia  &	Latn  &	\scriptsize{aaz-6l}  &	100  &	15.3 \\
nch  &	40K  &	Central Huasteca Nahuatl  &	200K  &	Americas  &	Latn  &	\scriptsize{azz-7l}  &	95.7  &	19.2 \\
ixl  &	40K  &	Ixil  &	100K  &	Americas  &	Latn  &	\scriptsize{ctp-4l}  &	100  &	15.1 \\
rwo  &	40K  &	Rawa  &	50K  &	Oceania  &	Latn  &	\scriptsize{gbi-8l}  &	99.9  &	- \\
azg  &	40K  &	San Pedro Amuzgos Amuzgo  &	50K  &	Americas  &	Latn  &	\scriptsize{azg-4l}  &	100  &	- \\
yaq  &	40K  &	Yaqui  &	50K  &	Americas  &	Latn  &	yaq  &	99.6  &	- \\
iou  &	40K  &	Tuma-Irumu  &	5K  &	Oceania  &	Latn  &	\scriptsize{awx-2l}  &	97.0  &	- \\
qvw  &	40K  &	Huaylla Wanca Quechua  &	300K  &	Americas  &	Latn  &	\scriptsize{qub-7l}  &	100  &	16.7 \\
snn  &	40K  &	Siona-Tetete  &	5K  &	Americas  &	Latn  &	\scriptsize{azg-4l}  &	100  &	- \\
sxu  &	40K  &	Central East Middle German  &	2M  &	Europe  &	Latn  &	\scriptsize{ang-7l}  &	99.9  &	64.3 \\
bch  &	40K  &	Bariai  &	5K  &	Oceania  &	Latn  &	\scriptsize{bch-10l}  &	98.8  &	- \\
bud  &	40K  &	Ntcham  &	300K  &	Africa  &	Latn  &	\scriptsize{amf-6l}  &	100  &	17.2 \\
tih  &	40K  &	Timugon Murut  &	5K  &	Asia  &	Latn  &	\scriptsize{tih-2l}  &	99.8  &	- \\
mcd  &	40K  &	Sharanahua  &	5K  &	Americas  &	Latn  &	\scriptsize{cbt-2l}  &	100  &	- \\
apn  &	40K  &	Apinayé  &	5K  &	Americas  &	Latn  &	\scriptsize{apn-3l}  &	91.4  &	- \\
yaa  &	40K  &	Yaminahua  &	5K  &	Americas  &	Latn  &	yaa  &	100  &	- \\
mqy  &	40K  &	Manggarai  &	900K  &	Asia  &	Latn  &	\scriptsize{abs-13l}  &	97.8  &	48.9 \\
yka  &	40K  &	Yakan  &	100K  &	Asia  &	Latn  &	\scriptsize{bpr-7l}  &	99.8  &	23.6 \\
rmc  &	40K  &	Central Romani  &	100K  &	Europe  &	Latn  &	\scriptsize{rmc-3l}  &	100  &	25.7 \\
tfr  &	40K  &	Teribe  &	5K  &	Americas  &	Latn  &	\scriptsize{pxm-2l}  &	100  &	- \\
tna  &	40K  &	Tacana  &	5K  &	Americas  &	Latn  &	\scriptsize{nhg-3l}  &	100  &	- \\
nhy  &	40K  &	Northern Oaxaca Nahuatl  &	50K  &	Americas  &	Latn  &	\scriptsize{nhy-3l}  &	99.7  &	- \\
nfa  &	40K  &	Dhao  &	5K  &	Asia  &	Latn  &	\scriptsize{bhw-9l}  &	99.1  &	- \\
zos  &	40K  &	Francisco León Zoque  &	50K  &	Americas  &	Latn  &	\scriptsize{ayo-4l}  &	98.4  &	- \\
zpz  &	40K  &	Texmelucan Zapotec  &	5K  &	Americas  &	Latn  &	\scriptsize{srq-3l}  &	100  &	- \\
mqj  &	40K  &	Mamasa  &	50K  &	Asia  &	Latn  &	\scriptsize{kje-6l}  &	97.9  &	14.5 \\
xsi  &	40K  &	Sio  &	5K  &	Oceania  &	Latn  &	\scriptsize{kij-2l}  &	99.9  &	- \\
srr  &	40K  &	Sereer  &	1M  &	Africa  &	Latn  &	\scriptsize{aa-13l}  &	81.6  &	21.4 \\
hac  &	40K  &	Gurani  &	400K  &	Asia  &	Arab  &	\scriptsize{fay-14l}  &	68.6  &	52.1 \\
yal  &	40K  &	Yalunka  &	100K  &	Africa  &	Latn  &	\scriptsize{bas-11l}  &	100  &	16.4 \\
aey  &	40K  &	Amele  &	5K  &	Oceania  &	Latn  &	\scriptsize{aey-6l}  &	97.3  &	- \\
lep-Tibt  &	40K  &	Lepcha  &	50K  &	Asia  &	Tibt  &	lep-Tibt  &	100  &	9.9 \\
kia  &	40K  &	Kim  &	50K  &	Africa  &	Latn  &	\scriptsize{cut-5l}  &	99.9  &	- \\
sef  &	40K  &	Cebaara Senoufo  &	3M  &	Africa  &	Latn  &	\scriptsize{ada-13l}  &	98.7  &	41.1 \\
lac  &	40K  &	Lacandon  &	5K  &	Americas  &	Latn  &	lac  &	99.8  &	- \\
rro  &	40K  &	Waima  &	50K  &	Oceania  &	Latn  &	\scriptsize{apb-13l}  &	97.7  &	- \\
cav  &	40K  &	Cavineña  &	5K  &	Americas  &	Latn  &	\scriptsize{aak-3l}  &	99.9  &	- \\
bvz  &	40K  &	Bauzi  &	5K  &	Asia  &	Latn  &	\scriptsize{adz-3l}  &	100  &	- \\
qxh  &	40K  &	Panao Huánuco Quechua  &	50K  &	Americas  &	Latn  &	\scriptsize{qub-7l}  &	99.5  &	17.3 \\
mcf  &	40K  &	Matsés  &	5K  &	Americas  &	Latn  &	\scriptsize{con-3l}  &	100  &	- \\
gvl  &	40K  &	Gulay  &	200K  &	Africa  &	Latn  &	\scriptsize{dzg-5l}  &	92.6  &	- \\
alq  &	40K  &	Algonquin  &	5K  &	Americas  &	Latn  &	\scriptsize{alq-4l}  &	99.9  &	- \\
gum  &	40K  &	Guambiano  &	50K  &	Americas  &	Latn  &	\scriptsize{cjp-4l}  &	100  &	- \\
cr-Latn  &	40K  &	Cree  &	200K  &	Americas  &	Latn  &	\scriptsize{ach-10l}  &	24.2  &	- \\
guh  &	40K  &	Guahibo  &	50K  &	Americas  &	Latn  &	\scriptsize{acu-8l}  &	100  &	10.8 \\
sri  &	40K  &	Siriano  &	5K  &	Americas  &	Latn  &	\scriptsize{cbv-6l}  &	99.2  &	- \\
kvn  &	40K  &	Border Kuna  &	50K  &	Americas  &	Latn  &	\scriptsize{guo-2l}  &	100  &	- \\
acn  &	40K  &	Longchuan Achang  &	50K  &	Asia  &	Latn  &	\scriptsize{acn-11l}  &	100  &	14.4 \\
mxb  &	40K  &	Tezoatlán Mixtec  &	50K  &	Americas  &	Latn  &	\scriptsize{kcg-9l}  &	98.9  &	- \\
ppo  &	40K  &	Folopa  &	5K  &	Oceania  &	Latn  &	\scriptsize{pab-2l}  &	98.2  &	- \\
atd  &	40K  &	Ata Manobo  &	50K  &	Asia  &	Latn  &	\scriptsize{atd-4l}  &	70.9  &	- \\
zam  &	40K  &	Cuixtla-Xitla Zapotec  &	5K  &	Americas  &	Latn  &	\scriptsize{srq-3l}  &	100  &	- \\
mti  &	40K  &	Maiwa (Papua New Guinea)  &	5K  &	Oceania  &	Latn  &	\scriptsize{dgz-5l}  &	99.8  &	- \\
jic  &	40K  &	Tol  &	5K  &	Americas  &	Latn  &	\scriptsize{jic-3l}  &	100  &	- \\
cui  &	40K  &	Cuiba  &	5K  &	Americas  &	Latn  &	\scriptsize{acu-8l}  &	100  &	- \\
ata  &	40K  &	Pele-Ata  &	5K  &	Oceania  &	Latn  &	\scriptsize{ata-5l}  &	98.8  &	- \\
cnk  &	40K  &	Khumi Chin  &	50K  &	Asia  &	Latn  &	\scriptsize{cdo-5l}  &	100  &	16.9 \\
bnc  &	40K  &	Bontok  &	50K  &	Asia  &	Latn  &	\scriptsize{bnc-8l}  &	99.6  &	15.5 \\
con  &	40K  &	Cofán  &	5K  &	Americas  &	Latn  &	\scriptsize{con-3l}  &	100  &	- \\
guo  &	40K  &	Guayabero  &	5K  &	Americas  &	Latn  &	\scriptsize{guo-2l}  &	100  &	- \\
caa  &	40K  &	Chortí  &	50K  &	Americas  &	Latn  &	caa  &	100  &	12.5 \\
toh  &	30K  &	Gitonga  &	400K  &	Africa  &	Latn  &	\scriptsize{cce-8l}  &	99.9  &	29.9 \\
xon  &	30K  &	Konkomba  &	900K  &	Africa  &	Latn  &	\scriptsize{daa-3l}  &	98.6  &	17.3 \\
rug  &	30K  &	Roviana  &	5K  &	Oceania  &	Latn  &	\scriptsize{akb-4l}  &	99.9  &	- \\
bbb  &	30K  &	Barai  &	5K  &	Oceania  &	Latn  &	\scriptsize{asg-10l}  &	99.6  &	- \\
chf  &	30K  &	Tabasco Chontal  &	50K  &	Americas  &	Latn  &	\scriptsize{chd-6l}  &	100  &	20.3 \\
cub  &	30K  &	Cubeo  &	5K  &	Americas  &	Latn  &	\scriptsize{cub-2l}  &	100  &	- \\
tke  &	30K  &	Takwane  &	2M  &	Africa  &	Latn  &	tke  &	97.3  &	- \\
ape  &	30K  &	Bukiyip  &	50K  &	Oceania  &	Latn  &	\scriptsize{aon-13l}  &	69.9  &	- \\
tuo  &	30K  &	Tucano  &	5K  &	Americas  &	Latn  &	\scriptsize{meh-3l}  &	99.3  &	- \\
npl  &	30K  &	Nahuatl, Southeastern Puebla  &	300K  &	Americas  &	Latn  &	\scriptsize{nhy-3l}  &	94.5  &	- \\
mmx  &	30K  &	Madak  &	5K  &	Oceania  &	Latn  &	\scriptsize{avt-12l}  &	98.4  &	- \\
ifa  &	30K  &	Amganad Ifugao  &	50K  &	Asia  &	Latn  &	\scriptsize{bnc-8l}  &	99.7  &	- \\
boj  &	30K  &	Anjam  &	5K  &	Oceania  &	Latn  &	\scriptsize{boj-4l}  &	100  &	- \\
ter  &	30K  &	Terena-Kinikinao-Chane  &	50K  &	Americas  &	Latn  &	\scriptsize{enx-4l}  &	100  &	- \\
mpx  &	30K  &	Misima-Paneati  &	50K  &	Oceania  &	Latn  &	\scriptsize{aui-14l}  &	97.5  &	- \\
msk  &	30K  &	Mansaka  &	50K  &	Asia  &	Latn  &	\scriptsize{agn-14l}  &	100  &	34.2 \\
sdh  &	30K  &	Southern Kurdish  &	3M  &	Asia  &	Arab  &	\scriptsize{fay-14l}  &	70.2  &	- \\
ber-Arab  &	30K  &	Berber  &	2M  &	Africa  &	Arab  &	\scriptsize{abv-10l}  &	99.5  &	21.1 \\
zia  &	30K  &	Zia  &	5K  &	Oceania  &	Latn  &	\scriptsize{avt-12l}  &	96.2  &	- \\
mna  &	30K  &	Mbula  &	5K  &	Oceania  &	Latn  &	\scriptsize{bmu-6l}  &	99.0  &	- \\
apy  &	30K  &	Apalaí  &	5K  &	Americas  &	Latn  &	\scriptsize{apy-7l}  &	99.2  &	- \\
kdi  &	30K  &	Kumam  &	200K  &	Africa  &	Latn  &	\scriptsize{adh-5l}  &	99.9  &	15.8 \\
aon  &	30K  &	Bumbita Arapesh  &	5K  &	Oceania  &	Latn  &	\scriptsize{aon-13l}  &	76.9  &	- \\
mpg  &	30K  &	Marba  &	200K  &	Africa  &	Latn  &	\scriptsize{agt-6l}  &	100  &	13.1 \\
mlp  &	30K  &	Bargam  &	5K  &	Oceania  &	Latn  &	\scriptsize{hni-8l}  &	98.4  &	- \\
kyc  &	30K  &	Kyaka  &	50K  &	Oceania  &	Latn  &	\scriptsize{avt-12l}  &	99.2  &	- \\
huu  &	30K  &	Murui Huitoto  &	5K  &	Americas  &	Latn  &	\scriptsize{aak-3l}  &	100  &	- \\
sll  &	30K  &	Salt-Yui  &	5K  &	Oceania  &	Latn  &	\scriptsize{aon-13l}  &	97.6  &	- \\
quw  &	30K  &	Tena Lowland Quichua  &	1M  &	Americas  &	Latn  &	\scriptsize{quw-5l}  &	99.9  &	- \\
bkq  &	30K  &	Bakairí  &	5K  &	Americas  &	Latn  &	\scriptsize{abr-5l}  &	99.2  &	- \\
yaf  &	30K  &	Yaka-Pelende-Lonzo  &	900K  &	Africa  &	Latn  &	\scriptsize{jbu-8l}  &	99.6  &	27.3 \\
ncl  &	30K  &	Michoacán Nahuatl  &	5K  &	Americas  &	Latn  &	\scriptsize{azz-7l}  &	98.8  &	- \\
sgb  &	30K  &	Mag-Anchi Ayta  &	5K  &	Asia  &	Latn  &	\scriptsize{kqp-4l}  &	100  &	- \\
rop  &	30K  &	Kriol  &	5K  &	Oceania  &	Latn  &	\scriptsize{bg-Latn-5l}  &	99.7  &	- \\
rkb  &	30K  &	Rikbaktsa  &	5K  &	Americas  &	Latn  &	\scriptsize{apy-7l}  &	99.2  &	- \\
kqc  &	30K  &	Doromu-Koki  &	5K  &	Oceania  &	Latn  &	\scriptsize{adz-3l}  &	99.5  &	- \\
kew  &	30K  &	West Kewa  &	50K  &	Oceania  &	Latn  &	\scriptsize{aau-14l}  &	95.4  &	11.5 \\
snp  &	30K  &	Siane  &	50K  &	Oceania  &	Latn  &	\scriptsize{knj-8l}  &	100  &	- \\
cjp  &	30K  &	Cabécar  &	5K  &	Americas  &	Latn  &	\scriptsize{cjp-4l}  &	100  &	- \\
viv  &	30K  &	Iduna  &	5K  &	Oceania  &	Latn  &	\scriptsize{aui-14l}  &	99.5  &	- \\
sim  &	30K  &	Mende (Papua New Guinea)  &	5K  &	Oceania  &	Latn  &	\scriptsize{apb-13l}  &	98.2  &	- \\
xsm  &	30K  &	Kasem  &	200K  &	Africa  &	Latn  &	\scriptsize{lia-3l}  &	100  &	12.2 \\
iws  &	30K  &	Sepik Iwam  &	5K  &	Oceania  &	Latn  &	\scriptsize{avt-12l}  &	98.8  &	- \\
amm  &	30K  &	Ama (Papua New Guinea)  &	5K  &	Oceania  &	Latn  &	amm  &	99.4  &	- \\
opm  &	30K  &	Oksapmin  &	5K  &	Oceania  &	Latn  &	\scriptsize{agg-5l}  &	98.8  &	- \\
kgk  &	30K  &	Kaiwá  &	50K  &	Americas  &	Latn  &	\scriptsize{gn-6l}  &	98.8  &	- \\
mta  &	30K  &	Cotabato Manobo  &	50K  &	Asia  &	Latn  &	mta  &	100  &	17.0 \\
bef  &	30K  &	Benabena  &	50K  &	Oceania  &	Latn  &	\scriptsize{avt-12l}  &	99.7  &	12.2 \\
ura  &	30K  &	Urarina  &	5K  &	Americas  &	Latn  &	\scriptsize{bmr-3l}  &	100  &	- \\
tim  &	30K  &	Timbe  &	50K  &	Oceania  &	Latn  &	\scriptsize{bom-5l}  &	99.9  &	- \\
ife  &	30K  &	Ifè  &	100K  &	Africa  &	Latn  &	ife  &	99.9  &	18.7 \\
gqr  &	30K  &	Gor  &	50K  &	Africa  &	Latn  &	\scriptsize{dzg-5l}  &	93.0  &	- \\
haz  &	30K  &	Hazaragi  &	2M  &	Asia  &	Arab  &	\scriptsize{fay-14l}  &	13.9  &	57.8 \\
row  &	30K  &	Dela-Oenale  &	5K  &	Asia  &	Latn  &	\scriptsize{bhw-9l}  &	98.6  &	- \\
lcm  &	30K  &	Tungag  &	50K  &	Oceania  &	Latn  &	\scriptsize{apb-13l}  &	97.0  &	- \\
wos  &	30K  &	Hanga Hundi  &	5K  &	Oceania  &	Latn  &	\scriptsize{aau-14l}  &	99.5  &	- \\
agn  &	30K  &	Agutaynen  &	50K  &	Asia  &	Latn  &	\scriptsize{agn-14l}  &	99.9  &	- \\
lew  &	30K  &	Ledo Kaili  &	400K  &	Asia  &	Latn  &	\scriptsize{kzf-3l}  &	100  &	18.0 \\
mbc  &	30K  &	Macushi  &	50K  &	Americas  &	Latn  &	mbc  &	100  &	- \\
byx  &	30K  &	Qaqet  &	5K  &	Oceania  &	Latn  &	\scriptsize{acn-11l}  &	97.7  &	- \\
apz  &	30K  &	Safeyoka  &	5K  &	Oceania  &	Latn  &	\scriptsize{amp-12l}  &	99.2  &	- \\
qxn  &	30K  &	Northern Conchucos Ancash Quechua  &	200K  &	Americas  &	Latn  &	\scriptsize{qub-7l}  &	99.1  &	18.2 \\
rgu  &	30K  &	Ringgou  &	50K  &	Asia  &	Latn  &	\scriptsize{bhw-9l}  &	98.0  &	- \\
for  &	30K  &	Fore  &	50K  &	Oceania  &	Latn  &	\scriptsize{awb-4l}  &	99.7  &	- \\
sny  &	30K  &	Saniyo-Hiyewe  &	5K  &	Oceania  &	Latn  &	\scriptsize{amp-12l}  &	97.1  &	- \\
mop  &	30K  &	Mopán Maya  &	5K  &	Americas  &	Latn  &	\scriptsize{chd-6l}  &	100  &	- \\
men  &	30K  &	Mende (Sierra Leone)  &	3M  &	Africa  &	Latn  &	men  &	100  &	20.8 \\
arl  &	30K  &	Arabela  &	5K  &	Americas  &	Latn  &	arl  &	100  &	- \\
hvn  &	30K  &	Hawu  &	100K  &	Asia  &	Latn  &	\scriptsize{aom-10l}  &	100  &	15.8 \\
bba  &	30K  &	Baatonum  &	600K  &	Africa  &	Latn  &	\scriptsize{bas-11l}  &	100  &	17.0 \\
des  &	30K  &	Desano  &	5K  &	Americas  &	Latn  &	\scriptsize{des-4l}  &	99.9  &	- \\
tue  &	30K  &	Tuyuca  &	5K  &	Americas  &	Latn  &	\scriptsize{cbc-2l}  &	100  &	- \\
ino  &	30K  &	Inoke-Yate  &	50K  &	Oceania  &	Latn  &	\scriptsize{ata-5l}  &	100  &	- \\
tod  &	30K  &	Toma  &	200K  &	Africa  &	Latn  &	\scriptsize{moa-2l}  &	100  &	16.9 \\
ots  &	30K  &	Estado de México Otomi  &	50K  &	Americas  &	Latn  &	\scriptsize{kcg-9l}  &	99.9  &	- \\
bpr  &	30K  &	Koronadal Blaan  &	200K  &	Asia  &	Latn  &	\scriptsize{bpr-7l}  &	97.3  &	13.6 \\
aau  &	30K  &	Abau  &	5K  &	Oceania  &	Latn  &	\scriptsize{aau-14l}  &	98.2  &	- \\
khz  &	30K  &	Keapara  &	50K  &	Oceania  &	Latn  &	\scriptsize{gah-11l}  &	97.6  &	- \\
bum  &	30K  &	Bulu (Cameroon)  &	900K  &	Africa  &	Latn  &	\scriptsize{bas-11l}  &	98.6  &	29.9 \\
aer  &	30K  &	Eastern Arrernte  &	5K  &	Oceania  &	Latn  &	aer  &	100  &	- \\
azz  &	30K  &	Highland Puebla Nahuatl  &	200K  &	Americas  &	Latn  &	\scriptsize{azz-7l}  &	99.7  &	13.5 \\
dah  &	30K  &	Gwahatike  &	5K  &	Oceania  &	Latn  &	\scriptsize{aon-13l}  &	97.2  &	- \\
yml  &	30K  &	Iamalele  &	5K  &	Oceania  &	Latn  &	\scriptsize{fj-4l}  &	99.7  &	- \\
bzh  &	30K  &	Mapos Buang  &	50K  &	Oceania  &	Latn  &	\scriptsize{apb-13l}  &	97.5  &	- \\
kgp  &	30K  &	Kaingang  &	50K  &	Americas  &	Latn  &	\scriptsize{bzj-9l}  &	100  &	- \\
tpz  &	30K  &	Tinputz  &	5K  &	Oceania  &	Latn  &	\scriptsize{ach-10l}  &	99.3  &	- \\
bex  &	30K  &	Jur Modo  &	50K  &	Africa  &	Latn  &	\scriptsize{bbo-3l}  &	100  &	11.0 \\
mek  &	30K  &	Mekeo  &	50K  &	Oceania  &	Latn  &	\scriptsize{fj-4l}  &	99.1  &	- \\
sbl  &	30K  &	Botolan Sambal  &	50K  &	Asia  &	Latn  &	\scriptsize{kqp-4l}  &	100  &	23.0 \\
uvl  &	30K  &	Lote  &	5K  &	Oceania  &	Latn  &	\scriptsize{ata-5l}  &	98.4  &	- \\
abx  &	30K  &	Inabaknon  &	50K  &	Asia  &	Latn  &	\scriptsize{abx-11l}  &	96.4  &	- \\
ese  &	30K  &	Ese Ejja  &	5K  &	Americas  &	Latn  &	\scriptsize{ese-2l}  &	100  &	- \\
mcb  &	30K  &	Machiguenga  &	5K  &	Americas  &	Latn  &	\scriptsize{apu-13l}  &	100  &	- \\
pma  &	30K  &	Paama  &	5K  &	Oceania  &	Latn  &	\scriptsize{ktm-4l}  &	99.4  &	- \\
qvm  &	30K  &	M-Y-L Quechua  &	50K  &	Americas  &	Latn  &	\scriptsize{qub-7l}  &	99.6  &	15.6 \\
mbj  &	30K  &	Nadëb  &	5K  &	Americas  &	Latn  &	\scriptsize{bgp-13l}  &	99.2  &	- \\
gso  &	30K  &	Southwest Gbaya  &	200K  &	Africa  &	Latn  &	\scriptsize{bas-11l}  &	100  &	16.8 \\
chd  &	30K  &	Highland Oaxaca Chontal  &	5K  &	Americas  &	Latn  &	\scriptsize{chd-6l}  &	99.6  &	- \\
ssx  &	30K  &	Samberigi  &	5K  &	Oceania  &	Latn  &	\scriptsize{aui-14l}  &	99.9  &	- \\
zaa  &	30K  &	Sierra de Juárez Zapotec  &	5K  &	Americas  &	Latn  &	\scriptsize{zaa-5l}  &	100  &	- \\
bqp  &	30K  &	Busa  &	50K  &	Africa  &	Latn  &	\scriptsize{akp-4l}  &	99.7  &	- \\
kgf  &	30K  &	Kulungtfu-Yuanggeng-Tobo  &	5K  &	Oceania  &	Latn  &	\scriptsize{kgf-4l}  &	99.6  &	- \\
tvk  &	30K  &	Southeast Ambrym  &	5K  &	Oceania  &	Latn  &	\scriptsize{gaw-8l}  &	99.1  &	- \\
pah  &	30K  &	Tenharim-Parintintin-Diahoi  &	5K  &	Americas  &	Latn  &	\scriptsize{gn-6l}  &	99.7  &	- \\
mog  &	30K  &	Mongondow  &	200K  &	Asia  &	Latn  &	\scriptsize{awb-4l}  &	100  &	18.6 \\
zas  &	30K  &	Santo Domingo Albarradas Zapotec  &	5K  &	Americas  &	Latn  &	\scriptsize{jic-3l}  &	100  &	- \\
zpl  &	30K  &	Lachixío Zapotec  &	5K  &	Americas  &	Latn  &	\scriptsize{cbr-6l}  &	99.4  &	- \\
pib  &	30K  &	Yine  &	5K  &	Americas  &	Latn  &	pib  &	71.2  &	- \\
omw  &	30K  &	South Tairora  &	50K  &	Oceania  &	Latn  &	\scriptsize{amp-12l}  &	99.9  &	- \\
faa  &	30K  &	Fasu  &	5K  &	Oceania  &	Latn  &	\scriptsize{apr-8l}  &	99.6  &	- \\
dob  &	30K  &	Dobu  &	100K  &	Oceania  &	Latn  &	\scriptsize{aui-14l}  &	99.1  &	13.2 \\
dnw  &	30K  &	Western Dani  &	200K  &	Asia  &	Latn  &	\scriptsize{bmu-6l}  &	100  &	13.5 \\
gor  &	30K  &	Gorontalo  &	1M  &	Asia  &	Latn  &	\scriptsize{bco-8l}  &	100  &	18.4 \\
hak  &	30K  &	Hakka Chinese  &	30M  &	Asia  &	Latn  &	\scriptsize{cdo-5l}  &	91.6  &	28.0 \\
poe  &	30K  &	San Juan Atzingo Popoloca  &	5K  &	Americas  &	Latn  &	\scriptsize{mix-4l}  &	99.5  &	- \\
gwi  &	30K  &	Gwich'in  &	5K  &	Americas  &	Latn  &	\scriptsize{agr-3l}  &	99.9  &	- \\
npy  &	30K  &	Napu  &	5K  &	Asia  &	Latn  &	\scriptsize{gah-11l}  &	100  &	- \\
kpr  &	30K  &	Korafe-Yegha  &	5K  &	Oceania  &	Latn  &	\scriptsize{aom-10l}  &	97.6  &	- \\
vid  &	30K  &	Vidunda  &	50K  &	Africa  &	Latn  &	\scriptsize{cwe-10l}  &	100  &	17.1 \\
yva  &	30K  &	Yawa  &	50K  &	Asia  &	Latn  &	\scriptsize{wnc-3l}  &	100  &	- \\
llg  &	30K  &	Lole  &	50K  &	Asia  &	Latn  &	\scriptsize{bhw-9l}  &	94.6  &	- \\
sps  &	30K  &	Saposa  &	5K  &	Oceania  &	Latn  &	\scriptsize{apb-13l}  &	97.3  &	- \\
wmw  &	30K  &	Mwani  &	100K  &	Africa  &	Latn  &	\scriptsize{bep-4l}  &	99.5  &	19.9 \\
mif  &	30K  &	Mofu-Gudur  &	50K  &	Africa  &	Latn  &	\scriptsize{meq-3l}  &	100  &	16.1 \\
fuf  &	30K  &	Pular  &	3M  &	Africa  &	Latn  &	\scriptsize{aa-13l}  &	99.9  &	20.9 \\
kbq  &	30K  &	Kamano  &	50K  &	Oceania  &	Latn  &	\scriptsize{aau-14l}  &	98.8  &	11.8 \\
fuv  &	30K  &	Nigerian Fulfulde  &	12M  &	Africa  &	Latn  &	\scriptsize{aa-13l}  &	98.1  &	- \\
ptu  &	30K  &	Bambam  &	50K  &	Asia  &	Latn  &	\scriptsize{atq-2l}  &	99.7  &	- \\
pri  &	30K  &	Paicî  &	5K  &	Oceania  &	Latn  &	\scriptsize{ae-Latn-10l}  &	97.2  &	- \\
en-Deva  &	30K  &	English  &	550M  &	Europe  &	Deva  &	\scriptsize{awa-14l}  &	97.1  &	75.9 \\
ymm  &	30K  &	Maay  &	3M  &	Africa  &	Latn  &	\scriptsize{aa-13l}  &	99.8  &	- \\
bmh  &	30K  &	Kein  &	5K  &	Oceania  &	Latn  &	\scriptsize{acn-11l}  &	98.1  &	- \\
zpo  &	30K  &	Amatlán Zapotec  &	50K  &	Americas  &	Latn  &	\scriptsize{zaa-5l}  &	88.4  &	- \\
smk  &	30K  &	Bolinao  &	50K  &	Asia  &	Latn  &	\scriptsize{br-4l}  &	99.0  &	28.6 \\
amp  &	30K  &	Alamblak  &	5K  &	Oceania  &	Latn  &	\scriptsize{amp-12l}  &	97.6  &	- \\
mic  &	30K  &	Mi'kmaq  &	5K  &	Americas  &	Latn  &	\scriptsize{hni-8l}  &	100  &	- \\
gkn  &	30K  &	Gokana  &	50K  &	Africa  &	Latn  &	\scriptsize{bcq-Latn-12l}  &	100  &	22.4 \\
cbs  &	30K  &	Cashinahua  &	5K  &	Americas  &	Latn  &	cbs  &	100  &	- \\
gvc  &	30K  &	Kotiria  &	5K  &	Americas  &	Latn  &	\scriptsize{des-4l}  &	100  &	- \\
xla  &	30K  &	Kamula  &	5K  &	Oceania  &	Latn  &	\scriptsize{ktm-4l}  &	99.6  &	- \\
sab  &	30K  &	Buglere  &	50K  &	Americas  &	Latn  &	sab  &	100  &	- \\
spp  &	30K  &	Supyire Senoufo  &	500K  &	Africa  &	Latn  &	spp  &	100  &	12.0 \\
gvn  &	30K  &	Kuku-Yalanji  &	5K  &	Oceania  &	Latn  &	\scriptsize{gvn-6l}  &	99.6  &	- \\
trc  &	30K  &	Copala Triqui  &	50K  &	Americas  &	Latn  &	\scriptsize{cut-5l}  &	88.0  &	- \\
thk  &	30K  &	Tharaka  &	200K  &	Africa  &	Latn  &	thk  &	100  &	20.9 \\
poy  &	30K  &	Pogolo  &	200K  &	Africa  &	Latn  &	\scriptsize{dov-9l}  &	100  &	15.7 \\
gej  &	30K  &	Gen  &	300K  &	Africa  &	Latn  &	\scriptsize{gej-4l}  &	100  &	24.1 \\
aak  &	30K  &	Ankave  &	5K  &	Oceania  &	Latn  &	\scriptsize{aak-3l}  &	100  &	- \\
rnl  &	30K  &	Halam  &	5K  &	Asia  &	Latn  &	\scriptsize{akb-4l}  &	100  &	- \\
kpf  &	30K  &	Komba  &	50K  &	Oceania  &	Latn  &	\scriptsize{avt-12l}  &	98.6  &	- \\
ghs  &	30K  &	Guhu-Samane  &	50K  &	Oceania  &	Latn  &	\scriptsize{agg-5l}  &	99.2  &	- \\
ajg  &	30K  &	Aja (Benin)  &	600K  &	Africa  &	Latn  &	\scriptsize{ajg-8l}  &	99.8  &	29.7 \\
bmu  &	30K  &	Burum-Mindik  &	5K  &	Oceania  &	Latn  &	\scriptsize{bmu-6l}  &	97.4  &	- \\
fuq  &	30K  &	Central-Eastern Niger Fulfulde  &	500K  &	Africa  &	Latn  &	\scriptsize{aa-13l}  &	100  &	16.3 \\
gvf  &	30K  &	Golin  &	50K  &	Oceania  &	Latn  &	\scriptsize{aon-13l}  &	98.2  &	12.6 \\
cbv  &	30K  &	Kakua  &	5K  &	Americas  &	Latn  &	\scriptsize{cbv-6l}  &	100  &	- \\
ake  &	30K  &	Akawaio  &	50K  &	Americas  &	Latn  &	ake  &	100  &	- \\
usp  &	30K  &	Uspanteco  &	5K  &	Americas  &	Latn  &	\scriptsize{acr-4l}  &	100  &	- \\
kpj  &	30K  &	Karajá  &	5K  &	Americas  &	Latn  &	\scriptsize{kpj-5l}  &	98.6  &	- \\
mtp  &	30K  &	Wichí Lhamtés Nocten  &	5K  &	Americas  &	Latn  &	\scriptsize{kqp-4l}  &	100  &	- \\
tku  &	30K  &	Upper Necaxa Totonac  &	5K  &	Americas  &	Latn  &	\scriptsize{tku-2l}  &	99.6  &	- \\
wnc  &	30K  &	Wantoat  &	5K  &	Oceania  &	Latn  &	\scriptsize{wnc-3l}  &	99.8  &	- \\
sgw  &	30K  &	Sebat Bet Gurage  &	1M  &	Africa  &	Ethi  &	\scriptsize{gez-8l}  &	99.2  &	35.6 \\
pwg  &	20K  &	Gapapaiwa  &	5K  &	Oceania  &	Latn  &	\scriptsize{asg-10l}  &	97.7  &	- \\
huv  &	20K  &	San Mateo del Mar Huave  &	50K  &	Americas  &	Latn  &	huv  &	100  &	- \\
tly-Cyrl-RU  &	20K  &	North-Central Talysh  &	900K  &	Europe  &	Cyrl  &	\scriptsize{az-RU-13l}  &	99.8  &	28.7 \\
kmu  &	20K  &	Kanite  &	5K  &	Oceania  &	Latn  &	\scriptsize{gah-11l}  &	99.9  &	- \\
bhb  &	20K  &	Bhili  &	3M  &	Asia  &	Deva  &	\scriptsize{awa-14l}  &	100  &	44.2 \\
ljp  &	20K  &	Lampung Api  &	800K  &	Asia  &	Latn  &	\scriptsize{hui-3l}  &	100  &	25.3 \\
wrs  &	20K  &	Waris  &	5K  &	Oceania  &	Latn  &	\scriptsize{acn-11l}  &	98.7  &	- \\
ttc  &	20K  &	Tektiteko  &	5K  &	Americas  &	Latn  &	\scriptsize{azz-7l}  &	100  &	- \\
myb  &	20K  &	Mbay  &	50K  &	Africa  &	Latn  &	\scriptsize{kpj-5l}  &	99.1  &	14.2 \\
mhx  &	20K  &	Maru  &	50K  &	Asia  &	Latn  &	\scriptsize{lsi-2l}  &	100  &	15.9 \\
knv  &	20K  &	Tabo  &	5K  &	Oceania  &	Latn  &	\scriptsize{bco-8l}  &	99.9  &	- \\
mbl  &	20K  &	Maxakalí  &	5K  &	Americas  &	Latn  &	\scriptsize{arn-6l}  &	99.3  &	- \\
nhi  &	20K  &	Zacatlán-Ahuacatlán-Tepetzintla Nahuatl  &	50K  &	Americas  &	Latn  &	\scriptsize{bjr-4l}  &	98.7  &	- \\
tby  &	20K  &	Tabaru  &	50K  &	Asia  &	Latn  &	\scriptsize{gbi-8l}  &	100  &	- \\
soq  &	20K  &	Kanasi  &	5K  &	Oceania  &	Latn  &	\scriptsize{acn-11l}  &	98.8  &	- \\
agg  &	20K  &	Angor  &	5K  &	Oceania  &	Latn  &	\scriptsize{agg-5l}  &	99.0  &	- \\
kwd  &	20K  &	Kwaio  &	50K  &	Oceania  &	Latn  &	\scriptsize{knj-8l}  &	99.9  &	- \\
ssg  &	20K  &	Seimat  &	5K  &	Oceania  &	Latn  &	\scriptsize{gaw-8l}  &	96.3  &	- \\
zaj  &	20K  &	Zaramo  &	5K  &	Africa  &	Latn  &	\scriptsize{cwe-10l}  &	99.2  &	- \\
ahr  &	20K  &	Ahirani  &	2M  &	Asia  &	Deva  &	\scriptsize{ahr-13l}  &	99.1  &	32.3 \\
pmf  &	20K  &	Pamona  &	100K  &	Asia  &	Latn  &	\scriptsize{pmf-2l}  &	100  &	14.2 \\
mco  &	20K  &	Coatlán Mixe  &	50K  &	Americas  &	Latn  &	\scriptsize{dyo-3l}  &	100  &	- \\
txq  &	20K  &	Tii  &	50K  &	Asia  &	Latn  &	\scriptsize{bhw-9l}  &	94.5  &	- \\
lif  &	20K  &	Limbu  &	800K  &	Asia  &	Deva  &	lif  &	99.9  &	- \\
ycn  &	20K  &	Yucuna  &	5K  &	Americas  &	Latn  &	\scriptsize{acu-8l}  &	100  &	- \\
kln  &	20K  &	Kalenjin  &	9M  &	Africa  &	Latn  &	\scriptsize{cro-3l}  &	99.9  &	12.5 \\
xsu  &	20K  &	Sanumá  &	5K  &	Americas  &	Latn  &	xsu  &	100  &	- \\
nus  &	20K  &	Nuer  &	900K  &	Africa  &	Latn  &	\scriptsize{din-3l}  &	100  &	22.8 \\
nak  &	20K  &	Nakanai  &	50K  &	Oceania  &	Latn  &	\scriptsize{bch-10l}  &	98.9  &	- \\
lid  &	20K  &	Nyindrou  &	5K  &	Oceania  &	Latn  &	\scriptsize{gaw-8l}  &	99.2  &	- \\
spl  &	20K  &	Selepet  &	5K  &	Oceania  &	Latn  &	\scriptsize{gah-11l}  &	99.5  &	- \\
aom  &	20K  &	Ömie  &	5K  &	Oceania  &	Latn  &	\scriptsize{aom-10l}  &	99.9  &	- \\
nhg  &	20K  &	Tetelcingo Nahuatl  &	5K  &	Americas  &	Latn  &	\scriptsize{nhg-3l}  &	100  &	- \\
cpu  &	20K  &	Pichis Ashéninka  &	50K  &	Americas  &	Latn  &	\scriptsize{apu-13l}  &	96.9  &	13.3 \\
muh  &	20K  &	Mündü  &	50K  &	Africa  &	Latn  &	muh  &	100  &	- \\
nmz  &	20K  &	Nawdm  &	200K  &	Africa  &	Latn  &	\scriptsize{amf-6l}  &	99.9  &	14.7 \\
taj  &	20K  &	Eastern Tamang  &	300K  &	Asia  &	Deva  &	\scriptsize{taj-2l}  &	100  &	18.3 \\
nca  &	20K  &	Iyo  &	5K  &	Oceania  &	Latn  &	\scriptsize{kgf-4l}  &	99.3  &	- \\
ik  &	20K  &	Alaskan Inupiaq  &	5K  &	Americas  &	Latn  &	\scriptsize{bla-7l}  &	99.6  &	- \\
gam  &	20K  &	Kandawo  &	5K  &	Oceania  &	Latn  &	\scriptsize{gam-3l}  &	99.1  &	- \\
kue  &	20K  &	Kuman  &	100K  &	Oceania  &	Latn  &	\scriptsize{agt-6l}  &	99.1  &	13.1 \\
bbr  &	20K  &	Girawa  &	5K  &	Oceania  &	Latn  &	\scriptsize{acn-11l}  &	96.0  &	- \\
roo  &	20K  &	Rotokas  &	5K  &	Oceania  &	Latn  &	\scriptsize{amp-12l}  &	91.7  &	- \\
big  &	20K  &	Biangai  &	50K  &	Oceania  &	Latn  &	\scriptsize{big-5l}  &	100  &	- \\
dip  &	20K  &	Northeastern Dinka  &	5M  &	Africa  &	Latn  &	\scriptsize{din-3l}  &	100  &	14.9 \\
aeb-Latn  &	20K  &	Tunisian Arabic  &	11M  &	Africa  &	Latn  &	\scriptsize{aeb-Latn-11l}  &	76.9  &	15.4 \\
lee  &	20K  &	Lyélé  &	100K  &	Africa  &	Latn  &	lee  &	99.6  &	12.4 \\
bgs  &	20K  &	Tagabawa  &	50K  &	Asia  &	Latn  &	\scriptsize{bgs-9l}  &	92.3  &	- \\
apb  &	20K  &	Sa'a  &	50K  &	Oceania  &	Latn  &	\scriptsize{apb-13l}  &	99.5  &	- \\
kmh  &	20K  &	Kalam  &	50K  &	Oceania  &	Latn  &	\scriptsize{dtb-4l}  &	99.5  &	- \\
soy  &	20K  &	Miyobe  &	50K  &	Africa  &	Latn  &	soy  &	100  &	- \\
gfk  &	20K  &	Patpatar  &	5K  &	Oceania  &	Latn  &	\scriptsize{acn-11l}  &	98.4  &	- \\
nvm  &	20K  &	Namiae  &	5K  &	Oceania  &	Latn  &	\scriptsize{dgz-5l}  &	99.1  &	- \\
bgr  &	20K  &	Bawm Chin  &	50K  &	Asia  &	Latn  &	\scriptsize{bgr-6l}  &	100  &	- \\
gnn  &	20K  &	Gumatj  &	5K  &	Oceania  &	Latn  &	\scriptsize{bch-10l}  &	100  &	- \\
qxo  &	20K  &	Southern Conchucos Ancash Quechua  &	900K  &	Americas  &	Latn  &	\scriptsize{qub-7l}  &	98.9  &	17.1 \\
cbr  &	20K  &	Cashibo-Cacataibo  &	5K  &	Americas  &	Latn  &	\scriptsize{cbr-6l}  &	100  &	- \\
bhl  &	20K  &	Bimin  &	5K  &	Oceania  &	Latn  &	\scriptsize{aey-6l}  &	99.6  &	- \\
amr  &	20K  &	Amarakaeri  &	5K  &	Americas  &	Latn  &	amr  &	100  &	- \\
kde  &	20K  &	Makonde  &	1M  &	Africa  &	Latn  &	\scriptsize{cwe-10l}  &	98.3  &	24.4 \\
zad  &	20K  &	Cajonos Zapotec  &	5K  &	Americas  &	Latn  &	\scriptsize{ctp-4l}  &	100  &	- \\
bps  &	20K  &	Sarangani Blaan  &	200K  &	Asia  &	Latn  &	\scriptsize{bpr-7l}  &	97.4  &	15.3 \\
mks  &	20K  &	Silacayoapan Mixtec  &	50K  &	Americas  &	Latn  &	mks  &	99.6  &	- \\
muy  &	20K  &	Muyang  &	50K  &	Africa  &	Latn  &	\scriptsize{dgc-6l}  &	100  &	12.8 \\
ncu  &	20K  &	Chumburung  &	50K  &	Africa  &	Latn  &	ncu  &	100  &	15.0 \\
stn  &	20K  &	Owa  &	5K  &	Oceania  &	Latn  &	\scriptsize{apr-8l}  &	100  &	- \\
sd-Deva  &	20K  &	Sindhi  &	68M  &	Asia  &	Deva  &	\scriptsize{bfz-15l}  &	98.3  &	34.4 \\
hub  &	20K  &	Huambisa  &	5K  &	Americas  &	Latn  &	\scriptsize{acu-8l}  &	98.8  &	- \\
cpb  &	20K  &	Ucayali-Yurúa Ashéninka  &	5K  &	Americas  &	Latn  &	\scriptsize{apu-13l}  &	97.0  &	14.2 \\
mio  &	20K  &	Pinotepa Nacional Mixtec  &	50K  &	Americas  &	Latn  &	\scriptsize{cta-4l}  &	100  &	- \\
chz  &	20K  &	Ozumacín Chinantec  &	5K  &	Americas  &	Latn  &	\scriptsize{chz-9l}  &	100  &	- \\
aia  &	20K  &	Arosi  &	5K  &	Oceania  &	Latn  &	aia  &	100  &	- \\
ian  &	20K  &	Iatmul  &	5K  &	Oceania  &	Latn  &	\scriptsize{agg-5l}  &	99.9  &	- \\
way  &	20K  &	Wayana  &	5K  &	Americas  &	Latn  &	way  &	100  &	- \\
ndz  &	20K  &	Ndogo  &	50K  &	Africa  &	Latn  &	\scriptsize{bcq-Latn-12l}  &	100  &	- \\
agd  &	20K  &	Agarabi  &	50K  &	Oceania  &	Latn  &	\scriptsize{acn-11l}  &	98.5  &	- \\
mcp  &	20K  &	Makaa  &	50K  &	Africa  &	Latn  &	mcp  &	100  &	15.0 \\
wbp  &	20K  &	Warlpiri  &	5K  &	Oceania  &	Latn  &	\scriptsize{gvn-6l}  &	99.9  &	- \\
mox  &	20K  &	Molima  &	5K  &	Oceania  &	Latn  &	\scriptsize{aui-14l}  &	96.2  &	- \\
tbc  &	20K  &	Takia  &	50K  &	Oceania  &	Latn  &	\scriptsize{aau-14l}  &	97.7  &	12.4 \\
sxb  &	20K  &	Suba  &	100K  &	Africa  &	Latn  &	\scriptsize{cgg-15l}  &	99.9  &	25.0 \\
kbm  &	20K  &	Iwal  &	5K  &	Oceania  &	Latn  &	\scriptsize{gam-3l}  &	99.8  &	- \\
bon  &	20K  &	Bine  &	5K  &	Oceania  &	Latn  &	bon  &	99.7  &	- \\
pls  &	20K  &	San Marcos Tlalcoyalco Popoloca  &	50K  &	Americas  &	Latn  &	\scriptsize{knj-8l}  &	100  &	- \\
vmy  &	20K  &	Ayautla Mazatec  &	5K  &	Americas  &	Latn  &	\scriptsize{nhr-2l}  &	98.6  &	- \\
nop  &	20K  &	Numanggang  &	5K  &	Oceania  &	Latn  &	\scriptsize{aau-14l}  &	98.7  &	- \\
qvn  &	20K  &	North Junín Quechua  &	50K  &	Americas  &	Latn  &	qvn  &	100  &	17.1 \\
cgc  &	20K  &	Kagayanen  &	50K  &	Asia  &	Latn  &	\scriptsize{cgc-2l}  &	100  &	22.4 \\
kwj  &	20K  &	Kwanga  &	50K  &	Oceania  &	Latn  &	\scriptsize{avt-12l}  &	97.3  &	- \\
kcg  &	20K  &	Tyap  &	100K  &	Africa  &	Latn  &	\scriptsize{kcg-9l}  &	100  &	13.9 \\
enx  &	20K  &	Southern Lengua  &	5K  &	Americas  &	Latn  &	\scriptsize{enx-4l}  &	99.0  &	- \\
mfy  &	20K  &	Mayo  &	50K  &	Americas  &	Latn  &	\scriptsize{ese-2l}  &	99.9  &	- \\
aby  &	20K  &	Aneme Wake  &	5K  &	Oceania  &	Latn  &	\scriptsize{aby-4l}  &	100  &	- \\
cco  &	20K  &	Comaltepec Chinantec  &	50K  &	Americas  &	Latn  &	cco  &	100  &	- \\
boa  &	20K  &	Bora  &	5K  &	Americas  &	Latn  &	boa  &	100  &	- \\
ikw  &	20K  &	Ikwere  &	200K  &	Africa  &	Latn  &	\scriptsize{anw-8l}  &	99.9  &	20.1 \\
lmn-Knda  &	20K  &	Lambadi  &	6M  &	Asia  &	Knda  &	\scriptsize{lmn-Knda-2l}  &	100  &	33.3 \\
rai  &	20K  &	Ramoaaina  &	50K  &	Oceania  &	Latn  &	\scriptsize{apb-13l}  &	95.9  &	- \\
leu  &	20K  &	Kara (Papua New Guinea)  &	5K  &	Oceania  &	Latn  &	\scriptsize{apb-13l}  &	99.1  &	- \\
sue  &	20K  &	Suena  &	5K  &	Oceania  &	Latn  &	\scriptsize{avt-12l}  &	99.6  &	- \\
auc  &	20K  &	Waorani  &	5K  &	Americas  &	Latn  &	\scriptsize{auc-6l}  &	100  &	- \\
kdc  &	20K  &	Kutu  &	50K  &	Africa  &	Latn  &	\scriptsize{cwe-10l}  &	97.7  &	19.1 \\
awb  &	20K  &	Awa (Papua New Guinea)  &	5K  &	Oceania  &	Latn  &	\scriptsize{awb-4l}  &	99.9  &	- \\
kpw  &	20K  &	Kobon  &	50K  &	Oceania  &	Latn  &	\scriptsize{aon-13l}  &	99.6  &	- \\
mux  &	20K  &	Bo-Ung  &	50K  &	Oceania  &	Latn  &	\scriptsize{gbi-8l}  &	99.5  &	11.2 \\
mqb  &	20K  &	Mbuko  &	50K  &	Africa  &	Latn  &	\scriptsize{blw-2l}  &	100  &	- \\
kne  &	20K  &	Kankanaey  &	200K  &	Asia  &	Latn  &	\scriptsize{bnc-8l}  &	100  &	25.7 \\
blz  &	20K  &	Balantak  &	50K  &	Asia  &	Latn  &	\scriptsize{bjp-4l}  &	100  &	14.2 \\
vmw  &	20K  &	Makhuwa  &	3M  &	Africa  &	Latn  &	\scriptsize{apu-13l}  &	99.9  &	16.2 \\
tcs  &	20K  &	Torres Strait Creole  &	5K  &	Oceania  &	Latn  &	\scriptsize{bg-Latn-5l}  &	99.4  &	- \\
mie  &	20K  &	Ocotepec Mixtec  &	50K  &	Americas  &	Latn  &	mie  &	99.9  &	- \\
mhy  &	20K  &	Ma'anyan  &	200K  &	Asia  &	Latn  &	\scriptsize{amo-7l}  &	99.5  &	34.7 \\
btd  &	20K  &	Batak Dairi  &	1M  &	Asia  &	Latn  &	\scriptsize{aom-10l}  &	100  &	21.5 \\
lem  &	20K  &	Nomaande  &	5K  &	Africa  &	Latn  &	\scriptsize{box-3l}  &	100  &	- \\
hop  &	20K  &	Hopi  &	5K  &	Americas  &	Latn  &	hop  &	99.9  &	- \\
ken  &	20K  &	Kenyang  &	50K  &	Africa  &	Latn  &	\scriptsize{ken-3l}  &	97.3  &	- \\
ssd  &	20K  &	Siroi  &	5K  &	Oceania  &	Latn  &	\scriptsize{bch-10l}  &	99.1  &	- \\
kms  &	20K  &	Kamasau  &	5K  &	Oceania  &	Latn  &	\scriptsize{cwt-4l}  &	97.4  &	- \\
yuw  &	20K  &	Yau (Morobe Province)  &	5K  &	Oceania  &	Latn  &	\scriptsize{amp-12l}  &	97.6  &	- \\
agt  &	20K  &	Central Cagayan Agta  &	5K  &	Asia  &	Latn  &	\scriptsize{agt-6l}  &	97.6  &	- \\
maj  &	20K  &	Jalapa De Díaz Mazatec  &	50K  &	Americas  &	Latn  &	\scriptsize{kcg-9l}  &	100  &	- \\
tpm  &	20K  &	Tampulma  &	50K  &	Africa  &	Latn  &	\scriptsize{sil-3l}  &	100  &	- \\
khs  &	20K  &	Kasua  &	5K  &	Oceania  &	Latn  &	\scriptsize{aui-14l}  &	99.9  &	- \\
kzf  &	20K  &	Da'a Kaili  &	400K  &	Asia  &	Latn  &	\scriptsize{kzf-3l}  &	100  &	16.6 \\
ign  &	20K  &	Ignaciano  &	50K  &	Americas  &	Latn  &	\scriptsize{acu-8l}  &	100  &	11.3 \\
mwv  &	20K  &	Mentawai  &	50K  &	Asia  &	Latn  &	\scriptsize{bjp-4l}  &	100  &	14.7 \\
cul  &	20K  &	Culina  &	5K  &	Americas  &	Latn  &	cul  &	100  &	- \\
apu  &	20K  &	Apurinã  &	5K  &	Americas  &	Latn  &	\scriptsize{apu-13l}  &	99.9  &	- \\
ncj  &	20K  &	Northern Puebla Nahuatl  &	50K  &	Americas  &	Latn  &	\scriptsize{bjr-4l}  &	100  &	18.6 \\
xnn  &	20K  &	Northern Kankanay  &	50K  &	Asia  &	Latn  &	\scriptsize{bnc-8l}  &	99.2  &	24.4 \\
bvd  &	20K  &	Baeggu  &	5K  &	Oceania  &	Latn  &	\scriptsize{bvc-4l}  &	99.1  &	- \\
nko  &	20K  &	Nkonya  &	50K  &	Africa  &	Latn  &	\scriptsize{ktj-6l}  &	100  &	- \\
cwe  &	20K  &	Kwere  &	50K  &	Africa  &	Latn  &	\scriptsize{cwe-10l}  &	97.6  &	21.7 \\
tap  &	20K  &	Taabwa  &	300K  &	Africa  &	Latn  &	tap  &	100  &	23.5 \\
wrk  &	20K  &	Garrwa  &	5K  &	Oceania  &	Latn  &	wrk  &	100  &	- \\
gah  &	20K  &	Alekano  &	50K  &	Oceania  &	Latn  &	\scriptsize{gah-11l}  &	97.6  &	- \\
yam  &	20K  &	Yamba  &	50K  &	Africa  &	Latn  &	\scriptsize{pny-2l}  &	100  &	7.0 \\
geb  &	20K  &	Kire  &	5K  &	Oceania  &	Latn  &	\scriptsize{bmu-6l}  &	100  &	- \\
ctp  &	20K  &	Western Highland Chatino  &	50K  &	Americas  &	Latn  &	\scriptsize{ctp-4l}  &	83.9  &	- \\
mbs  &	20K  &	Sarangani Manobo  &	50K  &	Asia  &	Latn  &	\scriptsize{mbb-3l}  &	99.6  &	16.0 \\
nlg  &	20K  &	Gela  &	50K  &	Oceania  &	Latn  &	\scriptsize{bgt-4l}  &	99.0  &	- \\
nsn  &	20K  &	Nehan  &	5K  &	Oceania  &	Latn  &	\scriptsize{bgt-4l}  &	99.4  &	- \\
bao  &	20K  &	Waimaha  &	5K  &	Americas  &	Latn  &	bao  &	99.7  &	- \\
cle  &	20K  &	Lealao Chinantec  &	5K  &	Americas  &	Latn  &	\scriptsize{chz-9l}  &	100  &	- \\
mvn  &	20K  &	Minaveha  &	5K  &	Oceania  &	Latn  &	\scriptsize{gah-11l}  &	98.2  &	- \\
tpt  &	20K  &	Tlachichilco Tepehua  &	5K  &	Americas  &	Latn  &	\scriptsize{gam-3l}  &	100  &	- \\
zyp  &	20K  &	Zyphe  &	50K  &	Asia  &	Latn  &	\scriptsize{arn-6l}  &	99.9  &	- \\
pio  &	20K  &	Piapoco  &	5K  &	Americas  &	Latn  &	pio  &	100  &	- \\
bgx-Latn  &	20K  &	Rumelian Turkish  &	300K  &	Europe  &	Latn  &	\scriptsize{aeb-Latn-11l}  &	95.6  &	59.5 \\
myx  &	20K  &	Masaaba  &	3M  &	Africa  &	Latn  &	\scriptsize{cgg-15l}  &	99.8  &	25.0 \\
tsz  &	20K  &	Purepecha  &	50K  &	Americas  &	Latn  &	\scriptsize{apu-13l}  &	100  &	15.2 \\
dww  &	20K  &	Dawawa  &	5K  &	Oceania  &	Latn  &	\scriptsize{abt-3l}  &	98.3  &	- \\
gnw  &	20K  &	Western Bolivian Guaraní  &	5K  &	Americas  &	Latn  &	gnw  &	100  &	- \\
ksr  &	20K  &	Borong  &	5K  &	Oceania  &	Latn  &	\scriptsize{kpj-5l}  &	98.5  &	- \\
otn  &	20K  &	Tenango Otomi  &	50K  &	Americas  &	Latn  &	\scriptsize{kcg-9l}  &	100  &	15.1 \\
gdn  &	20K  &	Umanakaina  &	5K  &	Oceania  &	Latn  &	\scriptsize{asg-10l}  &	99.5  &	- \\
kmo  &	20K  &	Kwoma  &	5K  &	Oceania  &	Latn  &	\scriptsize{abt-3l}  &	100  &	- \\
nii  &	20K  &	Nii  &	50K  &	Oceania  &	Latn  &	nii  &	100  &	- \\
tsw  &	20K  &	Salka-Tsishingini  &	200K  &	Africa  &	Latn  &	\scriptsize{asg-10l}  &	98.1  &	15.9 \\
not  &	20K  &	Nomatsiguenga  &	5K  &	Americas  &	Latn  &	\scriptsize{cot-2l}  &	100  &	- \\
ury  &	20K  &	Orya  &	5K  &	Asia  &	Latn  &	\scriptsize{bzj-9l}  &	100  &	- \\
anv  &	20K  &	Denya  &	50K  &	Africa  &	Latn  &	anv  &	100  &	- \\
pbc  &	20K  &	Patamona  &	50K  &	Americas  &	Latn  &	pbc  &	100  &	- \\
blt-Latn  &	20K  &	Tai Dam  &	800K  &	Asia  &	Latn  &	\scriptsize{blt-Latn-4l}  &	99.7  &	24.4 \\
aii  &	20K  &	Assyrian Neo-Aramaic  &	800K  &	Europe  &	Syrc  &	\scriptsize{ach-10l}  &	100  &	35.7 \\
mhl  &	20K  &	Mauwake  &	5K  &	Oceania  &	Latn  &	\scriptsize{aon-13l}  &	65.5  &	- \\
myk  &	20K  &	Mamara Senoufo  &	700K  &	Africa  &	Latn  &	myk  &	100  &	13.2 \\
mee  &	20K  &	Mengen  &	5K  &	Oceania  &	Latn  &	\scriptsize{ata-5l}  &	99.8  &	- \\
kpx  &	20K  &	Mountain Koiali  &	5K  &	Oceania  &	Latn  &	\scriptsize{apr-8l}  &	99.1  &	- \\
udu  &	20K  &	Uduk  &	50K  &	Africa  &	Latn  &	\scriptsize{knj-8l}  &	100  &	- \\
mnp  &	20K  &	Min Bei Chinese  &	10M  &	Asia  &	Latn  &	\scriptsize{cdo-5l}  &	99.8  &	16.2 \\
alp  &	20K  &	Alune  &	50K  &	Asia  &	Latn  &	\scriptsize{alp-2l}  &	100  &	- \\
wiu  &	20K  &	Wiru  &	50K  &	Oceania  &	Latn  &	\scriptsize{gbi-8l}  &	99.0  &	- \\
gyr  &	20K  &	Guarayu  &	5K  &	Americas  &	Latn  &	\scriptsize{gn-6l}  &	100  &	- \\
meq  &	20K  &	Merey  &	50K  &	Africa  &	Latn  &	\scriptsize{meq-3l}  &	100  &	- \\
kjs  &	20K  &	East Kewa  &	50K  &	Oceania  &	Latn  &	\scriptsize{aau-14l}  &	95.3  &	11.4 \\
mcq  &	20K  &	Ese  &	50K  &	Oceania  &	Latn  &	\scriptsize{acn-11l}  &	96.7  &	- \\
ong  &	20K  &	Olo  &	50K  &	Oceania  &	Latn  &	\scriptsize{kpz-2l}  &	99.9  &	- \\
tnn  &	20K  &	North Tanna  &	5K  &	Oceania  &	Latn  &	\scriptsize{nwi-5l}  &	99.9  &	- \\
coe  &	20K  &	Koreguaje  &	5K  &	Americas  &	Latn  &	coe  &	99.9  &	- \\
hrx  &	20K  &	Hunsrik  &	3M  &	Americas  &	Latn  &	\scriptsize{hrx-2l}  &	99.6  &	48.8 \\
tte  &	20K  &	Bwanabwana  &	5K  &	Oceania  &	Latn  &	\scriptsize{aui-14l}  &	98.3  &	- \\
yut  &	20K  &	Yopno  &	5K  &	Oceania  &	Latn  &	\scriptsize{dtb-4l}  &	99.2  &	- \\
sgz  &	20K  &	Sursurunga  &	5K  &	Oceania  &	Latn  &	\scriptsize{apb-13l}  &	97.7  &	- \\
ceg  &	20K  &	Chamacoco  &	5K  &	Americas  &	Latn  &	\scriptsize{ceg-5l}  &	100  &	- \\
qul  &	20K  &	North Bolivian Quechua  &	100K  &	Americas  &	Latn  &	\scriptsize{ae-Latn-10l}  &	99.6  &	20.9 \\
mvp  &	20K  &	Duri  &	100K  &	Asia  &	Latn  &	\scriptsize{kje-6l}  &	99.7  &	14.4 \\
xed  &	20K  &	Hdi  &	50K  &	Africa  &	Latn  &	\scriptsize{cko-5l}  &	100  &	- \\
bqj  &	20K  &	Bandial  &	50K  &	Africa  &	Latn  &	\scriptsize{bkv-4l}  &	99.6  &	- \\
duo  &	20K  &	Dupaninan Agta  &	5K  &	Asia  &	Latn  &	\scriptsize{dgc-6l}  &	99.5  &	- \\
mza  &	20K  &	Santa María Zacatepec Mixtec  &	5K  &	Americas  &	Latn  &	\scriptsize{mpm-4l}  &	99.6  &	- \\
mva  &	20K  &	Manam  &	5K  &	Oceania  &	Latn  &	\scriptsize{gah-11l}  &	97.7  &	- \\
akh  &	20K  &	Angal Heneng  &	50K  &	Oceania  &	Latn  &	\scriptsize{aau-14l}  &	97.8  &	12.0 \\
blw  &	20K  &	Balangao  &	50K  &	Asia  &	Latn  &	\scriptsize{blw-2l}  &	92.5  &	- \\
luy  &	20K  &	Luhya  &	1M  &	Africa  &	Latn  &	\scriptsize{cgg-15l}  &	99.7  &	20.6 \\
mnb  &	20K  &	Muna  &	300K  &	Asia  &	Latn  &	\scriptsize{ata-5l}  &	100  &	13.5 \\
nou  &	20K  &	Ewage-Notu  &	50K  &	Oceania  &	Latn  &	\scriptsize{aom-10l}  &	99.9  &	- \\
ebk  &	20K  &	Eastern Bontok  &	50K  &	Asia  &	Latn  &	\scriptsize{bnc-8l}  &	97.7  &	- \\
tpp  &	20K  &	Pisaflores Tepehua  &	5K  &	Americas  &	Latn  &	\scriptsize{tee-2l}  &	100  &	- \\
wim  &	20K  &	Wik-Mungkan  &	5K  &	Oceania  &	Latn  &	\scriptsize{beu-2l}  &	99.9  &	- \\
avt  &	20K  &	Au  &	5K  &	Oceania  &	Latn  &	\scriptsize{avt-12l}  &	98.3  &	- \\
blh  &	20K  &	Kuwaa  &	50K  &	Africa  &	Latn  &	\scriptsize{bcq-Latn-12l}  &	100  &	- \\
mfi  &	20K  &	Wandala  &	50K  &	Africa  &	Latn  &	\scriptsize{acu-8l}  &	100  &	13.5 \\
eri  &	20K  &	Ogea  &	5K  &	Oceania  &	Latn  &	\scriptsize{agt-6l}  &	98.3  &	- \\
mpt  &	20K  &	Mian  &	5K  &	Oceania  &	Latn  &	\scriptsize{amp-12l}  &	98.1  &	- \\
swp  &	20K  &	Suau  &	5K  &	Oceania  &	Latn  &	\scriptsize{apb-13l}  &	99.4  &	- \\
nho  &	20K  &	Takuu  &	5K  &	Oceania  &	Latn  &	\scriptsize{bgt-4l}  &	96.0  &	- \\
aso  &	20K  &	Dano  &	50K  &	Oceania  &	Latn  &	aso  &	100  &	11.6 \\
wnu  &	20K  &	Usan  &	5K  &	Oceania  &	Latn  &	\scriptsize{amp-12l}  &	96.7  &	- \\
zav  &	20K  &	Yatzachi Zapotec  &	5K  &	Americas  &	Latn  &	\scriptsize{cut-5l}  &	100  &	- \\
bcw  &	20K  &	Bana  &	50K  &	Africa  &	Latn  &	\scriptsize{bcw-3l}  &	100  &	- \\
dgz  &	20K  &	Daga  &	5K  &	Oceania  &	Latn  &	\scriptsize{dgz-5l}  &	99.8  &	- \\
ipi  &	20K  &	Ipili  &	50K  &	Oceania  &	Latn  &	\scriptsize{gbi-8l}  &	99.5  &	- \\
nsu  &	20K  &	Sierra Negra Nahuatl  &	50K  &	Americas  &	Latn  &	\scriptsize{nhy-3l}  &	99.5  &	- \\
urt  &	20K  &	Urat  &	5K  &	Oceania  &	Latn  &	\scriptsize{aey-6l}  &	99.1  &	- \\
obo  &	20K  &	Obo Manobo  &	50K  &	Asia  &	Latn  &	\scriptsize{atd-4l}  &	100  &	16.1 \\
zpi  &	20K  &	Santa María Quiegolani Zapotec  &	5K  &	Americas  &	Latn  &	\scriptsize{zaa-5l}  &	100  &	- \\
sat-Deva  &	20K  &	Santali  &	6M  &	Asia  &	Deva  &	\scriptsize{bho-13l}  &	98.1  &	- \\
cjv  &	20K  &	Chuave  &	50K  &	Oceania  &	Latn  &	\scriptsize{bom-5l}  &	97.5  &	- \\
ifu  &	20K  &	Mayoyao Ifugao  &	50K  &	Asia  &	Latn  &	\scriptsize{bnc-8l}  &	100  &	- \\
atg  &	20K  &	Ivbie North-Okpela-Arhe  &	50K  &	Africa  &	Latn  &	\scriptsize{ajg-8l}  &	100  &	- \\
icr  &	20K  &	San Andres Creole English  &	50K  &	Americas  &	Latn  &	\scriptsize{bzj-9l}  &	100  &	- \\
neb  &	20K  &	Toura (Côte d'Ivoire)  &	50K  &	Africa  &	Latn  &	\scriptsize{avn-3l}  &	100  &	8.9 \\
apr  &	20K  &	Arop-Lokep  &	5K  &	Oceania  &	Latn  &	\scriptsize{apr-8l}  &	99.2  &	- \\
aoz  &	20K  &	Uab Meto  &	800K  &	Asia  &	Latn  &	\scriptsize{aaz-6l}  &	100  &	15.5 \\
mca  &	20K  &	Maca  &	5K  &	Americas  &	Latn  &	\scriptsize{hrx-2l}  &	100  &	- \\
tiy  &	20K  &	Tiruray  &	50K  &	Asia  &	Latn  &	\scriptsize{cjp-4l}  &	100  &	13.2 \\
gdr  &	20K  &	Wipi  &	5K  &	Oceania  &	Latn  &	\scriptsize{aon-13l}  &	98.4  &	- \\
kqw  &	20K  &	Kandas  &	5K  &	Oceania  &	Latn  &	\scriptsize{apb-13l}  &	93.8  &	- \\
zaw  &	20K  &	Mitla Zapotec  &	50K  &	Americas  &	Latn  &	\scriptsize{toj-4l}  &	100  &	- \\
atq  &	20K  &	Aralle-Tabulahan  &	50K  &	Asia  &	Latn  &	\scriptsize{atq-2l}  &	99.9  &	- \\
ruf  &	20K  &	Luguru  &	700K  &	Africa  &	Latn  &	\scriptsize{cwe-10l}  &	100  &	17.3 \\
kje  &	20K  &	Kisar  &	50K  &	Asia  &	Latn  &	\scriptsize{kje-6l}  &	100  &	- \\
fai  &	20K  &	Faiwol  &	5K  &	Oceania  &	Latn  &	\scriptsize{bnj-6l}  &	99.8  &	- \\
zpq  &	20K  &	Zoogocho Zapotec  &	5K  &	Americas  &	Latn  &	\scriptsize{cut-5l}  &	100  &	- \\
mcn  &	20K  &	Masana  &	200K  &	Africa  &	Latn  &	\scriptsize{dzg-5l}  &	50.7  &	47.6 \\
nlc  &	20K  &	Nalca  &	50K  &	Asia  &	Latn  &	\scriptsize{mps-3l}  &	100  &	- \\
yby  &	20K  &	Yaweyuha  &	5K  &	Oceania  &	Latn  &	\scriptsize{aau-14l}  &	97.6  &	- \\
pao  &	20K  &	Northern Paiute  &	5K  &	Americas  &	Latn  &	\scriptsize{chq-3l}  &	99.9  &	- \\
mcu  &	20K  &	Donga Mambila  &	50K  &	Africa  &	Latn  &	\scriptsize{mcu-2l}  &	100  &	7.6 \\
kbc  &	20K  &	Kadiwéu  &	5K  &	Americas  &	Latn  &	\scriptsize{etr-2l}  &	100  &	- \\
djr  &	20K  &	Djambarrpuyngu  &	5K  &	Oceania  &	Latn  &	\scriptsize{bch-10l}  &	100  &	- \\
zae  &	20K  &	Yareni Zapotec  &	5K  &	Americas  &	Latn  &	\scriptsize{rmc-3l}  &	100  &	- \\
wer  &	20K  &	Weri  &	50K  &	Oceania  &	Latn  &	\scriptsize{acn-11l}  &	99.2  &	- \\
maw  &	20K  &	Mampruli  &	200K  &	Africa  &	Latn  &	\scriptsize{cko-5l}  &	100  &	14.1 \\
kud  &	20K  &	Auhelawa  &	5K  &	Oceania  &	Latn  &	\scriptsize{aui-14l}  &	98.0  &	- \\
dad  &	20K  &	Marik  &	5K  &	Oceania  &	Latn  &	\scriptsize{amp-12l}  &	94.4  &	- \\
cax  &	20K  &	Lomeriano-Ignaciano Chiquitano  &	5K  &	Americas  &	Latn  &	\scriptsize{acu-8l}  &	100  &	- \\
ons  &	20K  &	Ono  &	50K  &	Oceania  &	Latn  &	\scriptsize{kgf-4l}  &	100  &	- \\
wuv  &	20K  &	Wuvulu-Aua  &	5K  &	Oceania  &	Latn  &	\scriptsize{aom-10l}  &	98.6  &	- \\
gay  &	20K  &	Gayo  &	300K  &	Asia  &	Latn  &	\scriptsize{abs-13l}  &	97.2  &	31.9 \\
gng  &	20K  &	Ngangam  &	50K  &	Africa  &	Latn  &	\scriptsize{ada-13l}  &	100  &	15.4 \\
nim  &	20K  &	Nilamba  &	700K  &	Africa  &	Latn  &	\scriptsize{ess-4l}  &	100  &	16.5 \\
snc  &	20K  &	Sinaugoro  &	50K  &	Oceania  &	Latn  &	\scriptsize{gah-11l}  &	97.6  &	- \\
nhu  &	20K  &	Noone  &	50K  &	Africa  &	Latn  &	\scriptsize{cme-5l}  &	100  &	10.9 \\
daa  &	20K  &	Dangaleat  &	50K  &	Africa  &	Latn  &	\scriptsize{daa-3l}  &	100  &	12.3 \\
yli  &	20K  &	Angguruk Yali  &	50K  &	Asia  &	Latn  &	yli  &	100  &	- \\
bgq-Arab  &	20K  &	Bagri  &	2M  &	Asia  &	Arab  &	\scriptsize{bgq-Arab-4l}  &	99.7  &	- \\
tnk  &	20K  &	Kwamera  &	5K  &	Oceania  &	Latn  &	\scriptsize{nwi-5l}  &	99.4  &	- \\
knf  &	20K  &	Mankanya  &	50K  &	Africa  &	Latn  &	\scriptsize{knf-2l}  &	100  &	13.5 \\
mbh  &	20K  &	Mangseng  &	5K  &	Oceania  &	Latn  &	\scriptsize{acn-11l}  &	98.3  &	- \\
cme  &	20K  &	Cerma  &	50K  &	Africa  &	Latn  &	\scriptsize{cme-5l}  &	100  &	10.3 \\
dgr  &	20K  &	Dogrib  &	5K  &	Americas  &	Latn  &	\scriptsize{bcq-Latn-12l}  &	99.8  &	- \\
kup  &	20K  &	Kunimaipa  &	50K  &	Oceania  &	Latn  &	\scriptsize{bjr-4l}  &	99.5  &	- \\
lww  &	20K  &	Lewo  &	5K  &	Oceania  &	Latn  &	\scriptsize{big-5l}  &	99.5  &	- \\
ktb  &	20K  &	Kambaata  &	900K  &	Africa  &	Ethi  &	\scriptsize{gez-8l}  &	99.9  &	13.5 \\
dop  &	20K  &	Lukpa  &	50K  &	Africa  &	Latn  &	dop  &	100  &	- \\
kki  &	20K  &	Kagulu  &	200K  &	Africa  &	Latn  &	\scriptsize{dov-9l}  &	100  &	17.3 \\
mpp  &	20K  &	Migabac  &	5K  &	Oceania  &	Latn  &	\scriptsize{bru-6l}  &	99.3  &	- \\
sua  &	20K  &	Sulka  &	5K  &	Oceania  &	Latn  &	\scriptsize{bwu-4l}  &	98.7  &	- \\
yon  &	20K  &	Yonggom  &	5K  &	Oceania  &	Latn  &	\scriptsize{wnc-3l}  &	100  &	- \\
eka  &	15K  &	Ekajuk  &	50K  &	Africa  &	Latn  &	\scriptsize{apn-3l}  &	100  &	12.6 \\
mih  &	15K  &	Chayuco Mixtec  &	50K  &	Americas  &	Latn  &	\scriptsize{gaw-8l}  &	99.5  &	- \\
cux  &	15K  &	Tepeuxila Cuicatec  &	50K  &	Americas  &	Latn  &	cux  &	100  &	- \\
tlb  &	15K  &	Tobelo  &	50K  &	Asia  &	Latn  &	\scriptsize{gbi-8l}  &	100  &	- \\
klv  &	15K  &	Maskelynes  &	5K  &	Oceania  &	Latn  &	\scriptsize{bkv-4l}  &	99.9  &	- \\
due  &	15K  &	Umiray Dumaget Agta  &	5K  &	Asia  &	Latn  &	\scriptsize{dgc-6l}  &	100  &	- \\
lmn-Deva  &	15K  &	Lambadi  &	6M  &	Asia  &	Deva  &	\scriptsize{awa-14l}  &	98.4  &	24.8 \\
xuo  &	15K  &	Kuo  &	50K  &	Africa  &	Latn  &	xuo  &	100  &	- \\
shk  &	15K  &	Shilluk  &	200K  &	Africa  &	Latn  &	shk  &	100  &	- \\
xsb  &	15K  &	Tinà Sambal  &	50K  &	Asia  &	Latn  &	\scriptsize{dgc-6l}  &	100  &	21.8 \\
amk  &	15K  &	Ambai  &	50K  &	Asia  &	Latn  &	\scriptsize{aby-4l}  &	100  &	- \\
zab  &	14K  &	San Juan Guelavía Zapotec  &	50K  &	Americas  &	Latn  &	\scriptsize{toj-4l}  &	99.7  &	- \\
mig  &	14K  &	San Miguel El Grande Mixtec  &	50K  &	Americas  &	Latn  &	mig  &	60.8  &	- \\
hla  &	14K  &	Halia  &	50K  &	Oceania  &	Latn  &	\scriptsize{aon-13l}  &	97.6  &	- \\
dgc  &	14K  &	Casiguran-Nagtipunan Agta  &	5K  &	Asia  &	Latn  &	\scriptsize{dgc-6l}  &	99.1  &	- \\
iry  &	14K  &	Iraya  &	50K  &	Asia  &	Latn  &	\scriptsize{agn-14l}  &	96.8  &	- \\
vun  &	14K  &	Vunjo  &	900K  &	Africa  &	Latn  &	\scriptsize{nnq-4l}  &	99.9  &	15.9 \\
hot  &	14K  &	Hote  &	5K  &	Oceania  &	Latn  &	\scriptsize{aey-6l}  &	93.1  &	- \\
otd  &	14K  &	Ot Danum  &	50K  &	Asia  &	Latn  &	\scriptsize{bhw-9l}  &	100  &	- \\
ade  &	14K  &	Adele  &	50K  &	Africa  &	Latn  &	\scriptsize{ade-4l}  &	100  &	- \\
nin  &	14K  &	Ninzo  &	50K  &	Africa  &	Latn  &	\scriptsize{ajg-8l}  &	100  &	13.4 \\
gaw  &	14K  &	Nobonob  &	5K  &	Oceania  &	Latn  &	\scriptsize{gaw-8l}  &	100  &	- \\
nwi  &	14K  &	Southwest Tanna  &	5K  &	Oceania  &	Latn  &	\scriptsize{nwi-5l}  &	100  &	- \\
naf  &	14K  &	Nabak  &	50K  &	Oceania  &	Latn  &	\scriptsize{miq-2l}  &	98.1  &	- \\
ztq  &	14K  &	Quioquitani-Quieri Zapotec  &	5K  &	Americas  &	Latn  &	\scriptsize{gv-3l}  &	99.7  &	- \\
bkd  &	14K  &	Talaandig-Binukid  &	50K  &	Asia  &	Latn  &	\scriptsize{abx-11l}  &	93.9  &	- \\
mit  &	14K  &	Southern Puebla Mixtec  &	5K  &	Americas  &	Latn  &	mit  &	100  &	- \\
kbr  &	14K  &	Kafa  &	800K  &	Africa  &	Latn  &	\scriptsize{drs-Latn-8l}  &	100  &	12.8 \\
cnt  &	14K  &	Tepetotutla Chinantec  &	5K  &	Americas  &	Latn  &	\scriptsize{chz-9l}  &	100  &	- \\
bdd  &	14K  &	Bunama  &	5K  &	Oceania  &	Latn  &	\scriptsize{aui-14l}  &	97.3  &	- \\
bm-Nkoo  &	14K  &	Bambara  &	14M  &	Africa  &	Nkoo  &	\scriptsize{bm-Nkoo-2l}  &	78.3  &	12.1 \\
myw  &	14K  &	Muyuw  &	5K  &	Oceania  &	Latn  &	\scriptsize{aon-13l}  &	97.9  &	- \\
box  &	14K  &	Buamu  &	200K  &	Africa  &	Latn  &	\scriptsize{box-3l}  &	100  &	13.1 \\
prf  &	13K  &	Paranan  &	50K  &	Asia  &	Latn  &	\scriptsize{dgc-6l}  &	99.6  &	- \\
ziw  &	13K  &	Zigula-Mushungulu  &	400K  &	Africa  &	Latn  &	\scriptsize{cwe-10l}  &	98.3  &	17.6 \\
maq  &	13K  &	Chiquihuitlán Mazatec  &	5K  &	Americas  &	Latn  &	\scriptsize{chz-9l}  &	99.9  &	- \\
yrb  &	13K  &	Yareba  &	5K  &	Oceania  &	Latn  &	\scriptsize{aby-4l}  &	100  &	- \\
lwo  &	13K  &	Luwo  &	50K  &	Africa  &	Latn  &	\scriptsize{chq-3l}  &	100  &	13.2 \\
ozm  &	13K  &	Koonzime  &	50K  &	Africa  &	Latn  &	\scriptsize{mcu-2l}  &	100  &	9.5 \\
csk  &	13K  &	Jola-Esulalu  &	50K  &	Africa  &	Latn  &	\scriptsize{csk-3l}  &	100  &	13.0 \\
kdl  &	13K  &	Tsikimba  &	50K  &	Africa  &	Latn  &	\scriptsize{asg-10l}  &	99.0  &	16.3 \\
hno  &	13K  &	Northern Hindko  &	4M  &	Asia  &	Arab  &	\scriptsize{bal-13l}  &	40.5  &	32.3 \\
waj  &	13K  &	Waffa  &	5K  &	Oceania  &	Latn  &	waj  &	99.8  &	- \\
kkc  &	12K  &	Odoodee  &	5K  &	Oceania  &	Latn  &	\scriptsize{adh-5l}  &	99.8  &	- \\
mil  &	12K  &	Peñoles Mixtec  &	50K  &	Americas  &	Latn  &	\scriptsize{mib-2l}  &	100  &	- \\
ndj  &	12K  &	Ndamba  &	50K  &	Africa  &	Latn  &	\scriptsize{dov-9l}  &	100  &	15.9 \\
ngp  &	12K  &	Ngulu  &	100K  &	Africa  &	Latn  &	\scriptsize{cwe-10l}  &	98.3  &	16.3 \\
okv  &	11K  &	Orokaiva  &	50K  &	Oceania  &	Latn  &	\scriptsize{aom-10l}  &	99.1  &	14.6 \\
nnq  &	11K  &	Ngindo  &	200K  &	Africa  &	Latn  &	\scriptsize{nnq-4l}  &	99.9  &	13.9 \\
grt  &	11K  &	Garo  &	1M  &	Asia  &	Beng  &	\scriptsize{bpy-7l}  &	100  &	- \\
psh  &	11K  &	Southwest Pashayi  &	50K  &	Asia  &	Arab  &	\scriptsize{fay-14l}  &	15.5  &	38.0 \\
hnd  &	11K  &	Southern Hindko  &	600K  &	Asia  &	Arab  &	\scriptsize{bal-13l}  &	17.3  &	- \\
tgp  &	11K  &	Movono  &	5K  &	Oceania  &	Latn  &	\scriptsize{akb-4l}  &	100  &	- \\
tbo  &	10K  &	Tawala  &	50K  &	Oceania  &	Latn  &	\scriptsize{aui-14l}  &	98.5  &	- \\
bxh  &	9K  &	Buhutu  &	5K  &	Oceania  &	Latn  &	\scriptsize{aui-14l}  &	93.2  &	- \\
mfa  &	7K  &	Kelantan-Pattani Malay  &	1M  &	Asia  &	Arab  &	\scriptsize{mfa-3l}  &	95.9  &	23.2 \\
 \label{tab:datasize_full}
\end{longtable}
\end{footnotesize}
}

\insertgianttable

\end{document}